%% file: neurips_2026.tex
\documentclass{article}

\usepackage[main, final]{neurips_2026}

\newcommand{\boxednum}[1]{%
  \setlength{\fboxsep}{0.5pt}%
  \fcolorbox{black}{black}{\textcolor{white}{#1}}%
}

\usepackage{amsmath}
\usepackage{amssymb}
\usepackage{mathtools}
\usepackage{amsthm}

\usepackage[utf8]{inputenc} 
\usepackage[T1]{fontenc}    
\usepackage{hyperref}       
\usepackage{url}            
\usepackage{booktabs}       
\usepackage{amsfonts}       
\usepackage{nicefrac}       
\usepackage{microtype}      
\usepackage{xcolor}         
\usepackage{graphicx}
\usepackage{subcaption}
\usepackage{enumitem}
\usepackage{wrapfig}
\usepackage{float}
\usepackage{placeins}

\usepackage{algorithm}
\usepackage{algorithmic}

\title{Steering Away from Memorization:
Reachability-Constrained Reinforcement Learning for Text-to-Image Diffusion}

\author{%
  Sathwik Karnik$^{1,*}$
  \hspace{0.5em}
  Juyeop Kim$^{2,*}$
  \hspace{0.5em}
  Sanmi Koyejo$^{1,\dagger}$
  \hspace{0.5em}
  Jong-Seok Lee$^{2,\dagger}$
  \hspace{0.5em}
  Somil Bansal$^{1,\dagger}$
  \\[12pt]
  $^{1}$Stanford University, United States
  \\
  $^{2}$School of Integrated Technology \&
  BK21 Graduate Program in \\
  Intelligent Semiconductor Technology,
  Yonsei University, Korea
  \\[6pt]
  {\small\texttt{\{sathwik,sanmi,somil\}@stanford.edu}}
  \\
  {\small\texttt{\{juyeopkim,jong-seok.lee\}@yonsei.ac.kr}}
}

\begin{document}

\renewcommand{\thefootnote}{*}
\footnotetext{Equal contribution.
\textsuperscript{$\dagger$}Correspondence:
Somil Bansal, Jong-Seok Lee, and Sanmi Koyejo.}
\renewcommand{\thefootnote}{\arabic{footnote}}

\maketitle

\input{macros}

\vspace{-12pt}
\begin{abstract}
\input{sec/0_abstract}
\end{abstract}

\input{sec/1_introduction}
\input{sec/2_related_work}
\input{sec/3_preliminaries}

\input{sec/4_method}
\input{sec/5_experiments}

\input{sec/6_results}
\input{sec/7_discussion_and_limitations}
\input{sec/8_acknowledgements}

\bibliography{bib/ref,bib/bansal_papers,bib/reachability}
\bibliographystyle{icml2026}

\newpage
\appendix
\onecolumn
\input{sec/appendix}
\newpage

\end{document}

%% file: macros.tex
\newcommand{\rads}{\textsc{rads}}

\newcommand{\controlspace}{\mathcal{U}}
\newcommand{\failureset}{\mathcal{F}}
\newcommand{\latentspace}{\mathcal{X}}
\newcommand{\imagespace}{\mathcal{Y}}
\newcommand{\fddim}{f_{\textsc{ddim}}}
\newcommand{\brt}{\mathcal{B}}
\newcommand{\gaussian}{\mathcal{N}(\mathbf{0}, \mathbf{I})}

\newcommand{\image}{\mathbf{y}}
\newcommand{\latent}{\mathbf{x}}
\newcommand{\prompt}{c}
\newcommand{\promptemb}{\mathbf{e}_{c}}
\newcommand{\promptembt}{\mathbf{e}_{c}'}
\newcommand{\promptembsigma}{\mathbf{e}_{c,\sigma}}
\newcommand{\promptembspace}{\mathcal{C}}
\newcommand{\noisepredictor}{\epsilon_\theta}
\newcommand{\newpromptemb}{\tilde{\mathbf{e}}}
\newcommand{\dyndm}{f_{\text{DM}}}

\newcommand{\mdp}{\mathcal{M}}
\newcommand{\statespace}{\mathcal{S}}
\newcommand{\actionspace}{\mathcal{A}}
\newcommand{\targetfunction}{\ell}
\newcommand{\state}{\mathbf{s}}
\newcommand{\control}{\mathbf{u}}
\newcommand{\noise}{\mathbf{\omega}}
\newcommand{\statecaption}{\mathbf{c}}

\newcommand{\clipim}{\textsc{clip}_{\text{im}}}
\newcommand{\cliptext}{\textsc{clip}_{\text{text}}}
\newcommand{\sscdprompts}{\text{SSCD}_{\text{prompts}}}
\newcommand{\sscdseeds}{\text{SSCD}_{\text{seeds}}}
\newcommand{\sscdtarget}{\text{SSCD}_{\text{target}}}
\newcommand{\clip}{\text{CLIP}}
\newcommand{\fid}{\text{FID}}

\newcommand{\Qsafe}{Q^{\text{safe}}}       
\newcommand{\Qsafep}{Q^{\text{safe}}_\psi} 

\newcommand{\Qtask}{Q^{\text{task}}}       
\newcommand{\Qtaskp}{Q^{\text{task}}_\omega} 

\newcommand{\hatQsafe}{\hat{Q}^{\text{safe}}} 

\newcommand{\lagrange}{\lambda}            

\newcommand{\vaelatentspace}{\mathcal{Z}_\text{act}}
\newcommand{\enc}{\texttt{Enc}}
\newcommand{\dec}{\texttt{Dec}}
\newcommand{\vaelatent}{\mathbf{z}}

%% file: sec/0_abstract.tex
Text-to-image diffusion models are susceptible to memorization, revealing a fundamental failure to generalize beyond the training set. Current mitigation approaches often sacrifice image quality or prompt alignment to reduce memorization. To address this, we propose \underline{R}eachability-\underline{A}ware \underline{D}iffusion \underline{S}teering ($\rads$), an offline-trained, sampling-time control framework that mitigates memorization while maintaining generation fidelity. $\rads$ models the diffusion denoising process as a dynamical system and learns a policy-conditioned reachability value function that estimates future memorization risk along the denoising trajectory. This motivates a constrained reinforcement learning (RL) formulation, where a policy learns to steer the trajectory away from memorization via perturbations in the caption embedding space. Empirical evaluations show that $\rads$ achieves a stronger empirical Pareto frontier between generation diversity (SSCD), quality (FID), and alignment (CLIP) compared to state-of-the-art inference-time baselines. Crucially, $\rads$ applies the learned controller during sampling time, separating memorization mitigation from the pretrained model. 
Our website is available at: \url{https://s-karnik.github.io/rads-memorization-project-page/}.

%% file: sec/1_introduction.tex
\section{Introduction}
\vspace{-0.5em}
\begin{wrapfigure}{R}{0.5\textwidth}
    \centering
    \small

    \begin{subfigure}[t]{0.31\linewidth}
        \centering
        \includegraphics[width=\linewidth]{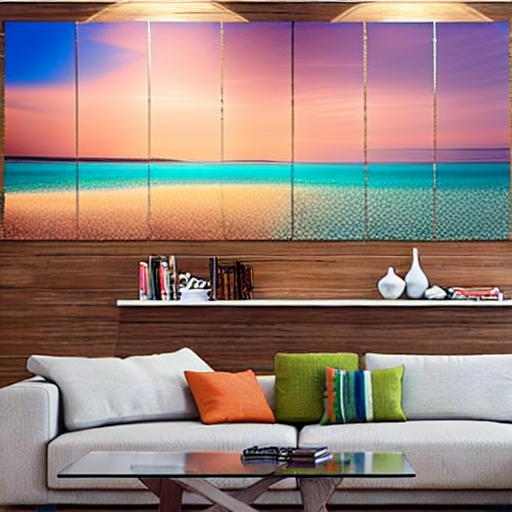}
        \caption{Memorized}
        \label{fig:teaser_memorized}
    \end{subfigure}
    \hfill
    \begin{subfigure}[t]{0.31\linewidth}
        \centering
        \includegraphics[width=\linewidth]{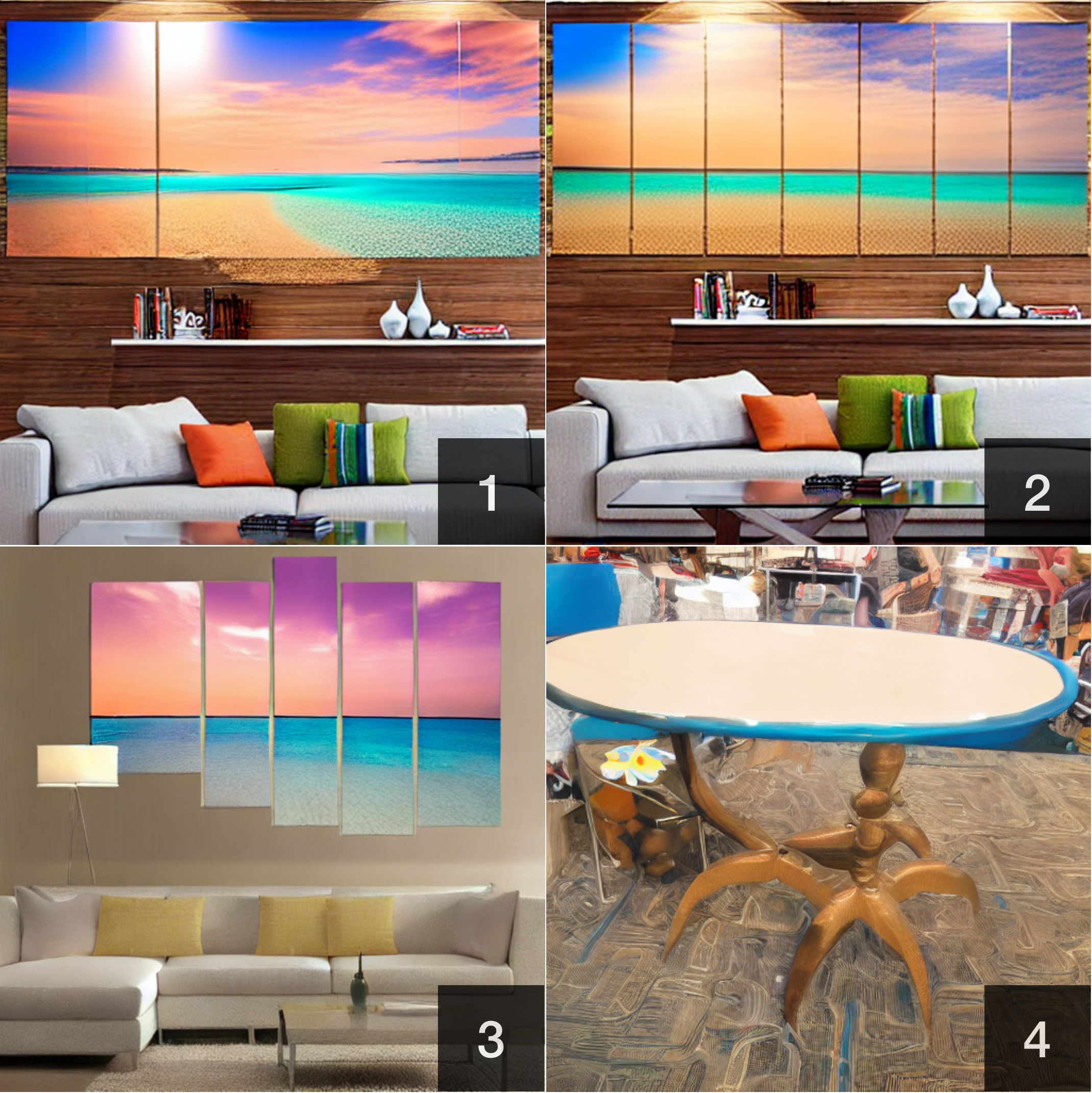}
        \caption{Prior work}
        \label{fig:teaser_prior_work}
    \end{subfigure}
    \hfill
    \begin{subfigure}[t]{0.31\linewidth}
        \centering
        \includegraphics[width=\linewidth]{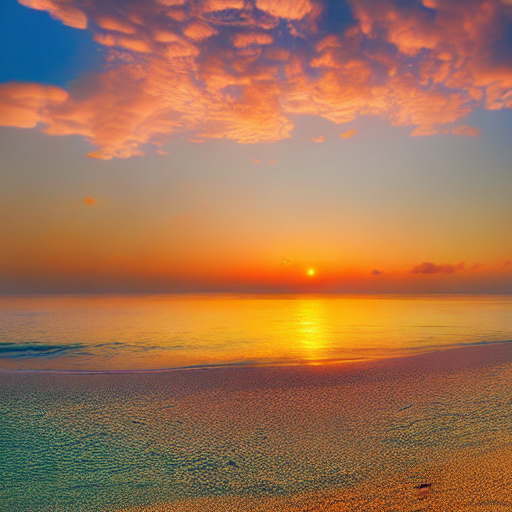}
        \caption{\textbf{$\rads$ {\footnotesize (Ours)}}}
        \label{fig:teaser_ours}
    \end{subfigure}

    \caption{
        \textbf{$\rads$ vs. Prior work.}
        Generated images for the prompt \textit{Designart Serene Maldives Seashore at Sunset Oversized Landscape Canvas Art - 4 Panels.}
        (a) Memorized image.
        (b) Results by prior work; \protect\boxednum{1}–\protect\boxednum{4} correspond to
        \protect\citeauthor{wen2024detecting, ren2024crossattn, hintersdorf2024nemo, jain2025attraction}
        (c) $\rads$ (ours).
        \vspace{-1em}
    }
    \label{fig:teaser}
\end{wrapfigure}

Diffusion models have undoubtedly emerged as the dominant paradigm for image generation in recent years~\citep{ho2020ddpm, song2021ddim, song2021score, dhariwal2021beatgan, rombach2022sdv1, ramesh2022dalle2}.
However, like other generative models~\citep{webster2021person, carlini2023quantifying, jagielski2023measuring, lee2023plagiarize}, diffusion models are susceptible to reproducing training data~\citep{somepalli2023forgery, carlini2023extracting, webster2023extraction}.
This phenomenon, called \textit{memorization}, is especially concerning in text-to-image generation, where natural language prompts can trigger the reproduction of copyrighted or private images~\citep{carlini2022onion, jiang2023art}.

To address this problem, prior work has proposed a range of mitigation strategies~\citep{wen2024detecting, ren2024crossattn, hintersdorf2024nemo, jain2025attraction}.
However, existing methods can leave substantial memorization or compromise image quality and prompt alignment.
This trade-off is illustrated in \figureautorefname~\ref{fig:teaser}.
\figureautorefname~\ref{fig:teaser_memorized} shows a memorized image generated by Stable Diffusion v1.4~\citep{rombach2022sdv1}, alongside outputs produced by representative prior mitigation methods in \figureautorefname~\ref{fig:teaser_prior_work}.
Prior methods preserve visual quality while retaining memorization, or reduce memorization at the cost of image quality or prompt alignment, while $\rads$ mitigates memorization while preserving prompt intent and image quality.

\begin{figure*}[!t]
    \centering
    \includegraphics[width=0.9\textwidth]{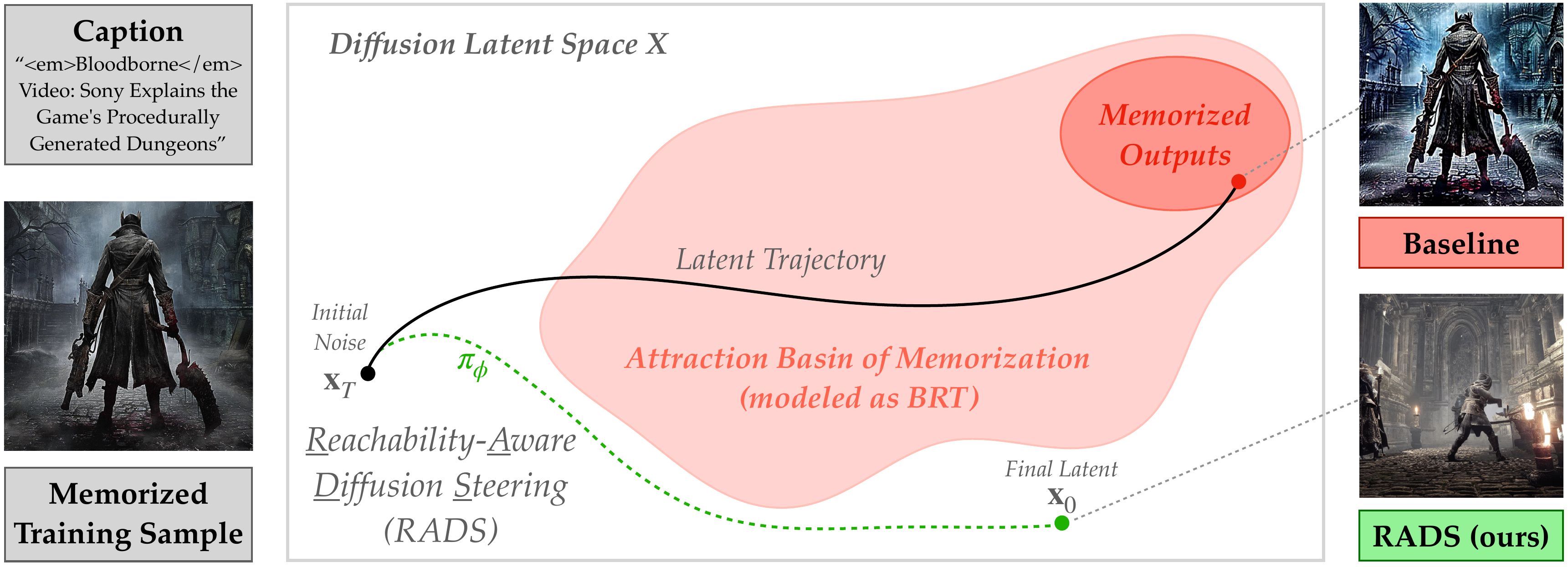}
    \caption{\textbf{Overview of $\rads$.} This diagram illustrates a real example of how $\rads$ prevents memorization in the Stable Diffusion v1.4 model, while the baseline (no mitigation) generates an image that closely resembles the training sample. $\rads$ does so by modeling the ``attraction basin'' of memorization as a backward reachable tube (BRT) and learning a policy $\pi_{\phi}$ for steering the caption embedding inputs to the diffusion model using reachability-constrained reinforcement learning.}
    \label{fig:rads_summary}
\end{figure*}

This motivates our central question:
\textit{how can we mitigate memorization without sacrificing image quality or prompt alignment?}
To answer this question, we introduce \emph{Reachability-Aware Diffusion Steering} (\rads), an offline-trained, sampling-time framework to actively reduce the risk of memorization (see Figure~\ref{fig:rads_summary}). 
Our core technique draws from \textit{reachability analysis}~\citep{bansal2017hamilton}, a method from control theory traditionally used to ensure safety in autonomous systems by identifying a system's ``failure set.'' 
Reachability analysis characterizes the ``backward reachable tube'' (BRT)---the set of all intermediate states from which a system will inevitably evolve into that failure set under its natural dynamics.
We adapt this concept to diffusion models by modeling the denoising process as a controlled dynamical system, where the latent noise represents the system state and the perturbed caption embeddings act as the control input. 
By learning a reachability-based safety value function, we can identify regions of the denoising trajectory that lead toward memorized images.

Building on this characterization, we formulate memorization mitigation as a constrained reinforcement learning (RL) problem. 
In this framework, the RL policy aims to maximize a reward signal based on image-prompt semantic alignment, while being constrained to avoid states within the BRT. 
By training this constrained optimization offline and applying the learned policy at inference time, $\rads$ learns to steer the diffusion trajectory away from memorization-inducing regions via perturbations in the caption embedding space. 
To the best of our knowledge, $\rads$ is the first framework to combine reachability analysis and RL to steer diffusion models away from memorization basins while preserving semantic alignment.

To summarize, our contributions are: (1) a reachability-based formulation of diffusion denoising as a controlled dynamical system over latent states and caption embeddings; (2) a reachability-constrained RL algorithm that trains a lightweight inference-time steering policy to mitigate memorization; (3) comprehensive experiments across multiple open-source diffusion models and datasets, demonstrating that $\rads$ achieves a stronger empirical Pareto frontier between generation diversity, perceptual quality, and prompt alignment compared to state-of-the-art inference-time baselines.

%% file: sec/2_related_work.tex
\section{Related Work}
\label{sec:related_work}

\textbf{Mitigating Memorization.} Prior work primarily addresses memorization through heuristic interventions on prompts or attention mechanisms. \citet{somepalli2023understanding} propose perturbing input prompts with random tokens to disrupt retrieval. Others focus on the diffusion architecture itself: \citet{ren2024crossattn} and \citet{hintersdorf2024nemo} mitigate memorization by masking cross-attention scores or disabling specific neurons responsible for reproducing training data. More recently, methods utilizing classifier-free guidance magnitude have emerged: \citet{wen2024detecting} replace tokens that trigger high guidance norms, while \citet{jain2025attraction} truncate guidance when an attraction basin is detected. Concurrent work by \citet{asthana2026detecting} studies memorization detection through anisotropy of the log-probability, combining classifier-free-guidance norm and angular-alignment signals for denoising-free detection and prompt-based mitigation. Unlike these discrete or prompt-level interventions, \ensuremath{\rads} learns a continuous, trajectory-level steering policy that controls caption embeddings throughout denoising, enabling state-dependent interventions that steer away from memorization while maintaining semantic alignment.

\textbf{Model Unlearning vs. Inference Steering.}
Another line of work mitigates undesired generations by modifying model weights, e.g., Erasing Stable Diffusion~\citep{gandikota2023erasing} and LoRA-based unlearning~\citep{kumari2023ablating, liu2025dyme}. However, memorization often involves idiosyncratic training instances rather than high-level concepts, making semantic unlearning targets difficult to specify. In contrast, $\rads$ operates at sampling time, steering trajectories away from memorization basins without modifying the diffusion backbone.

\textbf{Reachability in Safety-Critical Systems.}
We adapt reachability analysis \citep{bansal2021deepreach} to diffusion models.
Building on Reachability-Constrained Reinforcement Learning (RCRL) \citep{yu2022reachability}, which adapts Hamilton-Jacobi reachability to define feasible sets for safe policy learning, we propose the first application of latent reachability to the diffusion denoising process.
While prior work has applied similar concepts to large language models \citep{karnik2025preemptive}, $\rads$ learns a reachability-based value function to enable interventions before generation collapses.

\textbf{Inference-Time Steering via Reinforcement Learning.}
While RL has been used to fine-tune diffusion weights \citep{black2023training}, recent work explores \textit{inference-time steering} to guide generation without retraining.
Similar to PPLM \citep{dathathri2019plug} for LLMs, \citet{wagenmaker2025steering} propose DSRL to steer diffusion outputs toward external task rewards (e.g., robotic goals).
$\rads$ differs by formulating the \textit{internal} denoising process as a dynamical system, applying control directly to the caption embeddings to enforce safety constraints.
\vspace{-0.5em}

%% file: sec/3_preliminaries.tex
\section{Preliminaries: Denoising Diffusion}
\vspace{-0.5em}

Diffusion models commonly operate in a low-dimensional latent space $\latentspace$ learned by a well-trained autoencoder, which reduces computational cost compared to operating directly in the high-dimensional pixel space $\imagespace$~\citep{rombach2022sdv1}.
Instead of denoising directly in pixel space to produce an image $\image \in \imagespace$, diffusion models perform a gradual denoising process in latent space.
Starting from a random noise latent $\latent_T \in \latentspace$ sampled from $\gaussian$, the model iteratively denoises over $T$ steps (from timestep $\tau=T$ to $\tau=0$) to obtain $\latent_0$, which is then decoded by the autoencoder to produce the final image $\image$.
At each timestep $\tau$, denoising is performed by predicting the noise component to be removed from the intermediate latent $\latent_{\tau}$.
This prediction is made by a trained noise predictor $\noisepredictor$~\citep{ho2020ddpm}.
To enable conditional generation using text caption $c$, $\noisepredictor$ takes both $\latent_{\tau}$ and a caption embedding $\promptemb$ as input, derived from $c$ using CLIP~\citep{radford2021clip}.
The text embedding $\promptemb$ is injected into the model through cross-attention layers to guide the denoising process~\citep{rombach2022sdv1}.
The strength of text guidance is controlled by a guidance scale $g$~\citep{ho2021classifier}.

\vspace{-0.5em}

%% file: sec/4_method.tex
\section{Reachability-Aware Diffusion Steering}
\label{sec:method}

To mitigate memorization, we propose \emph{\underline{R}eachability-\underline{A}ware \underline{D}iffusion \underline{S}teering} ($\rads$).
We view the diffusion denoising process as a controlled dynamical system and aim to actively \textit{steer} the generation away from memorized outputs by introducing a control input $\control_t$.
\vspace{-0.5em}

\subsection{Denoising as a Controlled Dynamical System}
\label{sec:steering_space}

Since diffusion generation proceeds through sequential latent updates $\latent_{\tau} \rightarrow \latent_{\tau-1}$ governed by a fixed update rule, it can naturally be viewed as a dynamical system (e.g. via the lens of Langevin dynamics \citep{ho2020ddpm}).
In general, a dynamical system evolves as $\state_{t+1} = f(\state_t, \control_t, \noise_t)$ (from $0 \rightarrow T$), where $\state_t \in \statespace$ is the system state at time $t$, $\control_t \in \controlspace$ is the control input, $\noise_t$ is the noise (e.g., sampled noise in DDPM \citep{ho2020ddpm}), and $f$ is the update rule.
In this work, we denote the state $\state_t := (\latent_{T-t}, T-t)$, and we apply control inputs $\control_t$ for perturbing the base caption embedding $\promptemb$ to obtain the transient steered embedding $\promptembt$. Note that we treat the caption embedding $\promptemb$ as fixed context for each trajectory and suppress this conditioning in the notation. We denote the pre-trained diffusion update rule as $\latent_{T-(t+1)} = \dyndm(\latent_{T-t}, \promptembt, T-t)$. The complete system is defined as:
\begin{equation}
\small
\state_{t+1} = f(\state_t, \control_t, \noise_t) = \begin{bmatrix} 
    \latent_{T-(t+1)} \\ 
    T -(t+1) 
    \end{bmatrix} =
    \begin{bmatrix} 
    \dyndm(\latent_{T-t}, \promptembt, T-t) \\ 
    T -(t+1) 
    \end{bmatrix}
\end{equation}

\textbf{Remark:} In DDIM sampling, $\noise_t = 0$, while in DDPM sampling, $\noise_t \ne 0$.

Let $P(\cdot\mid s_t,u_t)$ denote the transition distribution
induced by these dynamics and the sampling noise $\omega_t$.
For DDIM sampling, this distribution is concentrated
at the next state specified by $f$.

\begin{figure}[!t]
    \centering
    \begin{subfigure}[t]{0.15\linewidth}
        \centering
        \includegraphics[width=\linewidth]{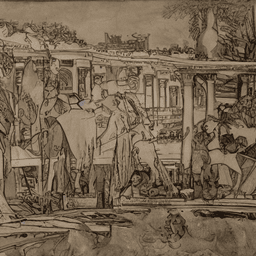}
        \caption{0 steps}
    \end{subfigure}
    \hspace{1.5em}
    \begin{subfigure}[t]{0.15\linewidth}
        \centering
        \includegraphics[width=\linewidth]{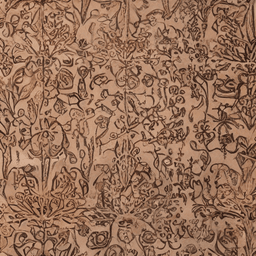}
        \caption{1 step}
    \end{subfigure}
    \hspace{1.5em}
    \begin{subfigure}[t]{0.15\linewidth}
        \centering
        \includegraphics[width=\linewidth]{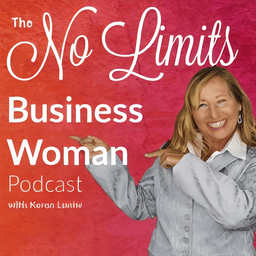}
        \caption{2 steps}
    \end{subfigure}
    \hspace{1.5em}
    \begin{subfigure}[t]{0.15\linewidth}
        \centering
        \includegraphics[width=\linewidth]{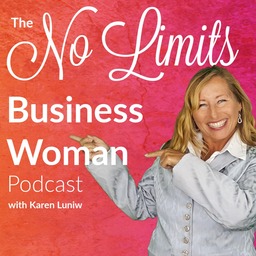}
        \caption{Memorized}
    \end{subfigure}
    \caption{
        \textbf{Memorization happens early.}
        Images generated with text guidance enabled for only the first
        $k$ steps ($k \in \{0,1,2\}$). Just 2 steps of guidance (c)
        are sufficient to reproduce the memorized image (d).
        Caption: \textit{The No Limits Business Woman~Podcast}.
    }
    \label{fig:toy_exp}
\end{figure}

\textbf{Choice of Action Space.}
Steering can be applied in either the image latent space $\latentspace$ or the caption embedding space (modifying $\promptemb$).
We select the caption space because text conditioning strongly influences memorization early in denoising \citep{kim2025diffusion}.
\figureautorefname~\ref{fig:toy_exp} illustrates this influence: in this example, applying guidance for only the first two steps is sufficient to regenerate the memorized image, even when guidance is disabled thereafter.
We further analyze this choice through random perturbations. We find that caption-space perturbations offer a more gradual mitigation-alignment tradeoff than image-latent perturbations on the evaluated prompts (Appendix~\ref{app:caption_latent_diagnostic}).
Thus, $\rads$ applies control to the caption embeddings to steer the denoising trajectory and reduce memorization.

\textbf{Controlled Dynamical System for Diffusion Denoising.} Standard text embeddings (e.g., CLIP) are high-dimensional (e.g., $77 \times 768$), resulting in a prohibitively large action space for steering.
To enable efficient steering, we learn a compact latent action space $\vaelatentspace \subseteq \mathbb{R}^d$ ($d \ll \text{dim}(\promptemb)$) using a variational autoencoder (VAE).
For the caption space $\promptembspace$ in which valid caption embeddings reside, we define an encoder $\enc: \promptembspace \to \vaelatentspace$ and a decoder $\dec: \vaelatentspace \to \promptembspace$.
Using this parameterization, we project the caption embedding $\promptemb$ into $\vaelatentspace$ to obtain $\enc(\promptemb)$, apply a perturbation in this projected space as $\enc(\promptemb) + \control_t$, and then decode it back into $\promptembspace$ to obtain the steered caption embedding $\promptembt = \dec(\enc(\promptemb) + \control_t)$.
Thus, the steered evolution of the denoising process follows the update:
\begin{equation}
\small
    \latent_{T-(t+1)} = \dyndm\Big(\latent_{T-t}, \underbrace{\dec(\enc(\promptemb) + \control_t)}_{\text{Steered caption embedding } \promptembt}, T-t\Big).
    \label{eq:caption_steering}
\end{equation}
\vspace{-1.5em}

\subsection{Memorization as a Reachability Problem}
\label{sec:memorization_as_reachability}
\vspace{-0.5em}
Diffusion-based image generation is inherently path-dependent: early latent states strongly bias the final output~\citep{choi2022perception, daras2022multiresolution, park2023riemannian, jain2025attraction, kim2025diffusion}.
If the denoising trajectory enters a ``basin of attraction'' corresponding to a memorized training example early in the denoising process, it can become increasingly difficult to recover and avoid memorization~\citep{jain2025attraction}.
This observation motivates the need for a framework that can identify and avoid such memorization-prone regions \emph{before} the image is fully generated.

To model these memorization-prone regions, we adopt \emph{reachability analysis}, drawing on tools from control theory traditionally used to reason about safety in dynamical systems.
Reachability addresses the following question: \textit{given a set of failure states and a steering policy, which initial states are predicted to evolve into this failure set under the system dynamics when following that policy?}

\vspace{-0.5em}

\paragraph{Failure Set and Backward Reachable Tube.}
In the context of memorization in diffusion models, we define the failure set $\mathcal{F}_{\mathrm{mem}} \subseteq \mathcal{S}$ as the set of states that decode to images closely resembling training examples. For a steering policy $\pi$ and deterministic closed-loop
dynamics, the policy-conditioned \emph{backward reachable tube (BRT)} is defined as $\brt^{\pi} =\{s_0\in\mathcal{S}\mid
\exists \sigma\in\{0,\ldots,T-1\},
s_\sigma^\pi\in\mathcal{F}_{\mathrm{mem}}\}$,
where $s_\sigma^\pi$ denotes the state reached by following $\pi$.
This set characterizes initial states from which the trajectory
enters the failure set under that policy; another policy may
avoid failure from the same initial state. In practice, exactly computing the BRT in high-dimensional latent space can be intractable.
Instead, we learn a discounted, policy-conditioned
safety value using a surrogate target.
\vspace{-1em}
\paragraph{Target Function and Policy-Conditioned Safety Value Function.}
We define a target function $\ell:\mathcal{S}\times\mathcal{U}
\rightarrow\mathbb{R}$ whose sub-zero level set defines a
surrogate failure set
$\mathcal{F}_{\ell}=\{(s,u)\in\mathcal{S}\times\mathcal{U}:
\ell(s,u)\leq 0\}$. For a realized trajectory, we define the minimum remaining safety margin $J_t = \min_{\sigma \in \{t,\ldots,T-1\}}
\ell(\state_\sigma,\control_\sigma)$, which is nonpositive if and only if a remaining denoising state-action pair enters the surrogate failure set. Let $\pi_\phi(\control\mid\state)$ denote the stochastic steering policy parameterized by $\phi$. Motivated by the minimum-over-time objective, we define a policy-conditioned safety value through the recursive backup:
\begin{equation}
Q_{\pi_\phi}^{\mathrm{safe}}(s_t,u_t)
=
\mathbb{E}_{\substack{
s_{t+1}\sim P(\cdot\mid s_t,u_t)\\
u_{t+1}\sim\pi_\phi(\cdot\mid s_{t+1})
}}
\left[
\min\left\{
\ell(s_t,u_t),
Q_{\pi_\phi}^{\mathrm{safe}}(s_{t+1},u_{t+1})
\right\}
\right].
\qquad t<T-1.
\end{equation}
The terminal condition is
$Q_{\pi_\phi}^{\mathrm{safe}}(\state_{T-1},\control_{T-1})
=\ell(\state_{T-1},\control_{T-1})$.

This value function provides a policy-conditioned safety margin for characterizing the memorization basin. 
In Section~\ref{subsec:csac}, we describe how we train $Q_{\pi_{\phi}}^{\text{safe}}(\state_t, \control_t)$ and enforce it as a constraint when training the steering policy $\pi_{\phi}$.
In doing so, the reachability-based safety critic provides a principled mechanism for identifying and avoiding memorization-inducing regions of the diffusion trajectory. 
\vspace{-1em}

\paragraph{Implemented Target Function.} In practice, we define the target function $\ell$ to detect and avoid memorization. To do so, we use the magnitude of the classifier-free guidance (CFG) vector. Prior work theoretically and empirically show that memorized generations often exhibit anomalously high guidance magnitudes, as the model overfits to the conditioning caption \citep{kim2025diffusion, jain2025attraction}.
Thus, we instantiate the target function $\ell(\cdot)$ to penalize extreme deviations between conditional and unconditional noise predictions:
\begin{equation}
\label{eqn:target_fn}
\small
\ell(\state_t, \control_t) = - \text{tanh}\left( \eta \cdot (\| \epsilon_\theta(\latent_{T-t}, \promptembt) - \epsilon_\theta(\latent_{T-t}, \emptyset) \|_2 - \beta) \right).
\end{equation}
The surrogate failure set $\mathcal{F}_{\ell}$ therefore consists of state-action pairs whose CFG norm is at least $\beta$, indicating elevated memorization risk. We use $\mathcal{F}_{\ell}$ as an empirically motivated surrogate: avoiding it does not guarantee avoidance of the actual memorization failure set $\mathcal{F}_{\mathrm{mem}}$. Here, $\eta$ controls the target's shaping scale and $\beta$ sets the CFG-norm threshold; Appendix~\ref{app:target_function_dets} details their selection. Because this target is based on a local CFG-norm signal, Appendix~\ref{app:reliability} evaluates how well the learned value function predicts future memorization risk.

Furthermore, we validate this surrogate failure set across SD v1.4, SD v2.1, and Realistic Vision: memorized prompts exhibit higher mean CFG norms across denoising steps (Appendix~\ref{app:cfg_validation}).

\vspace{-0.5em}

\subsection{Constrained Markov Decision Process}
\label{subsec:cmdp}
Having defined memorization-unsafe states via reachability, we now formulate memorization mitigation as a \emph{constrained Markov Decision Process (CMDP)}. The objective is to steer the diffusion denoising process during inference-time to maximize semantic alignment with a given text prompt while encouraging the generation to avoid memorization-prone trajectories.

\textbf{MDP Formulation.} We model the denoising process as a finite-horizon MDP $\mathcal{M} = (\statespace, \actionspace, P, r, T)$. Here, the state space remains $\statespace$, the action space $\actionspace$ is identically equal to $\controlspace$, and transition dynamics $P$ are the same as in Section~\ref{sec:steering_space}.
\vspace{-1em}

\paragraph{Constraint.} In this CMDP, we impose the safety constraint $\mathbb{E}_{\state \sim \mathcal{D},\, \control \sim \pi_\phi(\cdot \mid \state)}
[Q^{\text{safe}}(\state,\control)] \ge \delta$, where $\mathcal{D}$ is the replay buffer and $Q^{\text{safe}}$ is the discounted policy-conditioned surrogate from Section 4.2. This encourages the denoising trajectory to avoid memorization-prone regions identified by the learned safety value function.

\vspace{-1em}
\paragraph{Reward Function.} The primary task objective is to preserve semantic alignment between the generated image and the given text prompt. Let $\clipim(\image)$ denote the normalized CLIP embedding of the decoded image $\image$, and let $\cliptext$ denote the normalized CLIP embedding of the caption $c$. Then, $\clipim(\image) \cdot \cliptext(c)$ denotes the cosine similarity between the image and caption embeddings. We define a sparse terminal reward based on this cosine similarity:
\begin{equation}\small
\label{eqn:reward}
    r(\state_t) = 
    \begin{cases} 
        \clipim(\image_t) \cdot \cliptext(c) & \text{if } t = T-1 \\
        0 & \text{if } t < T-1
    \end{cases}
\end{equation}
\vspace{-1.5em}

\subsection{Constrained RL Solution (Soft Actor-Critic)}
\label{subsec:csac}
To solve the CMDP, we employ a constrained Soft Actor-Critic (SAC) algorithm~\citep{haarnoja2018soft} with Lagrangian relaxation. This approach enables sample-efficient, off-policy learning of a stochastic steering policy $\pi_{\phi}$ while enforcing the reachability-based safety constraint.
\vspace{-1em}
\paragraph{Architecture.} Our framework maintains three parameterized networks:
\vspace{-1em}
\begin{enumerate}[noitemsep, left=0pt]
    \item A \emph{stochastic policy} $\pi_\phi(\control \mid \state)$ with parameters $\phi$.
    \item A \emph{task critic} $\Qtaskp(\state, \control)$ with parameters $\omega$, estimating expected semantic-alignment return.
    \item A \emph{safety critic} $\Qsafep(\state, \control)$ with parameters $\psi$, estimating  reachability with respect to memorization.
\end{enumerate}
\vspace{-1em}
Additionally, we use a learnable parameter $\alpha$ to scale the entropy $\mathcal{H}(\pi)$, encouraging exploration.

\textbf{Safety Critic Learning (Reachability).} The safety critic $\Qsafe$ approximates a discounted variant of the recursive safety surrogate defined in Section~\ref{sec:memorization_as_reachability}. Given the immediate target function $\ell(\state_t, \control_t)$ and discount factor $\gamma$, we train the critic with a recursive target that combines current memorization risk with discounted future risk under the current policy.

Given a transition $(\state_t, \control_t, \state_{t+1})$ and a time horizon $T$, the target value $\hatQsafe_t$ is computed recursively:
\begin{equation}
\label{eqn:q_safe_target}
\small
    \hatQsafe_t = 
    \begin{cases} 
        \begin{aligned}
        (1-\gamma)\ell(\state_t, \control_t) + \gamma \min\big(&\ell(\state_t, \control_t),
        \Qsafe_{\psi'}(\state_{t+1}, \control'_{t+1})\big)
        \end{aligned} & \text{if } t < T-1 \\
        \ell(\state_{T-1}, \control_{T-1}) & \text{if } t = T -1
    \end{cases}
\end{equation}
where $\control'_{t+1} \sim \pi_\phi(\cdot|\state_{t+1})$ is the next action, and $\psi'$ denotes the parameters of the \emph{target} safety critic network (maintained via Polyak averaging). The parameters $\psi$ are updated to minimize the mean squared error $(\Qsafep(\state_t, \control_t) - \hatQsafe_t)^2$. 

\textbf{Task Critic Learning.}
The task critic $\Qtaskp$ is trained to estimate the cumulative task reward. Given the task reward $r_t$, we minimize the standard Bellman error
    $\mathcal{L}_{\text{task}}(\omega) = \mathbb{E} \big[ \big( \Qtaskp(\state_t, \control_t) - \big(r_t 
    + \gamma \mathbb{E}_{\state_{t+1}}[V^{\text{task}}(\state_{t+1})]\big) \big)^2\big]$, where $V^{\text{task}}$ denotes the value of the next state under the current policy.

\textbf{Policy and Dual Update.} 
We enforce the reachability constraint by requiring the expected safety value to meet a threshold $\delta$. We introduce a learnable Lagrange multiplier $\lagrange \geq 0$ to handle this constraint. The objective is defined over a distribution of states  sampled from the replay buffer $\mathcal{D}$:
\begin{equation}
J_\pi(\phi) = \mathbb{E}_{\state_t \sim \mathcal{D}, \control_t \sim \pi_\phi} \Big[ \Qtaskp(\state_t, \control_t) + \alpha \mathcal{H}(\pi_\phi(\cdot|\state_t)) +\lagrange \cdot (\Qsafep(\state_t, \control_t) - \delta) \Big]
\label{eq:policy_update}
\end{equation}
The value of $\lagrange$ is updated via dual gradient descent with gradient $\nabla_\lagrange J = \mathbb{E}[\Qsafep(\state, \control)] - \delta$. If the predicted safety margin falls below $\delta$, $\lagrange$ increases, biasing the policy toward safer steering.

Thus, unlike fixed heuristic interventions, $\rads$ learns a state-dependent control law that trades off prompt-image alignment against CFG-norm-based memorization risk throughout denoising.

%% file: sec/5_experiments.tex
\section{Experiments}
\label{sec:experiments}
\vspace{-0.5em}

\subsection{Experiment Setup}
\label{sec:exp_setup}
\textbf{Models and Datasets.}
Our benchmark evaluations cover three diffusion models: (i) Stable Diffusion (SD) v1.4~\citep{rombach2022sdv1} in both the DDIM \citep{song2021ddim} and DDPM \citep{ho2020ddpm} sampling settings; (ii) RealisticVision~\citep{civitai2023realvis} with DDIM sampling; (iii) SD v2.1~\citep{rombach2022sdv1} with DDIM sampling.
SD v1.4 and SD v2.1 are run in \texttt{float16}, while RealisticVision is run using \texttt{float32}.
For the sake of brevity, we report results for SD v1.4 in the main paper and defer results for the remaining models to \appendixautorefname~\ref{app:results}.

To reproduce memorized samples and conduct mitigation experiments, we first utilize the dataset of 500 memorized prompts identified by~\citet{webster2023extraction} for SD v1.4.
From this set, we construct our RL training split using the 430 prompts for which the corresponding target images are publicly available.
The remaining 70 prompts are strictly held out to evaluate on unseen prompts (see Section~\ref{sec:generalization}). Additionally, for zero-shot out-of-distribution evaluation, we utilize the MemBench dataset of 3000 memorized prompts identified by~\citet{hong2024membench} for SD v1.4. 

For generation, we set the number of steps to $T = 50$, the guidance scale to $g = 7.5$, and the initializations per prompt to $N = 10$. For the target function $\ell(\cdot)$ in $\rads$, we set $\eta = 0.1$ and $\beta = 9.0$ for SD v1.4; see Appendix~\ref{app:target_function_dets} and~\ref{app:target_sensitivity} for further details on these hyperparameters and sensitivity analysis. We report results for 5 train seeds.

To assess applicability beyond the older backbones covered by existing memorization benchmarks, we additionally evaluate on SDXL \citep{podell2024sdxl}, which uses a larger U-Net and dual text encoders while retaining standard classifier-free guidance. Because established SDXL memorization benchmarks with verified prompt-image pairs remain limited, we construct a controlled memorization setting by LoRA fine-tuning SDXL-base on 51 Webster prompt-image pairs. This allows us to test both whether the CFG-based risk signal remains informative and whether $\rads$ mitigates memorization on this architecture. Full details appear in \appendixautorefname~\ref{app:sdxl}.

\textbf{Baselines.}
We compare our method against four inference-time memorization mitigation strategies:~\citet{wen2024detecting, ren2024crossattn, hintersdorf2024nemo, jain2025attraction}.
While some of these works also propose training-time mitigation approaches, we exclude such methods from our comparison, as they require full retraining of the diffusion model.
\emph{Importantly, the $\rads$-trained policy likewise operates entirely at inference-time and does not require retraining the diffusion model.}
\vspace{-0.75em}

\subsection{Implementation of \rads}
\vspace{-0.5em}
To efficiently steer the high-dimensional CLIP text embeddings, we train a custom VAE that compresses the input into a 64-dimensional latent action space.
The policy and critic networks are parameterized as lightweight MLPs trained via SAC, utilizing entropy tuning to ensure stable exploration. For the algorithm pseudocode of $\rads$, please refer to Appendix~\ref{app:pseudocode}. For complete details on the experimental setup and training configurations, please refer to the rest of Appendix~\ref{app:implementation}.
\vspace{-0.5em}

\subsection{Evaluation Metrics}
We measure the effectiveness of $\rads$ using the following three complementary metrics:
\vspace{-0.5em}
\begin{itemize}[noitemsep, leftmargin=*]
    \item \textbf{SSCD (Diversity):} To measure the degree of memorization in generated images, we utilize Self-Supervised Copy Detection (SSCD)~\citep{pizzi2022sscd} to quantify visual similarity in two settings. SSCD has been reported to be one of the strongest metrics for replication~\citep{somepalli2023forgery}. We compute $\text{SSCD}_{\text{target}}$ (greatest pairwise similarity with the target images), $\text{SSCD}_{\text{seeds}}$ (similarity across random seeds for a fixed prompt) to measure replication, and $\text{SSCD}_{\text{prompts}}$ (similarity across prompts for a fixed seed) to detect mode collapse. For the $\text{SSCD}_{\text{target}}$, we evaluate on the 430 prompts from \citet{webster2023extraction} for which the ground-truth training images are publicly available. Lower scores indicate greater diversity. 
    \item \textbf{FID (Quality):} We assess quality via Fréchet Inception Distance (FID)~\citep{heusel2017fid} relative to the COCO validation set~\citep{lin2014microsoft}. Lower FID denotes more natural images. 
    \item \textbf{CLIP (Alignment):} We measure semantic consistency using the CLIP score~\citep{hessel2021clipscore} (cosine similarity between image and text embeddings). Higher scores indicate better alignment.
\end{itemize}
\vspace{-0.5em}

Additionally, we evaluate the computational efficiency of our approach. $\rads$ introduces minimal inference-time overhead, maintaining a generation speed comparable to the fastest, high-fidelity baselines, while \citet{hintersdorf2024nemo} is significantly slower (see Appendix~\ref{app:compute}).

%% file: sec/6_results.tex
\section{Results}
\label{sec:results}
\vspace{-0.5em}

\begin{figure}[h]
    \centering
    \includegraphics[width=0.9\textwidth]{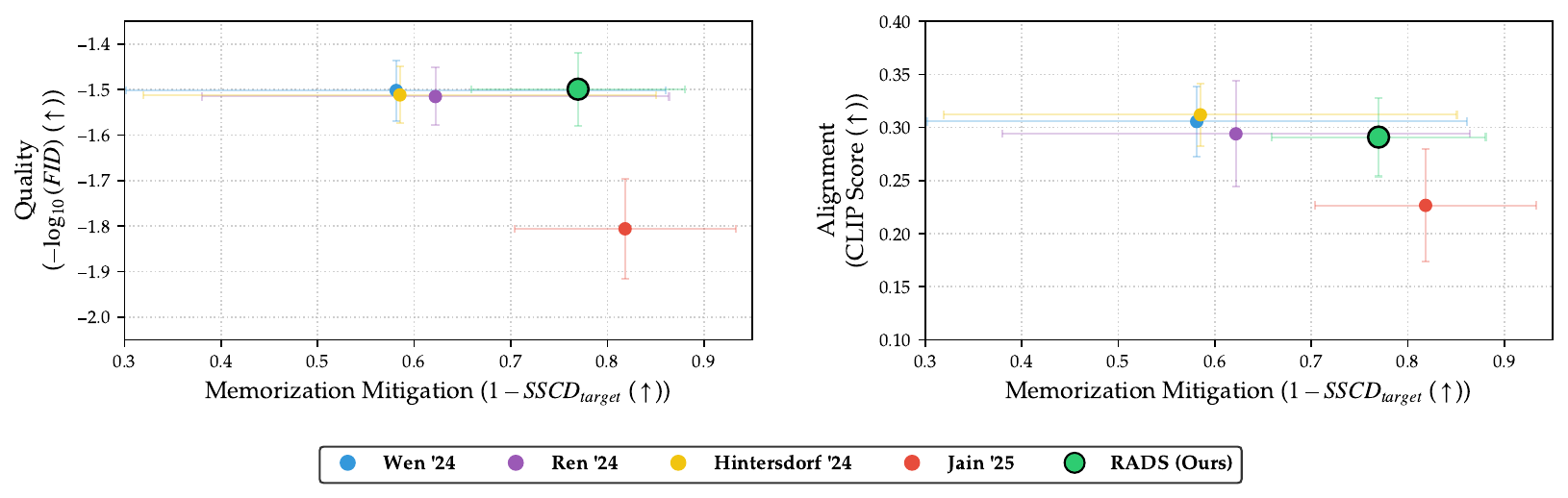}
    \caption{\textbf{In-Distribution Pareto Frontier: Quality and Alignment vs. Memorization Mitigation.} 
    We compare $\rads$ against baselines on the \protect\citet{webster2023extraction} dataset. 
    The left plot shows Quality ($-\log_{10}(\mathrm{FID}) \uparrow$), while the right plot displays Alignment (CLIP $\uparrow$). The x-axis shows the $(1-\sscdtarget) \uparrow$ scores.
    $\rads$ occupies the upper-right region of the frontier, maintaining high utility and alignment while significantly reducing memorization. \textbf{Note:} $\rads$ is trained on the \citeauthor{webster2023extraction} training split; prior baselines are zero-shot. \emph{Thus, we include zero-shot evaluation in Section~\ref{sec:generalization}.}}
    \label{fig:webster_pareto_frontier}
\end{figure}

We evaluate how well $\rads$ mitigates memorization while preserving image quality and prompt alignment. As summarized in \figureautorefname~\ref{fig:webster_pareto_frontier}, $\rads$ achieves a stronger empirical Pareto frontier compared to existing baselines, maintaining comparable CLIP and FID scores to existing methods, while significantly reducing the replication rate.  We use ``Pareto frontier'' to refer to the \emph{tradeoff} between memorization mitigation (SSCD) and image quality/fidelity; CLIP can be inflated by memorization, so small CLIP drops are expected after mitigation. Full quantitative results are provided in Appendix~\ref{app:results}.

\vspace{-0.85em}

\subsection{How diverse are images generated by $\rads$?}
\label{sec:diversity}

\begin{figure*}[!t]
    \small
    \centering

    \begin{subfigure}[t]{0.15\linewidth}
        \centering
        \includegraphics[width=\linewidth]{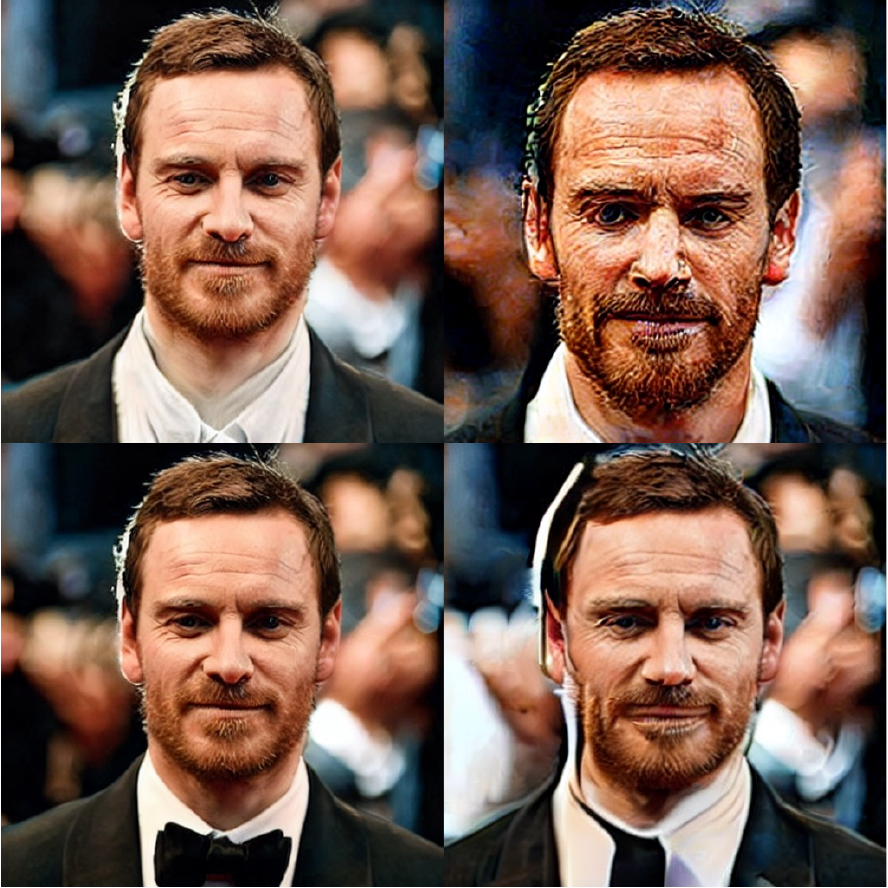}
        \caption{\footnotesize Memorized}
        \label{fig:diversity_memorized}
    \end{subfigure}
    \hfill
    \begin{subfigure}[t]{0.15\linewidth}
        \centering
        \includegraphics[width=\linewidth]{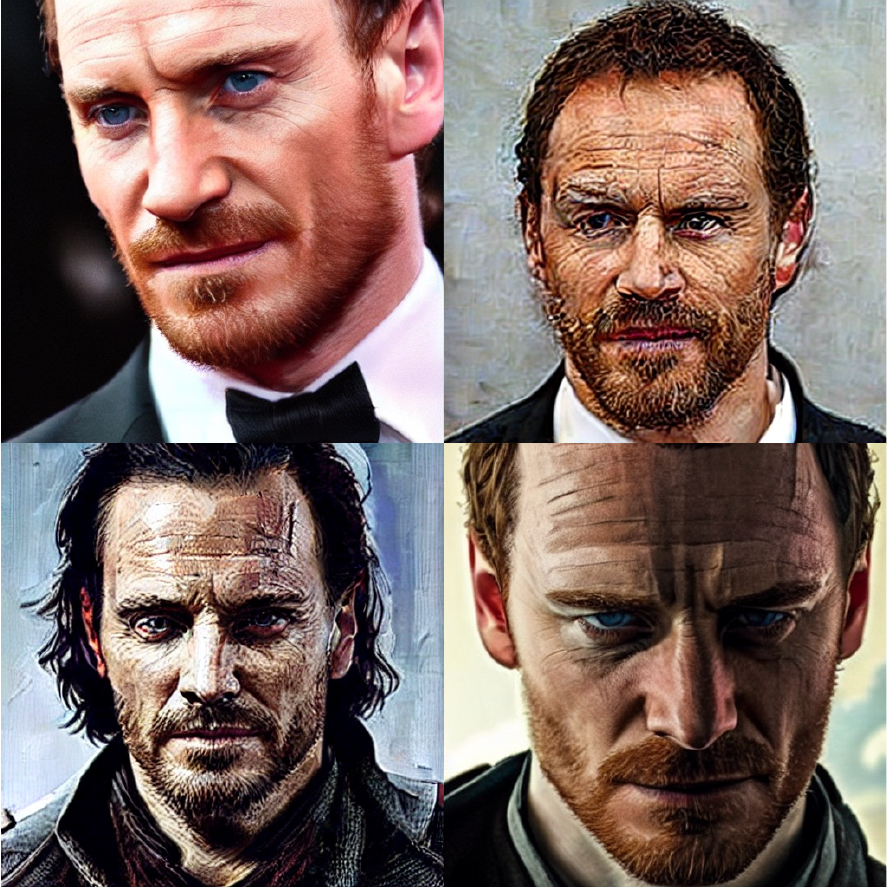}
        \caption{\footnotesize \citeauthor{wen2024detecting}}
        \label{fig:diversity_wen}
    \end{subfigure}
    \hfill
    \begin{subfigure}[t]{0.15\linewidth}
        \centering
        \includegraphics[width=\linewidth]{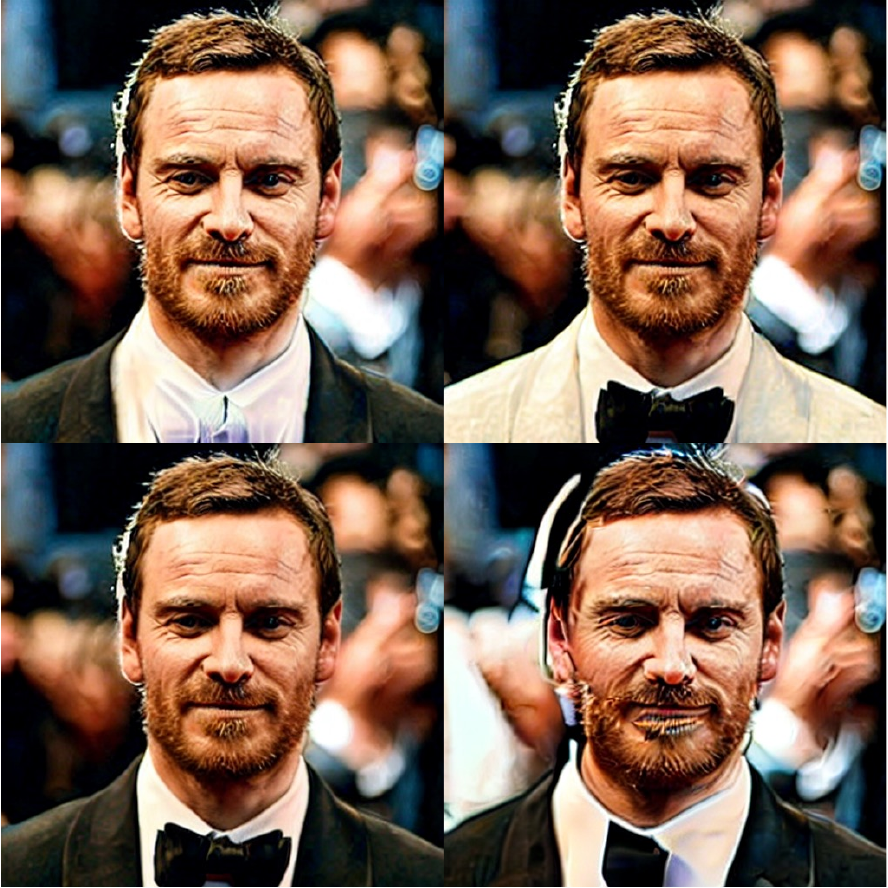}
        \caption{\footnotesize \citeauthor{ren2024crossattn}}
        \label{fig:diversity_ren}
    \end{subfigure}
    \hfill
    \begin{subfigure}[t]{0.15\linewidth}
        \centering
        \includegraphics[width=\linewidth]{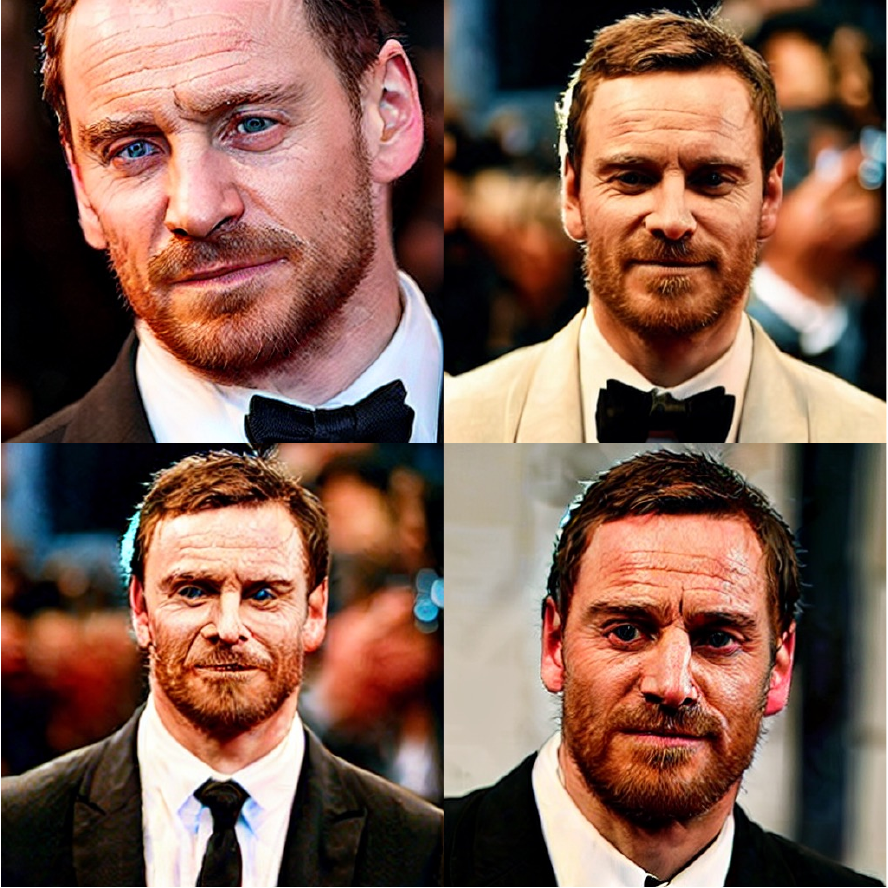}
        \caption{\tiny \citeauthor{hintersdorf2024nemo}}
        \label{fig:diversity_hintersdorf}
    \end{subfigure}
    \hfill
    \begin{subfigure}[t]{0.15\linewidth}
        \centering
        \includegraphics[width=\linewidth]{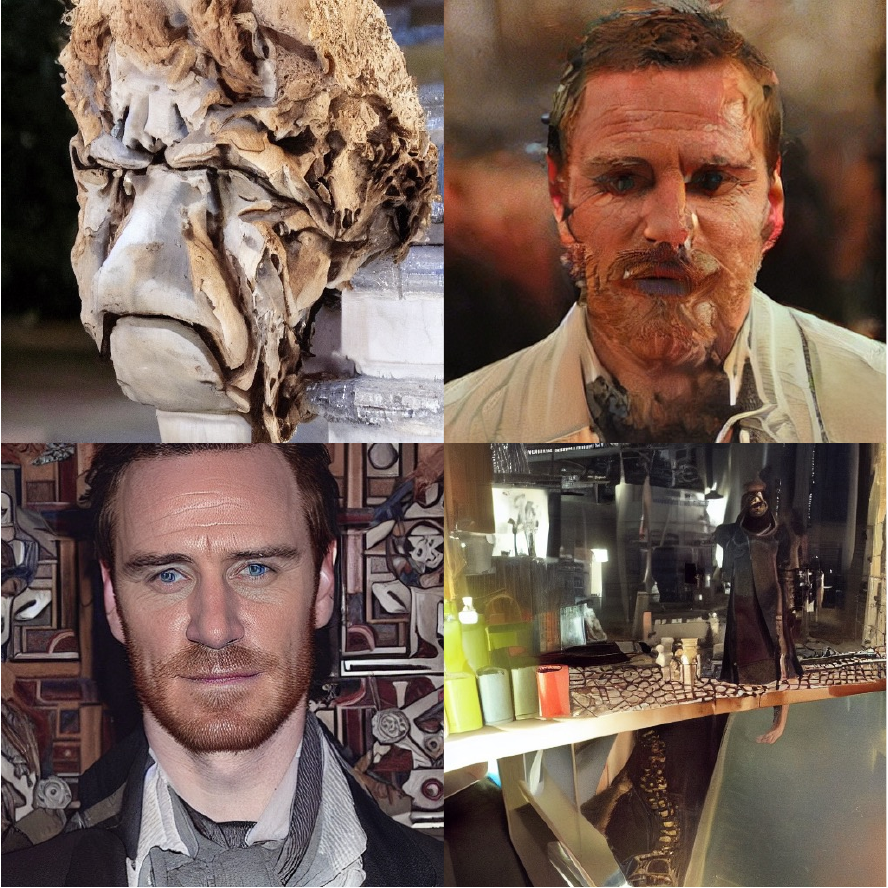}
        \caption{\footnotesize \citeauthor{jain2025attraction}}
        \label{fig:diversity_jain}
    \end{subfigure}
    \hfill
    \begin{subfigure}[t]{0.15\linewidth}
        \centering
        \includegraphics[width=\linewidth]{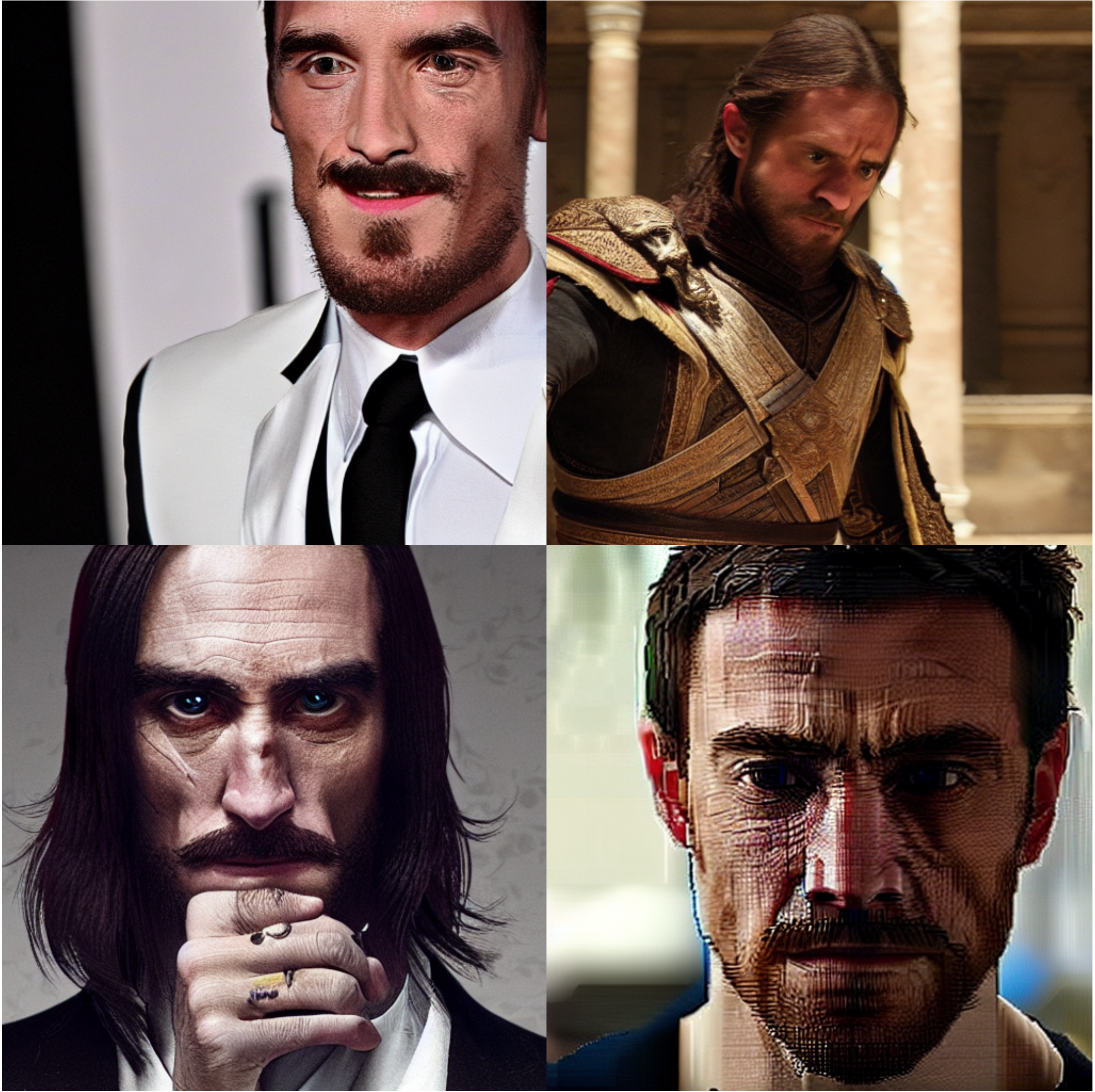}
        \caption{\footnotesize \textbf{$\rads$ (Ours)}}
        \label{fig:diversity_ours}
    \end{subfigure}

    \caption{
        \textbf{$\rads$ achieves mitigation with the highest generation diversity.}
        Generated images for the prompt
        \textit{Michael Fassbender to Star In \textless i\textgreater Assassin's Creed\textless /i\textgreater ~Movie}.
    }
    \label{fig:diversity}
\end{figure*}

A successful mitigation strategy should produce diverse outputs across random seeds for the same prompt, rather than repeatedly reproducing the same memorized instance. \figureautorefname~\ref{fig:diversity} compares generations from four initial latents across prior methods~\citep{wen2024detecting, ren2024crossattn, hintersdorf2024nemo, jain2025attraction} and $\rads$. $\rads$ produces substantially more diverse images across seeds (\figureautorefname~\ref{fig:diversity_ours}), while prior methods either produce similar outputs across seeds and methods (\figureautorefname s~\ref{fig:diversity_wen}--\ref{fig:diversity_hintersdorf}) or degrade image quality (\figureautorefname~\ref{fig:diversity_jain}). This is reflected in \tableautorefname~\ref{tab:results_webster_sdv1_ddim}: excluding \citet{jain2025attraction}, whose high FID reflects severe quality degradation, $\rads$ achieves the lowest $\sscdtarget$ and $\sscdseeds$ scores among high-fidelity methods.

\citet{jain2025attraction} exhibits high $\sscdprompts$ (0.2724 vs. 0.0409 for $\rads$), indicating mode collapse across prompts. Because the method turns guidance on and off, the initial latent can dominate the output, causing generations from the same $\latent_T$ to share similar style across prompts (\figureautorefname~\ref{fig:jain_diversity}).

\begin{figure*}[!t]
    \small
    \centering

    \begin{subfigure}[t]{0.15\linewidth}
        \centering
        \includegraphics[width=\linewidth]{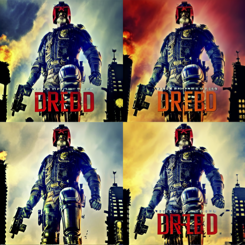}
        \caption{\footnotesize Memorized}
        \label{fig:challenging_memorized}
    \end{subfigure}
    \hfill
    \begin{subfigure}[t]{0.15\linewidth}
        \centering
        \includegraphics[width=\linewidth]{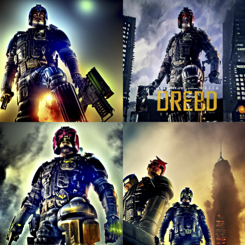}
        \caption{\footnotesize \citeauthor{wen2024detecting}}
        \label{fig:challenging_wen}
    \end{subfigure}
    \hfill
    \begin{subfigure}[t]{0.15\linewidth}
        \centering
        \includegraphics[width=\linewidth]{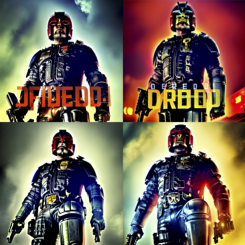}
        \caption{\footnotesize \citeauthor{ren2024crossattn}}
        \label{fig:challenging_ren}
    \end{subfigure}
    \hfill
    \begin{subfigure}[t]{0.15\linewidth}
        \centering
        \includegraphics[width=\linewidth]{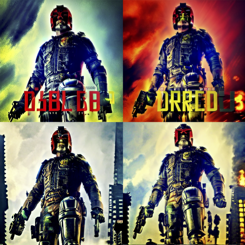}
        \caption{\tiny \citeauthor{hintersdorf2024nemo}}
        \label{fig:challenging_hintersdorf}
    \end{subfigure}
    \hfill
    \begin{subfigure}[t]{0.15\linewidth}
        \centering
        \includegraphics[width=\linewidth]{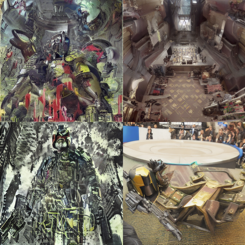}
        \caption{\footnotesize \citeauthor{jain2025attraction}}
        \label{fig:challenging_jain}
    \end{subfigure}
    \hfill
    \begin{subfigure}[t]{0.15\linewidth}
        \centering
        \includegraphics[width=\linewidth]{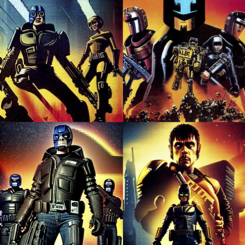}
        \caption{\footnotesize \textbf{$\rads$ (Ours)}}
        \label{fig:challenging_ours}
    \end{subfigure}

    \caption{
        \textbf{$\rads$ achieves mitigation on challenging prompts.}
        Generated images for the prompt
        \textit{\textless i\textgreater Dredd\textless /i$\textgreater$ Sequel Would Require More Star Power According to Producer}.
    }
    \vspace{-0.5em}
    \label{fig:challenging}
\end{figure*}

\vspace{-0.5em}

\subsection{How does $\rads$ perform on challenging, memorization-prone prompts?}
\label{sec:challenging}

We identify a subset of \textit{challenging prompts} from \citet{webster2023extraction}—typically containing specific entities (e.g., \textit{``Dredd''})—where standard mitigation strategies struggle.
\figureautorefname~\ref{fig:challenging} illustrates the failure modes of prior methods on one such prompt.
\citet{ren2024crossattn} and \citet{hintersdorf2024nemo} exhibit \textit{complete failure}, reproducing the memorized training data across all samples (\figureautorefname s~\ref{fig:challenging_ren}, \ref{fig:challenging_hintersdorf}). \citet{wen2024detecting} exhibits \textit{stochastic failure}: while it successfully steers certain random seeds (bottom right, Figure~\ref{fig:challenging_wen}), it fails to mitigate memorization in others, indicating that the method is sensitive to initialization $\latent_T$.
\citet{jain2025attraction} mitigates replication but suffers from severe degradation, rendering the output unrecognizable (\figureautorefname~\ref{fig:challenging_jain}).

In contrast, $\rads$ demonstrates \textit{consistent mitigation} (\figureautorefname~\ref{fig:challenging_ours}).
Regardless of the initial noise $\latent_T$, our method successfully prevents the generation of the memorized instance while preserving the semantic constraints of the prompt (e.g., the action-film style).
This indicates that $\rads$ is robust to initialization even in regions of the latent space with seemingly strong memorization basins.
\vspace{-0.5em}

\subsection{Does $\rads$ preserve quality and prompt alignment?}
\label{sec:quality}
\vspace{-0.75em}

In addition to memorization mitigation, the quality of the generated images is critical.
\figureautorefname~\ref{fig:quality} compares the image quality achieved by prior methods and $\rads$.
Qualitatively, $\rads$ produces high-fidelity images with natural details (\figureautorefname~\ref{fig:quality_ours}).
This is supported quantitatively by the $\fid$ results in \tableautorefname~\ref{tab:results_webster_sdv1_ddim}, where $\rads$ achieves a mean FID of 31.57 ± 5.82.
Given the overlapping standard deviations, this performance is competitive with the strongest baselines \citep{wen2024detecting, ren2024crossattn}, demonstrating that $\rads$ prevents memorization without the quality trade-offs often associated with steering.
Crucially, $\rads$ avoids the severe degradation observed in methods like \citet{jain2025attraction} (63.97 ± 16.17) and, in some cases, improves upon the unmitigated baseline.


In the \citet{webster2023extraction} dataset, $\rads$ achieves a CLIP score of 0.2909 ± 0.0370, which is numerically close to the unmitigated baseline (0.3129 ± 0.0279). While a marginal shift in the mean is observed, the high degree of variance across both methods suggests that $\rads$ preserves the core semantic alignment of the original model while successfully steering away from memorization basins. On the other hand, mitigation methods such as \citet{jain2025attraction} (0.2266 ± 0.0532) exhibit a significant degradation in CLIP alignment.

We argue that the maximal CLIP scores observed in unmitigated baselines do not reflect superior semantic understanding, but rather \emph{memorization bias}, where the model achieves high alignment scores primarily by reproducing the exact training data it has overfitted to. Consequently, any effective mitigation policy that steers away from the memorization manifold necessarily diverges from this overfitted maximum. The marginal reduction in CLIP scores for $\rads$ represents the removal of this bias, preserving generalized semantic intent rather than instance-specific replication. Thus, CLIP should be interpreted alongside SSCD and FID in memorization-prone settings.

Qualitatively, \figureautorefname~\ref{fig:alignment} confirms that $\rads$ can succeed where prior methods may fail.
While some baselines miss key semantic elements (e.g., \figureautorefname~\ref{fig:alignment_wen}; \textit{`Dinner With Friends'} is not reflected) or produce severe artifacts (\figureautorefname~\ref{fig:alignment_jain}), $\rads$ generates a high-quality image that clearly reflects the prompt's intent (\figureautorefname~\ref{fig:alignment_ours}).

\vspace{-0.5em}

\subsection{How well does $\rads$ generalize to unseen prompts and extend to newer architectures?}
\label{sec:generalization}

\begin{figure}[htbp]
    \vspace{-0.5em}
    \centering
    \includegraphics[width=0.9\textwidth]{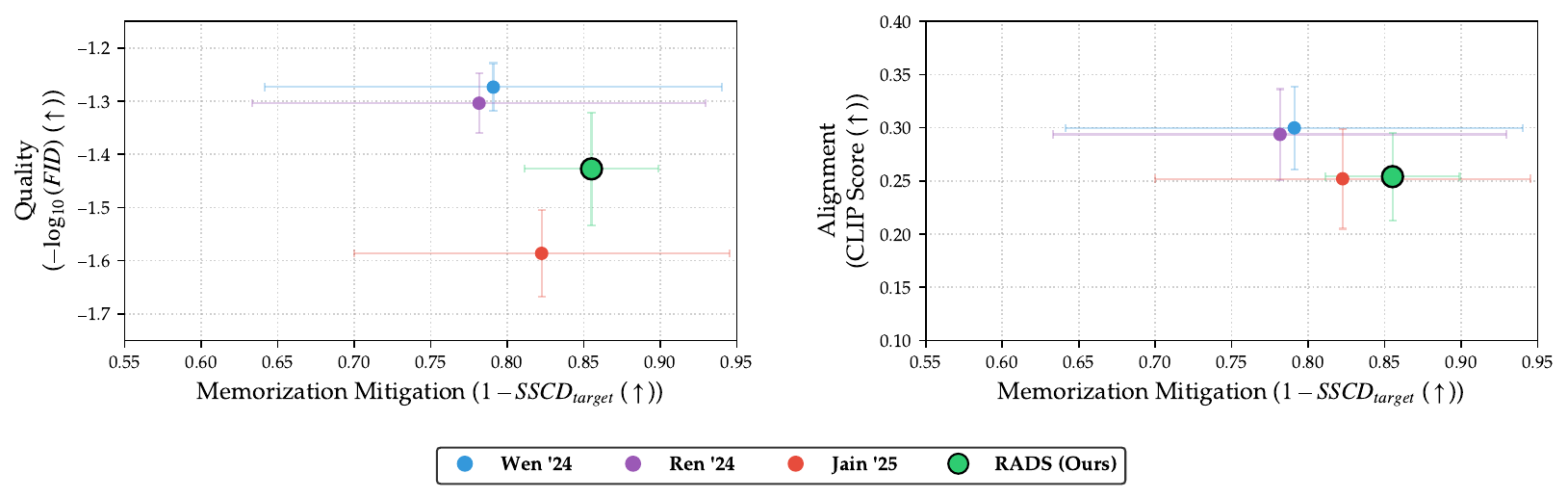}
    \caption{\textbf{Zero-Shot Generalization.} 
    We compare $\rads$ in a zero-shot evaluation against inference-time mitigation baselines on the MemBench dataset \citep{hong2024membench}. 
    The left plot shows Quality ($-\log_{10}(\mathrm{FID}) \uparrow$), while the right plot displays Alignment (CLIP Score $\uparrow$). The x-axis shows the $(1-\sscdtarget) \uparrow$ scores.
    $\rads$ achieves the strongest memorization mitigation, with better quality than \protect\citet{jain2025attraction} at comparable alignment; relative to \protect\citet{wen2024detecting} and \protect\citet{ren2024crossattn}, this comes at a moderate cost in quality and alignment. \textbf{Note:} The \protect\citet{hintersdorf2024nemo} baseline is omitted due to its computational infeasibility.}
    \label{fig:zero_shot_generalization}
\end{figure}

To evaluate if our steering policy—trained on a limited set of 430 prompts—learns a robust mitigation strategy, we assess zero-shot generalization on two unseen datasets. First, on the \citet{webster2023extraction} validation set (\tableautorefname~\ref{tab:results_webster_validation}), $\rads$ reduces $\sscdseeds$ from 0.5730 (unmitigated) to 0.1356, effectively halving the replication rate of the strongest high-fidelity baseline, \citet{wen2024detecting} (0.2690).

We further scrutinize this generalization on the 3,000 out-of-distribution prompts in MemBench (\tableautorefname~\ref{tab:results_membench}). Figure~\ref{fig:zero_shot_generalization} illustrates the Pareto frontier in the MemBench dataset. Once again, $\rads$ achieves the strongest memorization mitigation, with a moderate trade-off in quality and alignment relative to the highest-fidelity baselines \citep{wen2024detecting, ren2024crossattn}.

\emph{Remarkably, despite the limited training data, $\rads$ achieves the lowest target similarity ($\sscdtarget \approx 0.145$) among all evaluated methods.} This outperforms \citet{jain2025attraction} ($\sscdtarget \approx 0.178$), confirming that $\rads$ actively steers the generation away from memorization. In contrast, \citet{jain2025attraction} achieves low variance primarily through mode collapse to a state that remains proximal to the training data. $\rads$ maintains robust mitigation with competitive image fidelity ($\fid \approx 26.75$), avoiding the quality degradation observed in \citet{jain2025attraction}.
\vspace{-0.5em}
\paragraph{Extension to SDXL.} Beyond zero-shot evaluation on unseen prompts, we separately train and evaluate $\rads$ on a finetuned SDXL checkpoint. $\rads$ reduces $\sscdtarget$ from 0.4875 to 0.3562 and $\sscdseeds$ from 0.7056 to 0.3844, while CLIP decreases from 0.3066 to 0.2939. This supports applicability to a larger architecture with dual text encoders, within the scope of our induced-memorization setting (Appendix~\ref{app:sdxl}).
\vspace{-0.5em}
\subsection{How well does $\rads$ preserve generation quality on standard prompts?}
\label{sec:standard_prompt_preservation}
\vspace{-0.5em}

A memorization-mitigation method should also preserve generation quality and alignment on ordinary prompts that do not elicit memorization. To assess this, we evaluate $\rads$ on 1,000 randomly sampled COCO captions \citep{lin2014microsoft}, rather than prompts selected for their ability to reproduce training images. This provides a test of behavior on general-purpose prompts, although we do not assume that every caption is verified to be non-memorizing. $\rads$ largely preserves generation quality and alignment: CLIP changes from 0.3118 to 0.3077, while FID changes from 5.45 to 5.26 and seed diversity increases. Full results and evaluation details appear in
Appendix~\ref{app:coco_preservation}.
\vspace{-0.5em}
\subsection{Are all components of $\rads$ necessary?}
\label{sec:ablation}
\vspace{-0.5em}
We compare $\rads$ against four ablations: removing the reachability constraint, replacing the learned actor with classifier guidance using a state-only safety value, replacing the value-function target with a standard discounted-sum Bellman target, and applying only VAE reconstruction with $u_t=0$ at every step. As shown in Table~\ref{tab:rads_ablation}, removing the constraint substantially worsens memorization, increasing $\sscdtarget$ from 0.230 to 0.500, showing that alignment rewards alone are insufficient. VAE-only reconstruction reduces $\sscdtarget$ from 0.636 to 0.572, showing that reconstruction alone cannot explain the mitigation in memorization. Classifier guidance improves over the unconstrained variant but remains worse than full $\rads$. The standard Bellman target achieves low target similarity, but substantially worsens FID and CLIP, suggesting that it reduces memorization through degraded generations rather than a useful quality-diversity tradeoff; its training is also substantially less stable (Figure~\ref{fig:sum_ablation_training_curves}). Additional sensitivity experiments show that $\rads$ is robust to moderate variation in the target-function shaping scale $\eta$; see Appendix~\ref{app:ablations} and Appendix~\ref{app:target_sensitivity} for details. Our conclusion is that each component of $\rads$ contributes to effective memorization mitigation.

%% file: sec/7_discussion_and_limitations.tex
\vspace{-0.5em}

\section{Discussion and Limitations}
\label{sec:discussion}
\vspace{-0.5em}

Our results support the view that memorization is best treated as a trajectory-level failure rather than only a property of the final image. By applying state-dependent control throughout denoising, $\rads$ steers generations away from memorization-prone regions while preserving image fidelity and prompt alignment. More broadly, reachability-constrained steering may provide a general framework for controlling generative failures when an appropriate safety target can be specified.

While $\rads$ is effective in mitigating memorization, several limitations motivate future research:

\textbf{Semantic Drift via Data Scarcity.}
$\rads$ is trained on a limited set of verified memorized prompts, which can cause the policy to overfit to specific semantic tokens rather than memorization triggers. As a result, some out-of-distribution prompts can exhibit \textit{semantic drift} after steering (Appendix~\ref{app:failure_cases}). Future work should incorporate larger mixtures of memorized prompts and safe, non-memorized prompts to better separate memorization risk from general semantic content.

\textbf{Training and Calibration Requirements.}
Unlike zero-shot heuristic baselines, $\rads$ requires offline
training of an auxiliary controller (approximately 15 hours
on a single A100 GPU for SD v1.4). Applying it to a new backbone
requires training the policy and safety critic and calibrating
the CFG-norm threshold. The caption VAE can be reused across backbones sharing the same text-encoder embedding space.

\textbf{Benchmark Coverage for Modern Models.}
Our SDXL evaluation provides a controlled LoRA-induced memorization study,
rather than real training samples generated from the original pretrained
model. Broader evaluation therefore requires verified memorization
benchmarks for modern diffusion architectures, such as SDXL~\citep{podell2024sdxl} and FLUX~\citep{labs2025flux1kontextflowmatching}.
\vspace{-0.5em}

%% file: sec/8_acknowledgements.tex
\begin{ack}
We gratefully acknowledge research support from Open Philanthropy, the NSF CAREER program (2240163), and the Stanford University School of Engineering. We are also grateful to the Stanford Institute for Human-Centered Artificial Intelligence and Google Cloud for providing computational resources. Additionally, this work was supported by Institute of Information \& communications Technology Planning \& Evaluation (IITP) under 6G Cloud Research and Education Open Hub grant (IITP-2025-RS-2024-00428780) and by the National Research Foundation of Korea (NRF) grant (RS-2024-00453301, RS-2025-00517159) funded by the Korea government (MSIT).
\end{ack}
\clearpage

%% file: sec/appendix.tex
\section*{\Large Appendix}
\noindent
This appendix is organized as follows:
\begin{itemize}[left=0pt]
    \item \textbf{Implementation details} are provided in Appendix~\ref{app:implementation}.
    \item \textbf{Complete quantitative results} are provided in Appendix~\ref{app:results}.
    \item \textbf{Safety Critic Reliability Analysis} is provided in Appendix~\ref{app:reliability}.
    \item \textbf{Ablation studies and hyperparameter sensitivity experiments} are provided in Appendix~\ref{app:ablations}.
    \item \textbf{Classifier-free guidance analysis} is provided in Appendix~\ref{app:safety_analysis}.
    \item \textbf{Computational cost} is reported in Appendix~\ref{app:compute}.
    \item \textbf{Additional qualitative results and failure cases} are shown in Appendix~\ref{app:qualitative}.
\end{itemize}

\section{Implementation Details}
\label{app:implementation}

\subsection{Algorithm Pseudocode}
\label{app:pseudocode}

The $\rads$ training and inference-time stages are provided in Algorithms~\ref{alg:rads_training} and~\ref{alg:rads_inference}. 

\begin{algorithm}[h]
\caption{\rads \text{ Training}}
\label{alg:rads_training}
\begin{algorithmic}[1]
\REQUIRE Pretrained diffusion model $f_{\mathrm{DM}}$, frozen text encoder, VAE $(\enc,\dec)$, prompts $\mathcal{C}$, horizon $T$, safety target $\ell$, replay buffer $\mathcal{D}$
\REQUIRE Policy $\pi_\phi$, twin task critics $Q^{\mathrm{task}}_{\omega_i}$, safety critic $Q^{\mathrm{safe}}_\psi$, corresponding target critics, Lagrange multiplier $\lambda$
\FOR{each training iteration}
    \STATE Sample $c\sim\mathcal{C}$ and $\latent_T\sim\mathcal{N}(0,I)$
    \STATE Compute $\promptemb$ and $\mathbf{z}_c=\enc(\promptemb)$
    \FOR{$t=0,\ldots,T-1$}
        \STATE Form $\state_t=(\latent_{T-t},T-t)$ and sample $\control_t\sim\pi_\phi(\cdot\mid\state_t)$
        \STATE Decode $\promptembt=\dec(\mathbf{z}_c+\control_t)$
        \STATE Step diffusion dynamics to obtain $\state_{t+1}$
        \STATE Compute $\ell_t=\ell(\state_t,\control_t)$ and set $d_t=\mathbf{1}\{t=T-1\}$
        \STATE Set $r_t=0$; on the terminal transition, decode $\image$ and set
        $r_t=\mathrm{CLIP}_{\mathrm{im}}(\image)\cdot\mathrm{CLIP}_{\mathrm{text}}(c)$
        \STATE Store $(\state_t,\control_t,\state_{t+1},\ell_t,r_t,d_t,c)$ in $\mathcal{D}$
    \ENDFOR
    \FOR{each gradient update}
        \STATE Sample a minibatch from $\mathcal{D}$
        \STATE For nonterminal transitions ($d_t=0$), sample
        $\control'_{t+1}\sim\pi_\phi(\cdot\mid\state_{t+1})$
        \STATE Compute $\widehat Q_t^{\mathrm{safe}}$ using Eq.~\ref{eqn:q_safe_target}
        \STATE Update the safety critic by minimizing
        $(Q_\psi^{\mathrm{safe}}(\state_t,\control_t)-\widehat Q_t^{\mathrm{safe}})^2$
        \STATE Compute the task target:
        \[
y_t^{\mathrm{task}} =
\begin{cases}
r_t+\gamma\left[
\min_{i=1,2}Q_{\omega'_i}^{\mathrm{task}}
(\state_{t+1},\control'_{t+1})
-\alpha\log\pi_\phi(\control'_{t+1}\mid\state_{t+1})
\right],
& d_t=0,\\
r_t,
& d_t=1.
\end{cases}
\]
        \STATE Update each task critic by minimizing
        $(Q_{\omega_i}^{\mathrm{task}}(\state_t,\control_t)-y_t^{\mathrm{task}})^2$
        \STATE Update the policy by maximizing:
        \[
        J_\pi(\phi)=
        \mathbb{E}_{\state_t\sim\mathcal{D},\control_t\sim\pi_\phi}
        \left[
        Q_\omega^{\mathrm{task}}(\state_t,\control_t)
        +\alpha\mathcal{H}(\pi_\phi(\cdot\mid\state_t))
        +\lambda Q_\psi^{\mathrm{safe}}(\state_t,\control_t)
        \right]
        \]
        \STATE Update $\lambda$ to enforce
        $\mathbb{E}[Q_\psi^{\mathrm{safe}}(\state_t,\control_t)]\ge\delta$
        \STATE Update target networks via Polyak averaging
    \ENDFOR
\ENDFOR
\STATE \textbf{return} Policy $\pi_\phi$ and safety critic $Q_\psi^{\mathrm{safe}}$
\end{algorithmic}
\end{algorithm}

\begin{algorithm}[t]
\caption{\rads \text{ Inference-Time Steering}}
\label{alg:rads_inference}
\begin{algorithmic}[1]
\REQUIRE Pretrained diffusion model $f_{\mathrm{DM}}$, trained VAE $(\enc,\dec)$, trained policy $\pi_\phi$, prompt $c$, initial latent $\latent_T\sim\mathcal{N}(0,I)$, horizon $T$
\STATE Compute base caption embedding $\promptemb$ and latent caption representation $\mathbf{z}_c=\enc(e_c)$
\FOR{$t=0,\ldots,T-1$}
    \STATE Form state $\state_t=(\latent_{T-t},T-t)$
    \STATE Sample or take mean steering action $\control_t\sim\pi_\phi(\cdot\mid \state_t)$
    \STATE Decode steered caption embedding: $ \promptembt=\dec(\mathbf{z}_c+\control_t)$
    \STATE Apply one denoising step with steered conditioning: $
    \latent_{T-(t+1)}=f_{\mathrm{DM}}(\latent_{T-t},\promptembt,T-t)$
\ENDFOR
\STATE Decode final latent $\latent_0$ into image $\image$
\STATE \textbf{return} Mitigated image $\image$
\end{algorithmic}
\end{algorithm}

\subsection{CLIP Embedding Variational Autoencoder}
\label{app:vae}

To enable efficient steering in the high-dimensional caption embedding space, we employ a Variational Autoencoder (VAE) that compresses CLIP text embeddings into a manageable 64-dimensional latent action space. The architecture and training configurations are detailed as follows.

\paragraph{Architecture.} 
The VAE utilizes a transformer-based architecture for both the encoder and decoder components.
\begin{itemize}[left=0pt]
    \item \textbf{Encoder:} The encoder consists of $N=4$ transformer layers with $H=8$ attention heads and a hidden dimension of $D=768$. It processes input CLIP embeddings of shape $77 \times 768$ after adding a learnable positional embedding. The output is passed through a LayerNorm bottleneck before being projected into the latent space.
    \item \textbf{Decoder:} The decoder projects the latent vector back to the $77 \times 768$ sequence shape and processes it through another $N=4$ transformer layers to reconstruct the embedding.
    \item \textbf{Regularization:} A dropout rate of $0.1$ is applied across all layers.
\end{itemize}

\paragraph{Training Protocol and Dataset Mix.} 
The VAE is trained to learn a robust and generalizable representation of the CLIP embedding manifold using the following setup:
\begin{itemize}[leftmargin=*]
    \item \textbf{Loss Function:} The model is trained to minimize a multi-objective loss function that combines semantic and structural reconstruction metrics with a Kullback–Leibler (KL) divergence penalty:
\begin{equation}
\mathcal{L} = \mathcal{L}_{\text{cos}} + 0.1 \cdot \mathcal{L}_{\text{MSE}} + 2\times 10^{-3} \cdot \mathcal{L}_{\text{KLD}}
\end{equation}
where $\mathcal{L}_{\text{cos}}$ represents the cosine embedding loss (1 - cosine similarity) to preserve the semantic orientation of the CLIP vectors, and $\mathcal{L}_{\text{MSE}}$ provides structural grounding. 
    \item \textbf{Optimization:} We use the Adam optimizer with a learning rate of $10^{-4}$ and a batch size of 64 for 45,000 steps.
    \item \textbf{Dataset Mix:} The model is trained on the Conceptual Captions dataset \citep{sharma2018conceptual} to ensure internet-scale robustness across the text manifold. To prevent overfitting and ensure high fidelity for target contexts, this data is mixed with the specific 430 captions from our experimental training dataset. Periodic evaluation of reconstruction quality (MSE and cosine similarity) is performed specifically on these captions during training.
\end{itemize}

\subsection{Constrained Soft Actor-Critic}
\label{app:csac_impl_dets}

To solve the constrained MDP, we employ a constrained Soft Actor-Critic (SAC) algorithm with Lagrangian relaxation to enforce the reachability-based safety constraint. The implementation details are as follows.

\paragraph{Network Architectures.} 
All agent networks are implemented as Multi-Layer Perceptrons (MLPs) with the following specifications:
\begin{itemize}[leftmargin=*]
    \item \textbf{Actor Policy ($\pi_\phi$):} A stochastic policy that outputs a Gaussian distribution $(\mu, \sigma)$ over the $d=64$ dimensional latent action space. The network consists of 3 hidden layers with 256 units each and ReLU activations. Actions are sampled using the reparameterization trick and squashed via a \texttt{tanh} layer to the range $[-1, 1]$.
    \item \textbf{Twin Task Critics ($Q_{\omega_1}^{\text{task}}$, $Q_{\omega_2}^{\text{task}}$):} We utilize twin $Q$-networks to mitigate overestimation bias. Each critic comprises 3 hidden layers (256 units each) and takes the concatenated state (diffusion latent and caption embedding) and action as input.
    \item \textbf{Safety Critic ($Q_\psi^{\text{safe}}$):} The safety critic shares the same architecture as the task critics but is trained to estimate the future reachability value with respect to the memorization failure set. We set $\gamma = 0.99$ from Equation~\ref{eqn:q_safe_target}.
\end{itemize}

\paragraph{Hyperparameters and Training Protocol.} 
The policy is trained to optimize the trade-off between semantic reward and the safety constraint using the following hyperparameters and training protocol:
\begin{itemize}[leftmargin=*]
    \item \textbf{Optimization:} We use the Adam optimizer for all networks. The learning rate is set to $10^{-4}$ for the actor and $3 \times 10^{-5}$ for the twin task critics, the safety critic, and the entropy temperature $\alpha$.
    \item \textbf{Constraint Handling:} The Lagrangian multiplier $\lambda$ is updated via dual gradient descent with a learning rate of $3 \times 10^{-5}$. We apply a \texttt{tanh} $L^2$ threshold of 9.0 for the safety reward transformation to separate the $L^2$-norms of the classifier guidance vector based on memorization ($L^2$-norm $>$ 9.0 denotes memorization, see Appendix~\ref{app:target_function_dets}).
    \item \textbf{Training Protocol:} We use a simulation batch size of 32 trajectories. The policy is trained for 90 epochs. We set the number of network updates to 1. Lastly, we perform evaluation every 5 epochs and track the epoch with the highest sum of the reward function and target function $r(\state_T)+\ell(\state_T)$.
\end{itemize}

\subsection{Additional Model-Specific Hyperparameters}
\label{app:model_specific_hparams}

For SD v2.1 experiments, we use the same denoising horizon ($T=50$), guidance scale ($g=7.5$), and DDIM scheduler as the SD v1.4 setting. Since SD v2.1 uses a different text-encoder embedding space, we train a separate caption-embedding VAE. We keep the latent action dimension fixed at $d=64$ and train the SAC policy for 90 epochs with simulation batch size 16, evaluation interval 5, and one update per environment step. For SD v2.1, we use actor learning rate $1\times 10^{-4}$, critic learning rate $3\times 10^{-5}$, entropy-temperature learning rate $1\times 10^{-5}$, Lagrange multiplier learning rate $1\times 10^{-4}$, action regularization coefficient $0$, and CFG-norm threshold $\beta=11$. 

For SD v2.1, we use $\beta=11.0$ rather than the standalone classification-optimal threshold near $\beta \approx 8.45$. While the latter gives higher static memorization/non-memorization classification accuracy, $\beta=11.0$ yields a more stable safety-critic target and better downstream policy learning. Since $\beta$ serves as a target-shaping parameter rather than a final classifier, we select it based on training stability and steering performance.

\subsection{Target Function and Threshold Selection}
\label{app:target_function_dets}

In Equation~\ref{eqn:target_fn}, we define our instantiation of the target function $\ell$, which is parameterized by the safety threshold $\beta$ and the scaling factor $\eta$. For SD v1.4, we set $\eta = 0.1$ and $\beta = 9.0$.

The threshold $\beta$ is specifically chosen to facilitate the separation between the guidance magnitudes of novel generations and those associated with memorization. We empirically justify this value by analyzing the $L^2$-norms of the classifier-free guidance vectors across the training set. Our analysis indicates that an $L^2$-norm threshold of approximately $8.61$ at the final timestep achieves a classification accuracy of $89.77\%$ for identifying memorized samples. By approximating $\beta = 9.0$ (corresponding to an accuracy of $89.5\%$), we establish a robust boundary that effectively flags memorization while minimizing false positives during steering. 

We set $\delta=0$ and impose the learned safety constraint
in expectation; this does not provide a per-prompt safety guarantee.

\subsection{Target Function Hyperparameter Sensitivity}
\label{app:target_sensitivity}

The target function is central to $\rads$, since it defines the surrogate failure set.
In our implementation, failure corresponds to the classifier-free guidance norm exceeding
a threshold $\beta$. We set $\beta=9.0$ using an empirical analysis of memorized and
non-memorized trajectories, which yields 89.5\% classification accuracy in distinguishing memorized from non-memorized states.

We also evaluate sensitivity to the shaping scale $\eta$ in the target function
(Equation~\ref{eqn:target_fn}). Table~\ref{tab:eta_sensitivity} reports results on
SD v1.4 for $\eta \in \{0.05,0.1,0.2\}$, with $\eta=0.1$ used in the main experiments.
Across this range, $\rads$ maintains similar SSCD, FID, and CLIP scores, suggesting that
the method is robust to moderate variation in $\eta$. While the precise target affects
performance, the reachability-style objective reduces reliance on pointwise thresholding
by reasoning about future memorization risk.

\begin{table}[h]
    \centering
    \small
    \caption{
        \textbf{Sensitivity to target-function shaping scale $\eta$.}
        Results are reported on SD v1.4. The main experiments use $\eta=0.1$.
    }
    \label{tab:eta_sensitivity}
    \resizebox{\linewidth}{!}{
    \begin{tabular}{lccccc}
        \toprule
        Method & $\sscdtarget \downarrow$ & $\sscdseeds \downarrow$ & $\sscdprompts \downarrow$ & FID $\downarrow$ & CLIP $\uparrow$ \\
        \midrule
        $\rads$ ($\eta=0.05$) & 0.2493 ± 0.1243 & 0.1731 ± 0.1311 & 0.0427 ± 0.0238 & 30.2442 ± 6.8452 & 0.2930 ± 0.0385 \\
        $\rads$ ($\eta=0.10$) & 0.2303 ± 0.1110 & 0.1553 ± 0.1099 & 0.0409 ± 0.0227 & 31.5678 ± 5.8163 & 0.2909 ± 0.0370 \\
        $\rads$ ($\eta=0.20$) & 0.2310 ± 0.1116 & 0.1595 ± 0.1099 & 0.0392 ± 0.0213 & 30.4647 ± 4.9243 & 0.2939 ± 0.0359 \\
        \bottomrule
    \end{tabular}
    }
\end{table}

\section{Complete Quantitative Results}
\label{app:results}

We provide complete quantitative results across datasets, samplers, and diffusion backbones. 
We report \ensuremath{\sscdtarget}, \ensuremath{\sscdseeds}, and \ensuremath{\sscdprompts} to measure memorization and diversity, \ensuremath{\fid} to measure image quality, and \ensuremath{\clip} to measure prompt alignment. Lower values are better for \ensuremath{\sscdtarget}, \ensuremath{\sscdseeds}, \ensuremath{\sscdprompts}, and \ensuremath{\fid}; higher values are better for \ensuremath{\clip}. Best results are shown in bold. All results are provided as mean ± standard deviation for 5 training seeds and 10 evaluation seeds. 
\vspace{-1em}
\subsection{Stable Diffusion v1.4 on the Webster Dataset}
\label{app:results_webster_sdv1_ddim}

Table~\ref{tab:results_webster_sdv1_ddim} reports results on the \citet{webster2023extraction} dataset using Stable Diffusion v1.4 with DDIM sampling. This is the primary in-distribution setting used in the main paper. While \citet{jain2025attraction} achieves the lowest \ensuremath{\sscdtarget} and \ensuremath{\sscdseeds}, it substantially degrades image quality and prompt alignment. In contrast, \ensuremath{\rads} achieves the best \ensuremath{\sscdprompts} and \ensuremath{\fid}, while maintaining competitive \ensuremath{\clip}.
\vspace{-1.25em}

\begin{table}[H]
    \centering
    \caption{\textbf{\citet{webster2023extraction} Dataset with Stable Diffusion v1.4 and DDIM Sampling.} The dataset contains 500 memorized prompts. \ensuremath{\sscdtarget} is computed only on the 430 prompts with publicly available target images.}
    \label{tab:results_webster_sdv1_ddim}
    \small
    \setlength{\tabcolsep}{4pt}
    \begin{tabular}{lccccc}
        \toprule
        Method & \ensuremath{\sscdtarget} (↓) & \ensuremath{\sscdseeds} (↓) & \ensuremath{\sscdprompts} (↓) & \ensuremath{\fid} (↓) & \ensuremath{\clip} (↑) \\
        \midrule
        No mitigation 
        & 0.6364 ± 0.2626 
        & 0.4910 ± 0.2983 
        & 0.0956 ± 0.1031 
        & 42.1392 ± 5.9058 
        & \textbf{0.3129 ± 0.0279} \\
        \citeauthor{wen2024detecting} 
        & 0.4187 ± 0.2796 
        & 0.2132 ± 0.1798 
        & 0.0613 ± 0.0686 
        & 31.7794 ± 4.8709 
        & 0.3056 ± 0.0329 \\
        \citeauthor{ren2024crossattn} 
        & 0.3781 ± 0.2420 
        & 0.2332 ± 0.1876 
        & 0.0695 ± 0.0744 
        & 32.7601 ± 4.7818
        & 0.2940 ± 0.0500 \\
        \citeauthor{hintersdorf2024nemo} 
        & 0.4149 ± 0.2657 
        & 0.2534 ± 0.2042 
        & 0.0573 ± 0.0584 
        & 32.4948 ± 4.6363 
        & 0.3119 ± 0.0296 \\
        \citeauthor{jain2025attraction} 
        & \textbf{0.1816 ± 0.1144} 
        & \textbf{0.0620 ± 0.0608} 
        & 0.2724 ± 0.1352 
        & 63.9720 ± 16.1660
        & 0.2266 ± 0.0532 \\
        \textbf{\ensuremath{\rads} (Ours)} 
        & 0.2303 ± 0.1110 
        & 0.1553 ± 0.1099 
        & \textbf{0.0409 ± 0.0227} 
        & \textbf{31.5678 ± 5.8163} 
        & 0.2909 ± 0.0370 \\
        \bottomrule
    \end{tabular}
\end{table}
\vspace{-1.4em}

\subsection{Held-Out Webster Prompts}
\label{app:results_webster_validation}

Table~\ref{tab:results_webster_validation} evaluates generalization to the 70 held-out prompts from \citet{webster2023extraction}. These prompts do not have publicly available target images, so \ensuremath{\sscdtarget} cannot be computed. \ensuremath{\rads} substantially reduces \ensuremath{\sscdseeds} compared to the unmitigated baseline and remains competitive with the strongest high-fidelity baselines.

\begin{table}[H]
    \centering
    \caption{\textbf{Held-Out \citet{webster2023extraction} Validation Prompts with Stable Diffusion v1.4 and DDIM Sampling.} The validation set contains 70 unseen prompts from the original 500 memorized prompts. }
    \label{tab:results_webster_validation}
    \small
    \setlength{\tabcolsep}{5pt}
    \begin{tabular}{lcccc}
        \toprule
        Method & \ensuremath{\sscdseeds} (↓) & \ensuremath{\sscdprompts} (↓) & \ensuremath{\fid} (↓) & \ensuremath{\clip} (↑) \\
        \midrule
        No mitigation 
        & 0.5730 ± 0.2934 
        & 0.0953 ± 0.0831 
        & 54.0369 ± 2.2041 
        & \textbf{0.3220 ± 0.0305} \\
        \citeauthor{wen2024detecting} 
        & 0.2690 ± 0.2186 
        & \textbf{0.0618 ± 0.0510} 
        & \textbf{46.2274 ± 3.8182} 
        & 0.3127 ± 0.0345 \\
        \citeauthor{ren2024crossattn} 
        & 0.2842 ± 0.1988 
        & 0.0655 ± 0.0498 
        & 49.8954 ± 2.9605 
        & 0.3120 ± 0.0320 \\
        \citeauthor{hintersdorf2024nemo} 
        & 0.2537 ± 0.1537 
        & 0.0764 ± 0.0612 
        & 49.5971 ± 4.4862 
        & 0.3118 ± 0.0282 \\
        \citeauthor{jain2025attraction} 
        & \textbf{0.0404 ± 0.0341} 
        & 0.2904 ± 0.1463 
        & 73.4757 ± 13.9227 
        & 0.2285 ± 0.0540 \\
        \textbf{\ensuremath{\rads} (Ours)} 
        & 0.1356 ± 0.0838 
        & 0.0718 ± 0.0535 
        & 47.9868 ± 5.4545 
        & 0.2785 ± 0.0497 \\
        \bottomrule
    \end{tabular}
\end{table}

\subsection{Zero-Shot Generalization to MemBench Dataset}
\label{app:results_membench}

Table~\ref{tab:results_membench} reports zero-shot generalization to the MemBench dataset of 3000 memorized prompts \citep{hong2024membench}. Although \ensuremath{\rads} is trained only on the Webster training prompts, it achieves the lowest \ensuremath{\sscdtarget} on MemBench, showing that the learned steering policy transfers beyond the training prompt distribution. Compared to \citet{jain2025attraction}, \ensuremath{\rads} also substantially improves image quality while avoiding the high \ensuremath{\sscdprompts} associated with mode collapse.

\begin{table}[H]
    \centering
    \caption{\textbf{Zero-Shot Generalization to MemBench with Stable Diffusion v1.4 and DDIM Sampling.} We evaluate on the MemBench dataset of 3000 memorized prompts \citep{hong2024membench}.}
    \label{tab:results_membench}
    \small
    \setlength{\tabcolsep}{4pt}
    \begin{tabular}{lccccc}
        \toprule
        Method & \ensuremath{\sscdtarget} (↓) & \ensuremath{\sscdseeds} (↓) & \ensuremath{\sscdprompts} (↓) & \ensuremath{\fid} (↓) & \ensuremath{\clip} (↑) \\
        \midrule
        No mitigation 
        & 0.5030 ± 0.2836 
        & 0.4751 ± 0.3080 
        & \textbf{0.0367 ± 0.0164} 
        & 22.0230 ± 0.8903 
        & \textbf{0.3005 ± 0.0392} \\
        \citeauthor{wen2024detecting} 
        & 0.2091 ± 0.1494 
        & 0.1219 ± 0.0710 
        & 0.0412 ± 0.0178 
        & \textbf{18.7822 ± 1.9536} 
        & 0.2998 ± 0.0388 \\
        \citeauthor{ren2024crossattn} 
        & 0.2184 ± 0.1482 
        & 0.1508 ± 0.1121 
        & 0.0465 ± 0.0192 
        & 20.1319 ± 2.6071 
        & 0.2937 ± 0.0426 \\
        \citeauthor{jain2025attraction} 
        & 0.1775 ± 0.1226 
        & \textbf{0.0726 ± 0.0741} 
        & 0.1782 ± 0.0933 
        & 38.5641 ± 7.2228 
        & 0.2520 ± 0.0470 \\
        \textbf{\ensuremath{\rads} (Ours)} 
        & \textbf{0.1449 ± 0.0439} 
        & 0.1006 ± 0.0598 
        & 0.0516 ± 0.0238 
        & 26.7537 ± 6.5509 
        & 0.2540 ± 0.0411 \\
        \bottomrule
    \end{tabular}
\end{table}

\subsection{DDPM Sampling}
\label{app:results_sdv1_ddpm}

Table~\ref{tab:results_sdv1_ddpm} evaluates Stable Diffusion v1.4 with DDPM sampling. \ensuremath{\rads} continues to achieve the best \ensuremath{\sscdtarget}, \ensuremath{\sscdprompts}, and \ensuremath{\fid}, showing that the steering policy is effective beyond deterministic DDIM sampling.

\begin{table}[H]
    \centering
    \caption{\textbf{\citet{webster2023extraction} Dataset with Stable Diffusion v1.4 and DDPM Sampling.} We evaluate the 500 memorized prompts using DDPM sampling \citep{ho2020ddpm}.}
    \label{tab:results_sdv1_ddpm}
    \small
    \setlength{\tabcolsep}{4pt}
    \begin{tabular}{lccccc}
        \toprule
        Method & \ensuremath{\sscdtarget} (↓) & \ensuremath{\sscdseeds} (↓) & \ensuremath{\sscdprompts} (↓) & \ensuremath{\fid} (↓) & \ensuremath{\clip} (↑) \\
        \midrule
        No mitigation 
        & 0.6328 ± 0.2506 
        & 0.4982 ± 0.2968 
        & 0.0942 ± 0.1027 
        & 42.0048 ± 6.5753
        & \textbf{0.3133 ± 0.0276} \\
        \citeauthor{wen2024detecting} 
        & 0.4400 ± 0.2872 
        & 0.2441 ± 0.1881 
        & 0.0639 ± 0.0755 
        & 32.0970 ± 3.3878 
        & 0.3077 ± 0.0324 \\
        \citeauthor{ren2024crossattn} 
        & 0.4039 ± 0.2665 
        & 0.2474 ± 0.1994 
        & 0.0561 ± 0.0600 
        & 29.4955 ± 3.9289 
        & 0.2962 ± 0.0502 \\
        \citeauthor{hintersdorf2024nemo} 
        & 0.4192 ± 0.2412 
        & 0.2801 ± 0.2234 
        & 0.0531 ± 0.0548 
        & 31.9626 ± 3.5206 
        & 0.3123 ± 0.0309 \\
        \citeauthor{jain2025attraction} 
        & 0.2507 ± 0.1999 
        & \textbf{0.0654 ± 0.0632} 
        & 0.1248 ± 0.0589 
        & 37.1632 ± 8.1677 
        & 0.2455 ± 0.0529 \\
        \textbf{\ensuremath{\rads} (Ours)} 
        & \textbf{0.2206 ± 0.1029} 
        & 0.1593 ± 0.1075 
        & \textbf{0.0337 ± 0.0173} 
        & \textbf{29.4008 ± 5.2703} 
        & 0.2875 ± 0.0404 \\
        \bottomrule
    \end{tabular}
\end{table}

\subsection{Realistic Vision}
\label{app:results_realvis_ddim}

Table~\ref{tab:results_realvis_ddim} evaluates Realistic Vision with DDIM sampling on the \citet{webster2023extraction} dataset. \ensuremath{\rads} again achieves the lowest \ensuremath{\sscdprompts}, indicating that it avoids prompt-level mode collapse. While \citet{jain2025attraction} achieves the lowest \ensuremath{\sscdtarget} and \ensuremath{\sscdseeds}, it substantially degrades \ensuremath{\fid} and \ensuremath{\clip}.

\begin{table}[H]
    \centering
    \caption{\textbf{\citet{webster2023extraction} Dataset with Realistic Vision and DDIM Sampling.} We evaluate the 500 memorized prompts using Realistic Vision.}
    \label{tab:results_realvis_ddim}
    \small
    \setlength{\tabcolsep}{4pt}
    \begin{tabular}{lccccc}
        \toprule
        Method & \ensuremath{\sscdtarget} (↓) & \ensuremath{\sscdseeds} (↓) & \ensuremath{\sscdprompts} (↓) & \ensuremath{\fid} (↓) & \ensuremath{\clip} (↑) \\
        \midrule
        No mitigation 
        & 0.5839 ± 0.2677 
        & 0.4901 ± 0.3017 
        & 0.0865 ± 0.0941 
        & 36.5444 ± 3.4349 
        & \textbf{0.3183 ± 0.0294} \\
        \citeauthor{wen2024detecting} 
        & 0.3827 ± 0.2573 
        & 0.2136 ± 0.1327 
        & 0.0596 ± 0.0623 
        & \textbf{28.2020 ± 4.3097} 
        & 0.3107 ± 0.0339 \\
        \citeauthor{ren2024crossattn}  
        & 0.4318 ± 0.2672 
        & 0.2950 ± 0.2093 
        & 0.0826 ± 0.0934 
        & 34.8076 ± 1.9886
        & 0.3014 ± 0.0501 \\
        \citeauthor{hintersdorf2024nemo} 
        & 0.4035 ± 0.2523 
        & 0.3154 ± 0.2328 
        & 0.0514 ± 0.0518 
        & 29.4062 ± 3.8879 
        & 0.3175 ± 0.0312 \\
        \citeauthor{jain2025attraction} 
        & \textbf{0.1544 ± 0.0527} 
        & \textbf{0.0489 ± 0.0394} 
        & 0.3199 ± 0.1804 
        & 84.8682 ± 40.7578	 
        & 0.2139 ± 0.0537 \\
        \textbf{\ensuremath{\rads} (Ours)} 
        & 0.1992 ± 0.0797 
        & 0.1834 ± 0.1066 
        & \textbf{0.0405 ± 0.0266} 
        & 33.0128 ± 9.7632 
        & 0.2839 ± 0.0391 \\
        \bottomrule
    \end{tabular}
\end{table}

\subsection{Generalization to Larger Architecture: Stable Diffusion v2.1 on the Webster Dataset}
\label{app:results_sdv2}

Table~\ref{tab:results_sdv2_webster} evaluates whether \ensuremath{\rads} generalizes to a larger diffusion backbone. We evaluate Stable Diffusion v2.1 ($768\times768$) on the \citet{webster2023extraction} dataset using the same memorization and utility metrics as before, with all methods using the same checkpoint and sampling settings. \ensuremath{\rads} achieves the lowest \ensuremath{\sscdtarget}, \ensuremath{\sscdseeds}, and \ensuremath{\sscdprompts}, reducing \ensuremath{\sscdtarget} by 25\% relative to the strongest baseline, while keeping \ensuremath{\fid} within one standard deviation of the best method. This comes at a modest cost in prompt alignment (\ensuremath{\clip} 0.271 vs.\ 0.290 for \citet{ren2024crossattn}), consistent with the trend across all methods that stronger memorization mitigation lowers \ensuremath{\clip}.

\begin{table}[H]
\centering
\caption{\textbf{\citet{webster2023extraction} Dataset with Stable Diffusion v2.1.} We evaluate generalization to a larger diffusion backbone. Lower values are better for \ensuremath{\sscdtarget}, \ensuremath{\sscdseeds}, \ensuremath{\sscdprompts}, and \ensuremath{\fid}; higher values are better for \ensuremath{\clip}.}
\label{tab:results_sdv2_webster}
\small
\setlength{\tabcolsep}{4pt}
\begin{tabular}{lccccc}
    \toprule
    Method & \ensuremath{\sscdtarget} (↓) & \ensuremath{\sscdseeds} (↓) & \ensuremath{\sscdprompts} (↓) & \ensuremath{\fid} (↓) & \ensuremath{\clip} (↑) \\
    \midrule
    No mitigation
    & 0.3576 ± 0.2703
    & 0.2444 ± 0.1316
    & 0.0970 ± 0.0999
    & 36.1380 ± 3.7306
    & \textbf{0.3130 ± 0.0351} \\

    \citeauthor{wen2024detecting}
    & 0.3066 ± 0.2445
    & 0.1651 ± 0.0841
    & 0.0717 ± 0.0704
    & \textbf{27.7961 ± 3.9140}
    & 0.3024 ± 0.0404 \\

    \citeauthor{ren2024crossattn}
    & 0.2733 ± 0.1960
    & 0.1729 ± 0.1188
    & 0.0672 ± 0.0552
    & 30.6398 ± 3.4869
    & 0.2897 ± 0.0548 \\

    \citeauthor{jain2025attraction}
    & 0.3108 ± 0.2413
    & 0.2023 ± 0.1001
    & 0.0902 ± 0.0878
    & 30.5580 ± 4.9648
    & 0.3052 ± 0.0392 \\

    \textbf{\ensuremath{\rads} (Ours)}
    & \textbf{0.2054 ± 0.1020}
    & \textbf{0.1225 ± 0.0804}
    & \textbf{0.0668 ± 0.0427}
    & 29.5959 ± 4.9567
    & 0.2712 ± 0.0573 \\
    \bottomrule
\end{tabular}
\end{table}

\subsection{Preservation on Standard Prompts}
\label{app:coco_preservation}

To evaluate whether $\rads$ preserves generation quality and prompt alignment outside memorization-focused benchmarks, we sample 1,000 captions from COCO~\citep{lin2014microsoft} and compare unmitigated SD v1.4 with the existing $\rads$ controller, without additional training on these captions. Unlike Webster, these captions are not selected for eliciting memorized outputs. This evaluation therefore assesses preservation on general-purpose prompts rather than a verified memorization-free subset.

We evaluate five initial-noise seeds and additionally average $\rads$ results over five training seeds. FID is computed against the COCO validation images. We report CLIP for prompt alignment, FID for distributional image quality, and $\sscdseeds$ and $\sscdprompts$ for diversity across seeds and prompts.

\begin{table}[h]
    \centering
    \small
    \caption{Preservation on 1,000 randomly sampled COCO captions.
    $\rads$ maintains similar FID and prompt-level diversity,
    with a small decrease in CLIP and increased seed diversity.
    Lower SSCD and FID are better; higher CLIP is better.}
    \label{tab:coco_preservation}
    \begin{tabular}{lcccc}
        \toprule
        Method & $\sscdseeds$ ($\downarrow$)
        & $\sscdprompts$ ($\downarrow$)
        & FID ($\downarrow$) & CLIP ($\uparrow$) \\
        \midrule
        No mitigation
        & 0.1591 $\pm$ 0.0681
        & 0.0451 $\pm$ 0.0159
        & 5.4529 $\pm$ 0.4544
        & 0.3118 $\pm$ 0.0276 \\
        $\rads$ (Ours)
        & 0.1379 $\pm$ 0.0656
        & 0.0446 $\pm$ 0.0157
        & 5.2603 $\pm$ 0.9479
        & 0.3077 $\pm$ 0.0290 \\
        \bottomrule
    \end{tabular}
\end{table}

As shown in Table~\ref{tab:coco_preservation}, CLIP decreases by approximately 1.3\%, while FID and prompt-level similarity remain close to their unmitigated values. Seed-level similarity decreases, indicating increased diversity across initializations. These results suggest that $\rads$ largely preserves nominal generation behavior on standard prompts, although they do not establish preservation for every prompt.

\subsection{Controlled Memorization Evaluation on SDXL}
\label{app:sdxl}

\paragraph{Motivation and Controlled Setup.}
We evaluate SDXL to assess whether the CFG-based risk signal and learned steering policy extend to a larger architecture with dual text encoders. Because established SDXL memorization benchmarks with verified prompt-image pairs remain limited, we construct a controlled memorization setting by LoRA fine-tuning SDXL-base on 51 Webster prompt-image pairs, comprising 22 MV and 29 RV examples. This experiment evaluates mitigation of induced memorization, rather than extraction of memorized examples from the original
pretrained SDXL model.

We apply rank-8 LoRA to the U-Net attention layers and train for 60 epochs (1,530 steps), using batch size 2, learning rate $5\times10^{-4}$, 5\% warm-up, linear learning-rate decay, and \texttt{bfloat16} precision. The selected checkpoint achieves mean $\sscdtarget=0.4875$ across the 51 trained pairs, with variation in memorization strength.

\paragraph{Validation of the CFG-Based Risk Signal.}
We compare CFG norms before and after LoRA fine-tuning on the 51 trained prompts and 296 held-out TV and non-memorized prompts. As shown in Table~\ref{tab:sdxl_cfg}, fine-tuning substantially increases the mean CFG norm on trained prompts, while the change on held-out prompts is much smaller. The same trend persists after normalization by the number of latent elements. These results support the sensitivity of the CFG-based signal to induced memorization in SDXL.

\begin{table}[t]
    \centering
    \small
    \caption{CFG norms averaged over denoising steps before
    and after SDXL LoRA fine-tuning.}
    \label{tab:sdxl_cfg}
    \begin{tabular}{lccc}
        \toprule
        Prompt set & Base SDXL & LoRA SDXL & Change \\
        \midrule
        Trained (51)
        & 3.344 $\pm$ 0.472
        & 7.207 $\pm$ 1.352
        & 3.863 $\pm$ 0.993 \\
        Held-out (296)
        & 3.683 $\pm$ 0.883
        & 3.777 $\pm$ 1.412
        & 0.094 $\pm$ 0.503 \\
        \bottomrule
    \end{tabular}
\end{table}

\paragraph{RADS Training and Evaluation.}
After constructing the memorizing checkpoint, we freeze its weights throughout $\rads$ training and evaluation. We train the caption VAE for 35,000 steps on Webster and Conceptual Captions, compressing the concatenated 2,048-dimensional dual-encoder representation into a 64-dimensional action space. We then train $\rads$ for a single training seed for 20 epochs and evaluate all 51 trained prompts using ten initial-noise seeds. This evaluates mitigation on the controlled memorization set, rather than held-out prompt generalization.

\begin{table}[t]
    \centering
    \small
    \setlength{\tabcolsep}{3pt}
    \caption{Memorization mitigation on the controlled SDXL
    checkpoint. FID is computed by pooling generations
    across initial-noise seeds.}
    \label{tab:sdxl_results}
    \begin{tabular}{lccccc}
        \toprule
        Method
        & $\sscdtarget$ ($\downarrow$)
        & $\sscdseeds$ ($\downarrow$)
        & $\sscdprompts$ ($\downarrow$)
        & FID ($\downarrow$)
        & CLIP ($\uparrow$) \\
        \midrule
        No mitigation
        & 0.4875 $\pm$ 0.1244
        & 0.7056 $\pm$ 0.1311
        & 0.2106 $\pm$ 0.0848
        & 263.30
        & 0.3066 $\pm$ 0.0183 \\
        $\rads$ (Ours)
        & 0.3562 $\pm$ 0.1911
        & 0.3844 $\pm$ 0.2427
        & 0.2027 $\pm$ 0.1107
        & 239.45
        & 0.2939 $\pm$ 0.0398 \\
        \bottomrule
    \end{tabular}
\end{table}

\paragraph{Results and Scope.}
$\rads$ reduces $\sscdtarget$ by 26.9\% and $\sscdseeds$ by 45.5\%, while CLIP decreases by 4.1\%. Pooled FID also decreases, although its absolute value should be interpreted cautiously because the small, specialized prompt set differs substantially from the COCO reference distribution. These results provide preliminary evidence that $\rads$ extends to SDXL under standard CFG. They do not establish effectiveness on naturally occurring memorization in pretrained SDXL or on models without an explicit CFG conditioning.

\section{Reliability of the Safety Critic}
\label{app:reliability}

We evaluate whether the learned safety critic ranks states according to their association with subsequent memorization. We query $Q_{\pi_\phi}^{\mathrm{safe}}(\state_t,0)$ on states from unsteered trajectories using SD v1.4 on the \cite{webster2023extraction} dataset. This value evaluates zero steering action at the current step, followed by the learned policy at subsequent steps. Zero action still applies VAE reconstruction to the caption embedding.

The evaluation labels instead indicate whether the corresponding uncontrolled rollout memorizes. Consequently, the reported threshold statistics measure agreement with uncontrolled outcomes, rather than prediction errors for outcomes under the learned policy. We compare the critic's ROC-AUC with that of the instantaneous CFG norm.

\begin{table}[h]
\centering
\small
\caption{Safety critic evaluation against uncontrolled-rollout
memorization labels, with 242 positive state samples at each
denoising step. The nonnegative-value fraction is the fraction
of these positive samples with
$Q_{\pi_\phi}^{\mathrm{safe}}(\state_t,0)\geq0$.
Because the critic is policy-conditioned while the labels
describe uncontrolled outcomes, this fraction is not a
false-negative rate for policy-conditioned outcomes.}
\label{tab:q_reliability}
\begin{tabular}{c c c c}
\toprule
Step & $Q$-AUC & CFG-norm AUC & Fraction $Q_{\pi_\phi}^{\mathrm{safe}}(\state_t,0)\geq0$ \\
\midrule
1  & 0.733 & 0.646 & 0.998 \\
5  & 0.856 & 0.704 & 0.655 \\
10 & 0.858 & 0.668 & 0.480 \\
\bottomrule
\end{tabular}
\end{table}

The critic achieves ROC-AUCs of 0.733, 0.856, and 0.858 at steps 1, 5, and 10, exceeding the corresponding instantaneous CFG-norm scores. This demonstrates useful ranking of states against uncontrolled memorization outcomes.

Many states whose uncontrolled rollouts memorize receive nonnegative policy-conditioned safety values. This is consistent with possible recovery under subsequent steering, but does not establish recovery without evaluating controlled continuations from those states. 

\vspace{-1em}
\section{Ablation Studies}
\label{app:ablations}
\vspace{-0.5em}
\ensuremath{\rads} consists of several components that jointly prevent memorization while preserving high-quality image generation. As discussed in Section~\ref{sec:ablation}, we perform ablations to isolate the contribution of each component. In particular, we compare full \ensuremath{\rads} against the following variants:

\begin{itemize}[left=0pt]
    \item \textbf{No constraint:} Instead of using a constrained RL solution, we remove the reachability constraint and train SAC to maximize only the semantic-alignment return from Equation~\ref{eqn:reward}. This ablation uses the same VAE action space, policy architecture, and task reward as \ensuremath{\rads}, but removes the safety constraint by setting the Lagrange multiplier to zero. The resulting policy objective becomes
    \begin{equation}
        J_{\pi}^{\text{no-constraint}}(\phi)
        =
        \mathbb{E}_{\state_t \sim \mathcal{D}, \control_t \sim \pi_\phi}
        \left[
            Q_{\omega}^{\text{task}}(\state_t, \control_t)
            +
            \alpha \mathcal{H}(\pi_\phi(\cdot \mid \state_t))
        \right].
        \label{eqn:no_constraint_ablation}
    \end{equation}
    This tests whether optimizing for prompt alignment alone is sufficient to avoid memorization.

    \item \textbf{Classifier guidance:} Instead of training a full RL policy, we train a state-only safety value function \ensuremath{V^{\text{safe}}_\psi(\state_t)} and use it for classifier-guided denoising. This ablation does not learn an actor and does not learn a control-value function \ensuremath{Q^{\text{safe}}(\state_t,\control_t)}. 
    
    The safety value function is trained with the same discounted
    reachability-inspired backup as $\rads$, but without action
    conditioning or caption VAE reconstruction. We evaluate the
    target using the original caption embedding $\promptemb$:
    \[
    \ell_{\mathrm{base}}(\state_t)
    =
    -\tanh\left(
    \eta\left(
    \|\epsilon_\theta(\latent_{T-t},\promptemb)
    -\epsilon_\theta(\latent_{T-t},\emptyset)\|_2-\beta
    \right)
    \right).
    \]
    For a transition $(\state_t,\state_{t+1})$, we use
    \begin{equation}
    \widehat V_t^{\mathrm{safe}} =
    \begin{cases}
    (1-\gamma)\ell_{\mathrm{base}}(\state_t)
    +\gamma\min\left\{
    \ell_{\mathrm{base}}(\state_t),
    V_{\psi'}^{\mathrm{safe}}(\state_{t+1})
    \right\},
    & t<T-1,\\
    \ell_{\mathrm{base}}(\state_{T-1}),
    & t=T-1.
    \end{cases}
    \end{equation}
    
    We then minimize
    \begin{equation}
        \mathcal{L}_{\text{safe}}(\psi)
        =
        \mathbb{E}_{\state_t \sim \mathcal{D}}
        \left[
            \left(
                V^{\text{safe}}_\psi(\state_t)
                -
                \widehat{V}^{\text{safe}}_t
            \right)^2
        \right].
        \label{eqn:classifier_guidance_value_loss}
    \end{equation}

    At inference time, we use \ensuremath{V^{\text{safe}}_\psi} analogously to classifier guidance. Rather than sampling a learned control from \ensuremath{\pi_\phi}, we modify the denoising update using the gradient of the safety value with respect to the current latent state. Writing the unmodified denoising update as $\latent_{T-(t+1)}
        =
        f_{\text{DM}}(\latent_{T-t}, \promptemb, T-t)$
    the classifier-guided variant applies an additional safety-gradient term:
    \begin{equation}
        \latent_{T-(t+1)}
        =
        f_{\text{DM}}(\latent_{T-t}, \promptemb, T-t)
        +
        \eta_{\text{cg}}
        \nabla_{\latent_{T-t}}
        V^{\text{safe}}_\psi(\state_t),
        \label{eqn:classifier_guidance_state_update}
    \end{equation}
    where \ensuremath{\eta_{\text{cg}}} is the classifier-guidance scale. Equivalently, this ablation uses the learned safety value for a gradient signal during denoising, rather than learning an amortized constrained policy over caption-embedding controls. This tests whether a safety value function alone is sufficient, or whether full constrained RL is needed to learn effective steering controls. 
    
    In our implementation of this ablation, the architecture is largely preserved from \ensuremath{\rads}, with three differences. First, the safety value function is state-only: it takes \ensuremath{\state_t} as input rather than the state-control pair \ensuremath{(\state_t,\control_t)}. Second, we set the dropout rate in the shared encoder to 0.4, whereas full \ensuremath{\rads} uses no dropout. Third, we train the value function for 30 epochs. We found that the latter two changes were necessary to prevent the state-only value function from overfitting.

Finally, we set the classifier-guidance scale to \ensuremath{\eta_{\mathrm{cg}} = 500.0}. 
Although this value may appear large, it was selected using a sweep over 
\ensuremath{\eta_{\mathrm{cg}} \in \{10.0, 50.0, 100.0, 500.0, 1000.0\}}. 
As shown in Table~\ref{tab:classifier_guidance_scale_sweep}, smaller guidance scales preserve CLIP alignment and FID but provide limited memorization mitigation, while 
\ensuremath{\eta_{\mathrm{cg}} = 1000.0} further reduces \ensuremath{\sscdtarget} at the cost of severe perceptual degradation, suggesting that overly strong guidance can push trajectories into an out-of-distribution regime rather than yielding meaningful mitigation. 
We therefore use \ensuremath{\eta_{\mathrm{cg}} = 500.0} as the strongest guidance scale that avoids this degenerate behavior while still reducing memorization. 
This small-scale sweep was performed over 5 evaluation seeds ($N=5$) and a single training seed.

\begin{table}[H]
\centering
\caption{Classifier-guidance scale sweep. We evaluate different values of the classifier-guidance scale \ensuremath{\eta_{\mathrm{cg}}}. Lower \ensuremath{\sscdtarget} indicates stronger memorization mitigation, lower \ensuremath{\fid} indicates higher perceptual quality, and higher \ensuremath{\clip} indicates better prompt alignment.}
\label{tab:classifier_guidance_scale_sweep}
\small
\setlength{\tabcolsep}{8pt}
\begin{tabular}{lccc}
    \toprule
    \ensuremath{\eta_{\mathrm{cg}}} & \ensuremath{\sscdtarget} (↓) & \ensuremath{\fid} (↓) & \ensuremath{\clip} (↑) \\
    \midrule
    10.0 
    & 0.5996 $\pm$ 0.2477 
    & 41.2483 $\pm$ 3.1731
    & 0.3061 $\pm$ 0.0322 \\
    50.0 
    & 0.5924 $\pm$ 0.2501 
    & 41.1191 $\pm$ 2.7407
    & \textbf{0.3069 $\pm$ 0.0310} \\
    100.0 
    & 0.5609 $\pm$ 0.2480 
    & \textbf{39.0535 $\pm$ 3.7951}
    & \textbf{0.3069 $\pm$ 0.0309} \\
    500.0 
    & 0.3317 $\pm$ 0.2283 
    & 42.7457 $\pm$ 10.2887 
    & 0.2877 $\pm$ 0.0419 \\
    1000.0 
    & \textbf{0.1966 $\pm$ 0.1422} 
    & 80.9265 $\pm$ 12.2767
    & 0.2419 $\pm$ 0.0494 \\
    \bottomrule
\end{tabular}
\end{table}

    \item \textbf{Standard Bellman:} Instead of using the reachability-style minimum over the trajectory, we replace the safety critic target with the standard discounted Bellman backup used in RL. This corresponds to optimizing the expected discounted sum of safety rewards, rather than the worst reachable safety value along the denoising trajectory. The target becomes
    \begin{equation}
        \widehat Q_t^{\mathrm{sum}} =
\begin{cases}
\ell(\state_t,\control_t)
+\gamma Q_{\psi'}^{\mathrm{sum}}
(\state_{t+1},\control'_{t+1}),
& t<T-1,\\
\ell(\state_{T-1},\control_{T-1}),
& t=T-1.
\end{cases}
        \label{eqn:standard_bellman_ablation}
    \end{equation}
    where \ensuremath{\control'_{t+1} \sim \pi_\phi(\cdot \mid \state_{t+1})}. The critic is trained by minimizing
    \begin{equation}
        \mathcal{L}_{\text{sum}}(\psi)
        =
        \mathbb{E}_{(\state_t,\control_t) \sim \mathcal{D}}
        \left[
            \left(
                Q^{\text{sum}}_{\psi}(\state_t,\control_t)
                -
                \widehat{Q}^{\text{sum}}_t
            \right)^2
        \right].
        \label{eqn:standard_bellman_loss}
    \end{equation}
    This ablation tests whether a standard cumulative safety objective is sufficient, or whether the reachability-style minimum is necessary for detecting trajectories that become unsafe at any point during denoising. In terms of the implementation, we use the same hyperparameters and training protocols as $\rads$, except for the Bellman update.

    \item \textbf{VAE-only:} We set $\control_t=0$ at every denoising step, so conditioning uses $\promptembt=\mathrm{Dec}(\mathrm{Enc}(\promptemb))$ without learned steering. This isolates mitigation induced by VAE reconstruction. The unmitigated baseline instead uses the original caption embedding $\promptemb$ without VAE reconstruction.
\end{itemize}

Table~\ref{tab:rads_ablation} reports the results across $N=10$ evaluation seeds and 5 training seeds in SD v1.4. Removing the constraint worsens memorization metrics, indicating that semantic-alignment optimization alone does not steer the model away from memorization. Classifier guidance improves over the no-constraint variant, but performs worse than full \ensuremath{\rads}, suggesting that directly using the safety value as a gradient-guidance signal is less effective than learning a constrained policy. The standard Bellman variant reduces memorization but degrades perceptual quality and prompt alignment, often producing highly diverse but artifacted generations. VAE reconstruction alone reduces target similarity but falls substantially short of full $\rads$, indicating that reconstruction effects alone do not explain the observed mitigation. In contrast, full \ensuremath{\rads} achieves the strongest overall tradeoff across memorization mitigation, perceptual quality, and prompt alignment. Appendix~\ref{app:ablate_sum} further discusses the training instability and qualitative degradation of the standard Bellman variant.

\begin{table}[t]
\centering
\caption{Ablation study of \ensuremath{\rads} variants. The standard Bellman variant reduces target similarity but substantially degrades FID and CLIP, while full \ensuremath{\rads} achieves the best overall tradeoff.}
\label{tab:rads_ablation}
\small
\setlength{\tabcolsep}{4pt}
\begin{tabular}{lccccc}
    \toprule
    Variant & \ensuremath{\sscdtarget} (↓) & \ensuremath{\sscdseeds} (↓) & \ensuremath{\sscdprompts} (↓) & \ensuremath{\fid} (↓) & \ensuremath{\clip} (↑) \\
    \midrule
    No constraint 
    & 0.500 $\pm$ 0.234 
    & 0.345 $\pm$ 0.245 
    & 0.079 $\pm$ 0.077 
    & 36.35 $\pm$ 4.17 
    & 0.300 $\pm$ 0.033 \\

    Classifier guidance 
    & 0.389 $\pm$ 0.258 
    & 0.327 $\pm$ 0.237 
    & 0.069 $\pm$ 0.054 
    & 38.58 $\pm$ 7.60 
    & 0.290 $\pm$ 0.043 \\

    Standard Bellman 
    & \textbf{0.182 $\pm$ 0.072} 
    & \textbf{0.078 $\pm$ 0.058} 
    & 0.065 $\pm$ 0.037 
    & 41.77 $\pm$ 16.76 
    & 0.240 $\pm$ 0.046 \\

    VAE-only & 0.572 $\pm$ 0.252 & 0.401 $\pm$ 0.274 & 0.081 $\pm$ 0.083 & 38.14 $\pm$ 4.89 & \textbf{0.305 $\pm$ 0.032} \\

    \textbf{Full \ensuremath{\rads}} 
    & 0.230 $\pm$ 0.111 
    & 0.155 $\pm$ 0.110 
    & \textbf{0.041 $\pm$ 0.023} 
    & \textbf{31.57 $\pm$ 5.82} 
    & 0.291 $\pm$ 0.037 \\
    \bottomrule
\end{tabular}
\end{table}

\subsection{Challenges of Training the Standard Bellman Variant}
\label{app:ablate_sum}

\begin{figure}[t]
    \centering
    \includegraphics[width=0.3\linewidth]{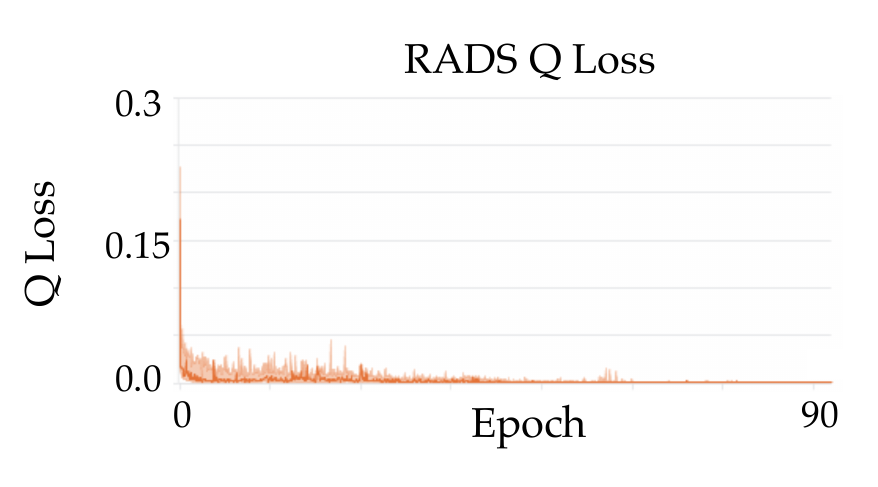}
    \includegraphics[width=0.3\linewidth]{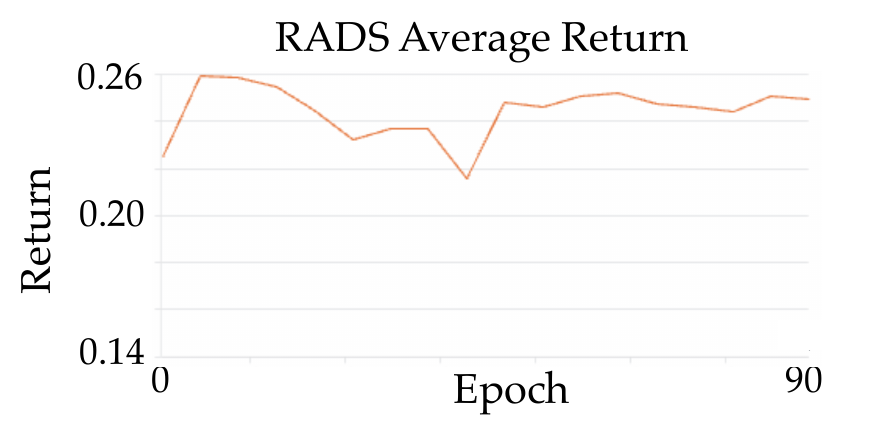}
    \includegraphics[width=0.3\linewidth]{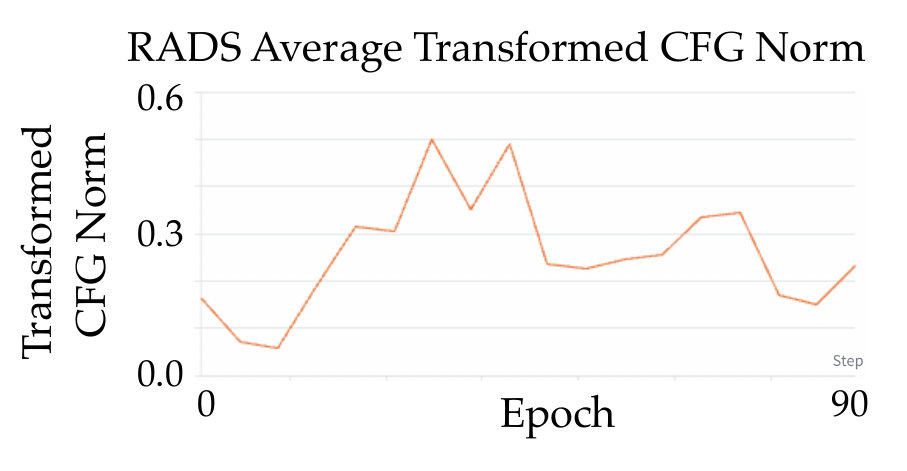}

    \includegraphics[width=0.3\linewidth]{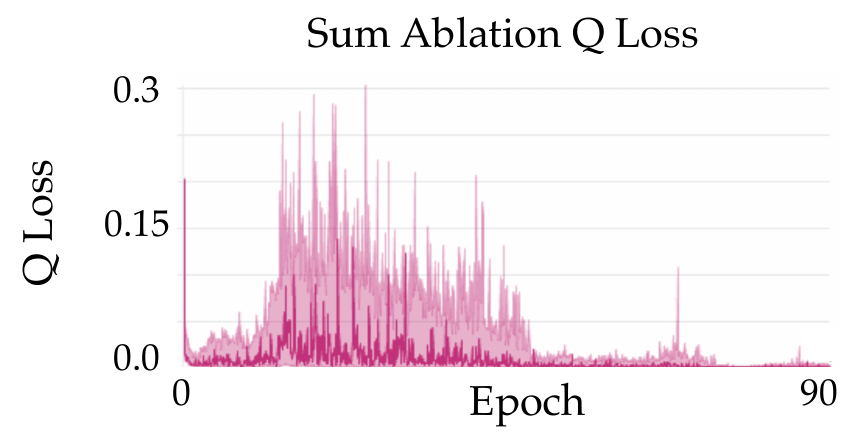}
    \includegraphics[width=0.3\linewidth]{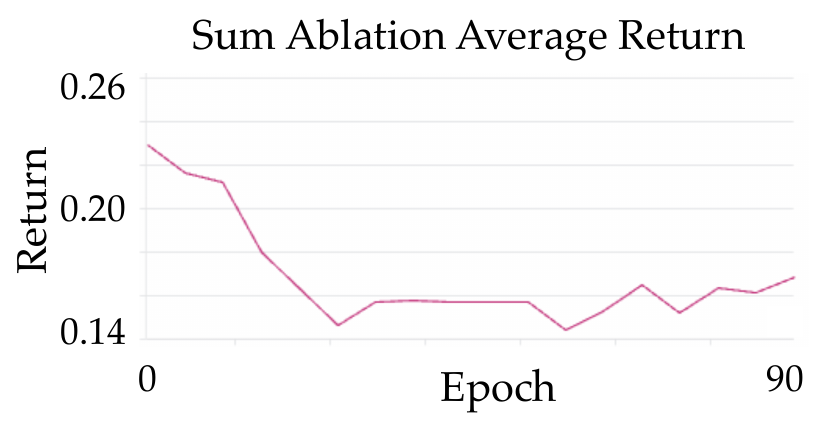}
    \includegraphics[width=0.3\linewidth]{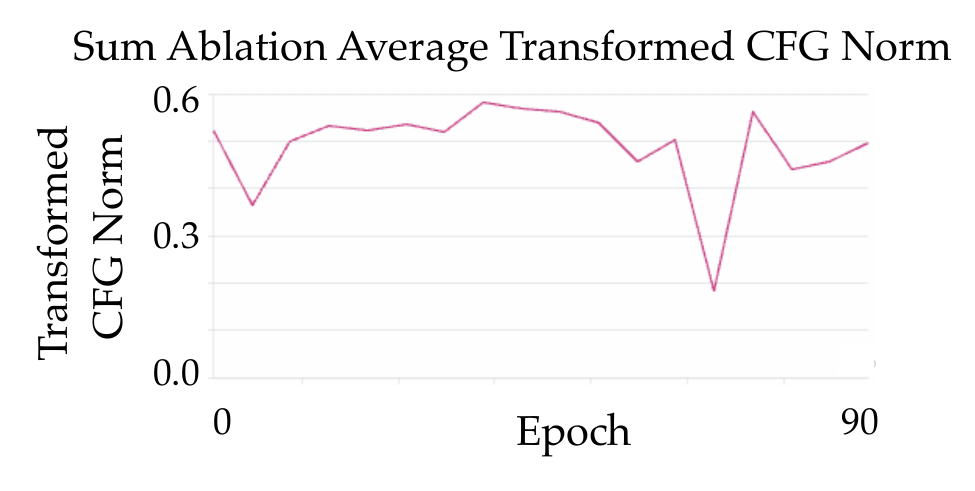}
    \caption{
        \textbf{Training instability of the standard Bellman ablation.}
        We compare full \ensuremath{\rads} against the standard Bellman sum ablation. Orange curves denote full \ensuremath{\rads}, while purple curves denote the sum ablation. Compared to full \ensuremath{\rads}, the sum ablation exhibits larger critic-loss spikes, greater variation in evaluation reward, and less stable transformed $L^2$-norms across training. These curves represent training dynamics for a single seed.
    }
    \label{fig:sum_ablation_training_curves}
\end{figure}

The standard Bellman ablation replaces the reachability-style minimum target with a cumulative discounted-sum safety objective. For this ablation, we select the best post-stabilization checkpoint by maximizing the validation sum of CLIP reward and $\ell(s, u)$. Figure~\ref{fig:sum_ablation_training_curves} shows one representative seed; Table~\ref{tab:rads_ablation} reports aggregate results over the full Webster prompt set and multiple seeds.

\begin{figure}[h]
    \centering
    \begin{subfigure}[t]{0.18\linewidth}
        \centering
        \includegraphics[width=\linewidth]{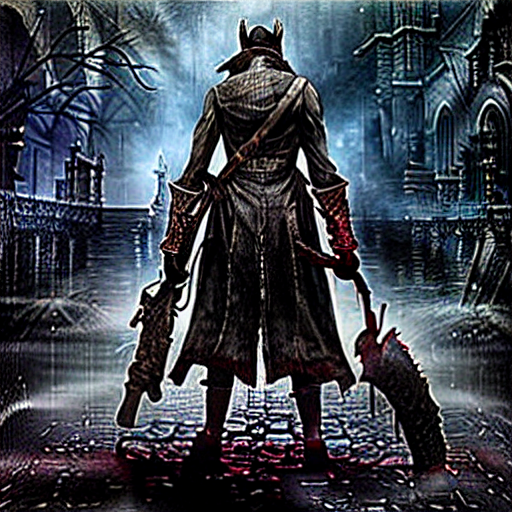}
        \caption{Memorized}
    \end{subfigure}
    \hspace{1em}
    \begin{subfigure}[t]{0.18\linewidth}
        \centering
        \includegraphics[width=\linewidth]{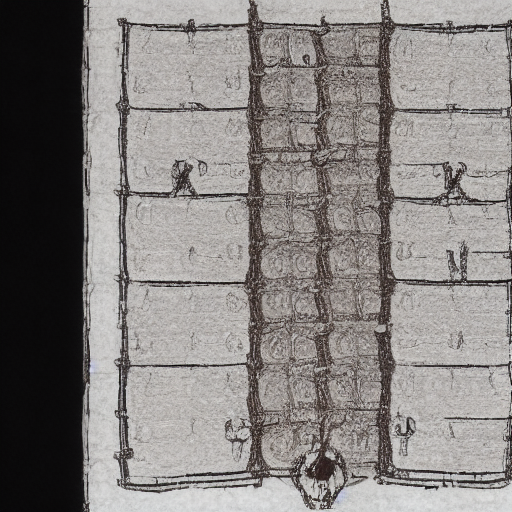}
        \caption{\footnotesize Standard Bellman}
    \end{subfigure}
    \hspace{1em}
    \begin{subfigure}[t]{0.18\linewidth}
        \centering
        \includegraphics[width=\linewidth]{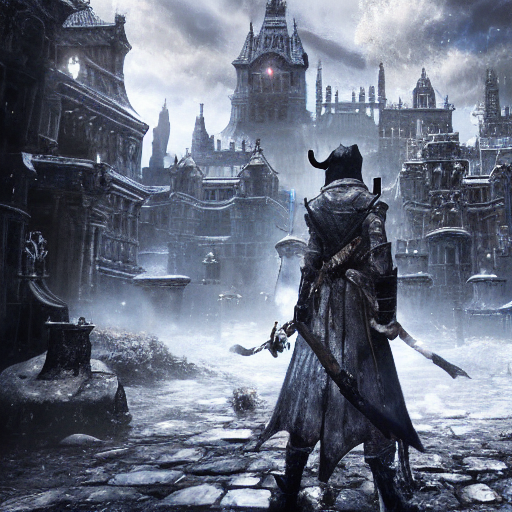}
        \caption{\ensuremath{\rads}}
    \end{subfigure}

    \caption{
        \textbf{Failure mode of the standard Bellman ablation.}
        Although the standard Bellman reduces target similarity, it can do so through significantly artifacted generations with much worse perceptual quality and prompt alignment. In contrast, \ensuremath{\rads} mitigates memorization while preserving a coherent image. This example uses the same training and evaluation seeds in the standard Bellman and $\rads$.
    }
    \label{fig:sum_ablation_failure}
\end{figure}

While the standard Bellman ablation can increase diversity, it is much less stable than full \ensuremath{\rads} and often sacrifices perceptual quality and prompt alignment, producing artifacted generations (Figure~\ref{fig:sum_ablation_failure}).

\vspace{-0.5em}

\subsection{Critic-Only Sampling Baseline}
\label{app:critic_only}

We also evaluate whether the learned actor can be replaced by direct inference-time search over actions using the learned safety critic. At each of the first five denoising steps, we sample $K=50$ candidate controls in the 64-dimensional latent action space and select the candidate that maximizes
\[
r(s_t) + \lambda \max(0, Q^{safe}(s_t,u_t)),
\]
where $r(s_t)$ is estimated from the predicted final image and $Q^{safe}$ is the learned safety critic. We sweep $\lambda \in \{0.1, 0.5, 1.0\}$ to vary the strength of the safety term.

Because the critic is trained on states throughout the denoising trajectory, we evaluate two variants. The first uses the true denoising timestep when querying $Q^{safe}$. The second remaps the first five intervention steps to the last five timesteps of the critic input, testing whether the critic is more informative near the end of denoising where memorization risk is easier to detect. We report results in Table~\ref{tab:critic_only_sampling}.

Overall, critic-only sampling does not substantially reduce memorization: $\sscdtarget$ remains near $0.47$ across settings, far above full \ensuremath{\rads}. Moreover, because this baseline requires evaluating many candidate actions at each denoising step, it is substantially slower than the learned actor. These results suggest that amortizing the search into a learned policy is important for effective steering in the high-dimensional caption-control space.

\begin{table}[h]
\centering
\caption{
Critic-only sampling baseline. We remove the learned actor and select controls by sampling \ensuremath{K=50} candidates during the first five denoising steps. Results are computed on 100 prompts with 5 evaluation seeds using a single training seed. Lower is better for \ensuremath{\sscdtarget}, \ensuremath{\sscdseeds}, \ensuremath{\sscdprompts}, and \ensuremath{\fid}; higher is better for \ensuremath{\clip}.
}
\label{tab:critic_only_sampling}
\small
\setlength{\tabcolsep}{3pt}
\resizebox{\linewidth}{!}{%
\begin{tabular}{lccccc}
    \toprule
    Method & \ensuremath{\sscdtarget} (↓) & \ensuremath{\sscdseeds} (↓) & \ensuremath{\sscdprompts} (↓) & \ensuremath{\fid} (↓) & \ensuremath{\clip} (↑) \\
    \midrule
    Critic-only, actual step, \ensuremath{\lambda=0.1}
    & 0.4701 ± 0.2777
    & 0.2883 ± 0.2062
    & 0.0876 ± 0.0884
    & 34.6970 ± 3.4496
    & 0.2972 ± 0.0343 \\

    Critic-only, actual step, \ensuremath{\lambda=0.5}
    & 0.4755 ± 0.2788
    & 0.2948 ± 0.2088
    & 0.0872 ± 0.0869
    & 35.2575 ± 4.0520
    & 0.2979 ± 0.0341 \\

    Critic-only, actual step, \ensuremath{\lambda=1.0}
    & 0.4754 ± 0.2802
    & 0.2981 ± 0.2107
    & 0.0875 ± 0.0870
    & 36.6125 ± 3.4206
    & \textbf{0.2986 ± 0.0323} \\

    Critic-only, last-5 remap, \ensuremath{\lambda=0.1}
    & 0.4683 ± 0.2736
    & 0.2855 ± 0.2081
    & 0.0863 ± 0.0882
    & 35.0568 ± 3.3399
    & 0.2961 ± 0.0352 \\

    Critic-only, last-5 remap, \ensuremath{\lambda=0.5}
    & 0.4752 ± 0.2799
    & 0.2993 ± 0.2116
    & 0.0872 ± 0.0882
    & 35.6130 ± 3.0133
    & 0.2976 ± 0.0345 \\

    Critic-only, last-5 remap, \ensuremath{\lambda=1.0}
    & 0.4726 ± 0.2805
    & 0.2903 ± 0.2040
    & 0.0869 ± 0.0879
    & 35.9110 ± 4.4518
    & 0.2981 ± 0.0341 \\

    \textbf{\ensuremath{\rads} (Ours)}
    & \textbf{0.2099 ± 0.0827}
    & \textbf{0.1532 ± 0.0922}
    & \textbf{0.0444 ± 0.0173}
    & \textbf{33.2905 ± 3.1018}
    & 0.2892 ± 0.0359 \\
    \bottomrule
\end{tabular}%
}
\end{table}

\subsection{Caption-Space versus Image-Latent Perturbations}
\label{app:caption_latent_diagnostic}

We compare the effects of perturbations in caption embedding space and image latent space on ten verified memorized SD v1.4 prompts with five initial-noise seeds. During the first ten denoising steps, we inject normalized Gaussian perturbations into either the conditional caption embedding or the diffusion latent, scaled relative to the norm of the corresponding representation. The higher unmitigated $\sscdtarget$ compared with Table~\ref{tab:rads_ablation} reflects the deliberately strongly memorized subset used here.

\begin{table}[t]
    \centering
    \small
    \caption{Caption-space versus image-latent perturbations.
    Caption perturbations provide a more gradual
    mitigation--alignment tradeoff on the evaluated prompts.}
    \label{tab:caption_latent_diagnostic}
    \begin{tabular}{lccc}
        \toprule
        Representation & Relative scale & CLIP $\uparrow$
        & $\sscdtarget$ $\downarrow$ \\
        \midrule
        None & 0.00 & 0.3404 $\pm$ 0.0215 & 0.8964 $\pm$ 0.0667 \\
        \midrule
        Caption & 0.10 & 0.3386 $\pm$ 0.0238 & 0.8669 $\pm$ 0.1506 \\
        Caption & 0.25 & 0.3391 $\pm$ 0.0230 & 0.8575 $\pm$ 0.1603 \\
        Caption & 0.50 & 0.3391 $\pm$ 0.0263 & 0.8032 $\pm$ 0.2445 \\
        Caption & 1.00 & 0.3325 $\pm$ 0.0325 & 0.6270 $\pm$ 0.3682 \\
        Caption & 2.00 & 0.3221 $\pm$ 0.0405 & 0.4719 $\pm$ 0.3909 \\
        Caption & 4.00 & 0.2867 $\pm$ 0.0607 & 0.3096 $\pm$ 0.3796 \\
        Caption & 8.00 & 0.2076 $\pm$ 0.0483 & 0.0538 $\pm$ 0.1034 \\
        \midrule
        Image latent & 0.05 & 0.3407 $\pm$ 0.0213 & 0.8951 $\pm$ 0.0679 \\
        Image latent & 0.10 & 0.3397 $\pm$ 0.0214 & 0.8923 $\pm$ 0.0686 \\
        Image latent & 0.25 & 0.2405 $\pm$ 0.0523 & 0.2992 $\pm$ 0.2911 \\
        Image latent & 0.50 & 0.1731 $\pm$ 0.0185 & 0.0212 $\pm$ 0.0317 \\
        Image latent & 1.00 & 0.1970 $\pm$ 0.0155 & 0.0100 $\pm$ 0.0188 \\
        \bottomrule
    \end{tabular}
\end{table}

Small image-latent perturbations have little effect on replication, while larger perturbations substantially reduce both target similarity and prompt alignment. Caption-space perturbations provide a more gradual tradeoff: at relative scale 1.0, $\sscdtarget$ decreases from 0.8964 to 0.6270, while CLIP decreases from 0.3404 to 0.3325. These results support our choice of caption-space control, but the diagnostic is limited to ten prompts and does not establish that a trained image-latent steering policy would be ineffective.


\section{Classifier-Free Guidance Analysis}
\label{app:safety_analysis}
\subsection{Cross-Model Validation of the CFG-Norm Signal}
\label{app:cfg_validation}

Our safety target relies on CFG norm as a proxy for memorization risk. To assess whether this relationship holds across diffusion backbones and denoising steps, we evaluate SD v1.4, SD v2.1, and Realistic Vision on Webster prompts. For each prompt, we generate ten unmitigated images from independent initial-noise seeds and record the conditional--unconditional prediction difference $\|\epsilon_\theta(x_{T-t},\promptemb) -\epsilon_\theta(x_{T-t},\emptyset)\|_2$ at all 50 denoising steps. At each step, we average the norms across seeds for each prompt and compute ROC-AUC against the Webster memorization labels, giving each prompt equal weight.

\begin{table}[t]
    \centering
    \small
    \setlength{\tabcolsep}{4pt}
    \caption{Cross-model validation of CFG norm as a memorization
    risk signal. We report mean CFG norms for memorized and
    non-memorized prompts, together with prompt-level ROC-AUC
    at selected denoising steps. Step 0 is the first denoising
    step and step 49 is the last.}
    \label{tab:cfg_validation}
    \begin{tabular}{llrrrrrr}
        \toprule
        Model & Statistic & Step 0 & Step 10 & Step 20
        & Step 30 & Step 40 & Step 49 \\
        \midrule
        SD v1.4 & Memorized
        & 3.38 & 6.01 & 8.53 & 10.18 & 11.98 & 18.86 \\
        & Non-memorized
        & 2.56 & 4.53 & 5.33 & 5.96 & 6.44 & 6.48 \\
        & ROC-AUC
        & 0.692 & 0.736 & 0.758 & 0.782 & 0.805 & 0.922 \\
        \midrule
        SD v2.1 & Memorized
        & 2.92 & 3.87 & 5.67 & 7.48 & 10.28 & 10.63 \\
        & Non-memorized
        & 1.58 & 2.40 & 3.18 & 3.68 & 3.99 & 4.43 \\
        & ROC-AUC
        & 0.933 & 0.882 & 0.927 & 0.948 & 0.953 & 0.918 \\
        \midrule
        Realistic Vision & Memorized
        & 3.60 & 5.86 & 9.04 & 11.58 & 13.17 & 24.05 \\
        & Non-memorized
        & 3.13 & 4.83 & 6.56 & 7.52 & 7.77 & 8.57 \\
        & ROC-AUC
        & 0.642 & 0.666 & 0.730 & 0.784 & 0.799 & 0.914 \\
        \bottomrule
    \end{tabular}
\end{table}

Memorized prompts exhibit higher mean CFG norms at every reported step across all three backbones. Final-step ROC-AUC reaches 0.922, 0.918, and 0.914 for SD v1.4, SD v2.1, and Realistic Vision, respectively. However, discrimination varies across models and timesteps, with weaker early-step separation for SD v1.4 and Realistic Vision. These results support CFG norm as an informative risk proxy, rather than an exact indicator of memorization. Its scale also varies across backbones, motivating model-specific calibration of the threshold $\beta$.

\subsection{Effect of RADS on CFG-Norm Trajectories}
The ablations in Appendix~\ref{app:ablations} show that the reachability constraint and learned policy are both important for mitigation. We next analyze how $\rads$ changes the denoising trajectory by examining the classifier-free guidance norm, which \citet{jain2025attraction} identifies as a signal of memorization risk.

\begin{wrapfigure}{r}{0.48\textwidth}
    \centering
    \vspace{-10pt}
    \includegraphics[width=0.43\textwidth]{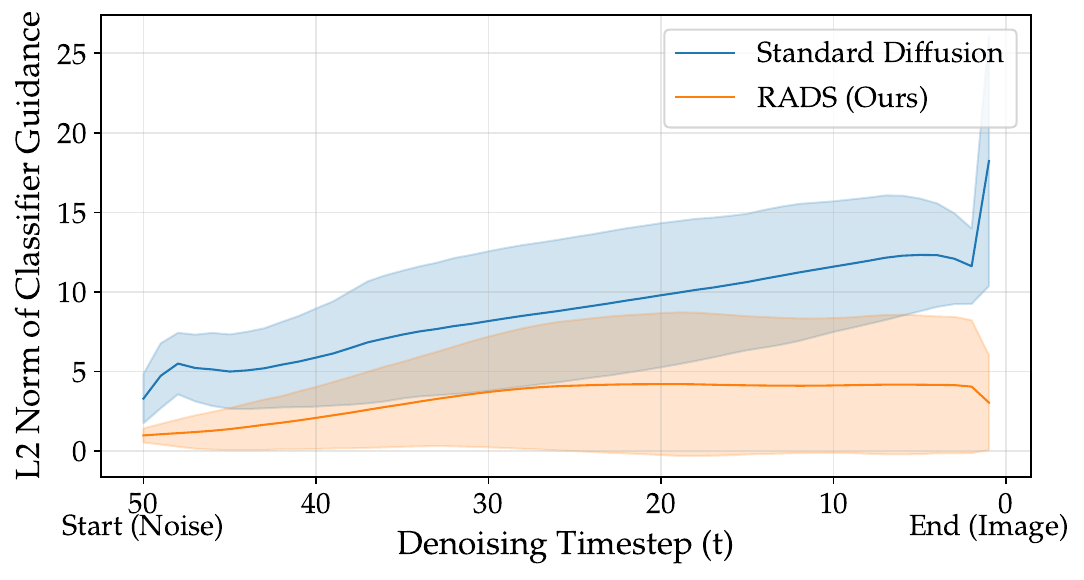}
    \caption{\textbf{Guidance norm with and without $\rads$.} Standard diffusion (blue) spikes as it collapses into the memorization basin. $\rads$ (orange) steers away from this peak.}
    \label{fig:guidance_norm}
\end{wrapfigure}

We track the classifier-free guidance norm $\|\epsilon_\theta(x_t,e_c)-\epsilon_\theta(x_t,\emptyset)\|_2$ over denoising. Unmitigated memorized prompts exhibit a sharp guidance-norm increase, consistent with the attraction-basin hypothesis of \citet{jain2025attraction}. In contrast, \ensuremath{\rads} suppresses this growth early, providing trajectory-level evidence that the learned policy steers away from high-guidance regions before memorization emerges. We compute these trajectories over 50 randomly sampled prompts, 10 evaluation seeds, and 5 training seeds.

\vspace{-0.5em}
\section{Computational Cost}
\label{app:compute}
\vspace{-0.5em}

\begin{wraptable}{r}{0.42\textwidth}
    \vspace{-10pt}
    \small
    \centering
    \caption{Average inference time per image on an A100 GPU.}
    \label{tab:inference_time_app}
    \begin{tabular}{@{}lc@{}}
        \toprule
        Method & Time (s) \\ 
        \midrule
        No mitigation & 2.295 ± 0.158 \\
        \citet{wen2024detecting} & 2.898 ± 0.046 \\
        \citet{ren2024crossattn} & 2.754 ± 0.040 \\
        \citet{hintersdorf2024nemo} & 104.090 ± 48.959 \\
        \citet{jain2025attraction} & $\textbf{2.232 ± 0.037}$ \\
        \textbf{$\rads$ (ours)} & 2.926 ± 0.034 \\
        \bottomrule
    \end{tabular}
    \vspace{-10pt}
\end{wraptable}

We evaluate the computational cost of $\rads$ in terms of offline training and inference-time overhead. Training the steering policy takes approximately 15 hours on a single A100 GPU. The caption-embedding VAE is trained separately and can be reused across diffusion models that share the same text-encoder embedding space.

At inference time, $\rads$ performs one actor forward pass per denoising step without modifying the diffusion model weights. Table~\ref{tab:inference_time_app} reports average inference time per image across 5 evaluation seeds and 10 prompts, with $\rads$ averaged over 5 training seeds. On a single A100 GPU, $\rads$ takes 2.926 $\pm$ 0.034 seconds per image, comparable to the fastest high-fidelity inference-time baselines and much faster than \citet{hintersdorf2024nemo}, whose neuron localization substantially increases latency.

This amortized policy avoids costly inference-time action search: unlike critic-only alternatives that must sample candidate actions during generation, $\rads$ pays the optimization cost during training and executes a learned state-conditioned control law at inference time.

\newpage

\section{Additional Qualitative Results}
\label{app:qualitative}
\vspace{-0.5em}

\subsection{Additional Comparisons Across Methods}
\label{app:additional_method_comparisons}

\begin{figure}[h]
    \small
    \centering
    \begin{subfigure}[t]{0.15\linewidth}
        \centering
        \includegraphics[width=\linewidth]{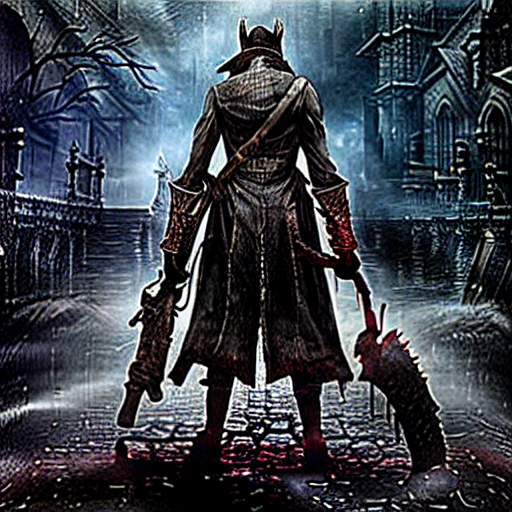}
        \caption{Memorized}
        \label{fig:quality_memorized}
    \end{subfigure}
    \hfill
    \begin{subfigure}[t]{0.15\linewidth}
        \centering
        \includegraphics[width=\linewidth]{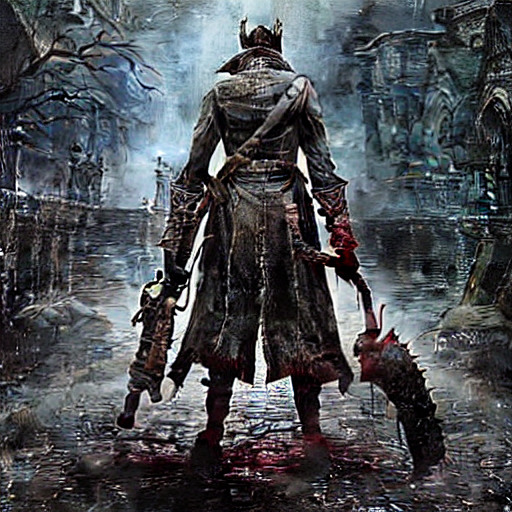}
        \caption{\citeauthor{wen2024detecting}}
        \label{fig:quality_wen}
    \end{subfigure}
    \hfill
    \begin{subfigure}[t]{0.15\linewidth}
        \centering
        \includegraphics[width=\linewidth]{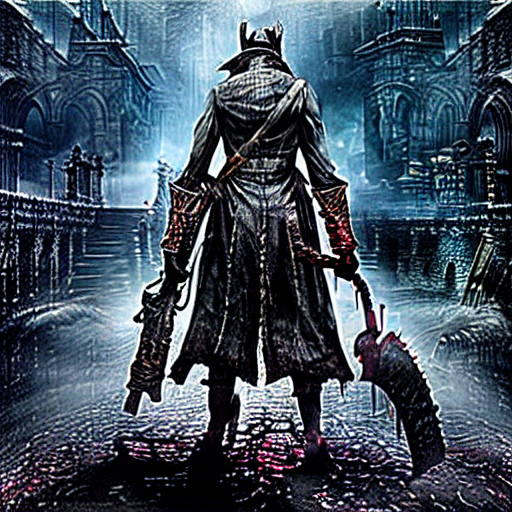}
        \caption{\citeauthor{ren2024crossattn}}
        \label{fig:quality_ren}
    \end{subfigure}
    \hfill
    \begin{subfigure}[t]{0.15\linewidth}
        \centering
        \includegraphics[width=\linewidth]{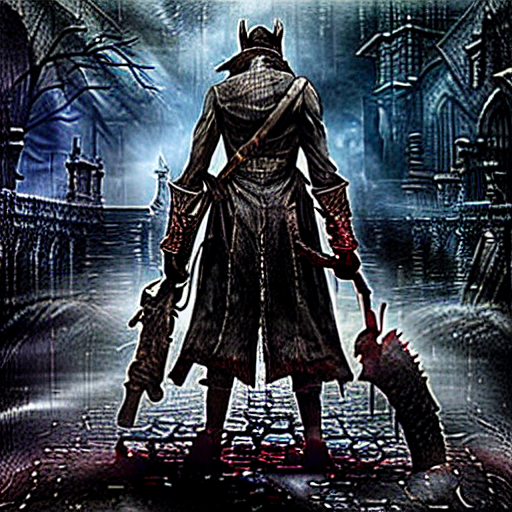}
        \caption{\tiny \citeauthor{hintersdorf2024nemo}}
        \label{fig:quality_hintersdorf}
    \end{subfigure}
    \hfill
    \begin{subfigure}[t]{0.15\linewidth}
        \centering
        \includegraphics[width=\linewidth]{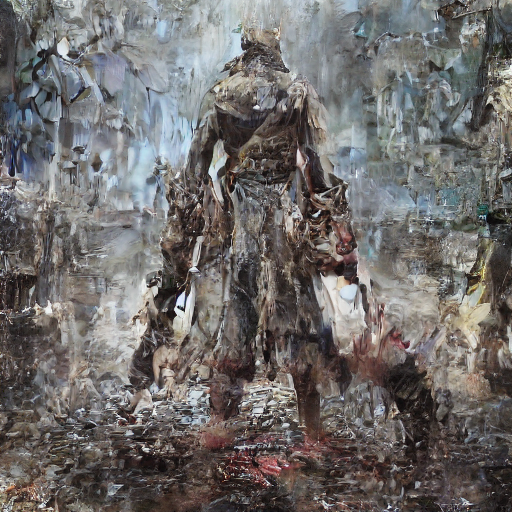}
        \caption{\citeauthor{jain2025attraction}}
        \label{fig:quality_jain}
    \end{subfigure}
    \hfill
    \begin{subfigure}[t]{0.15\linewidth}
        \centering
        \includegraphics[width=\linewidth]{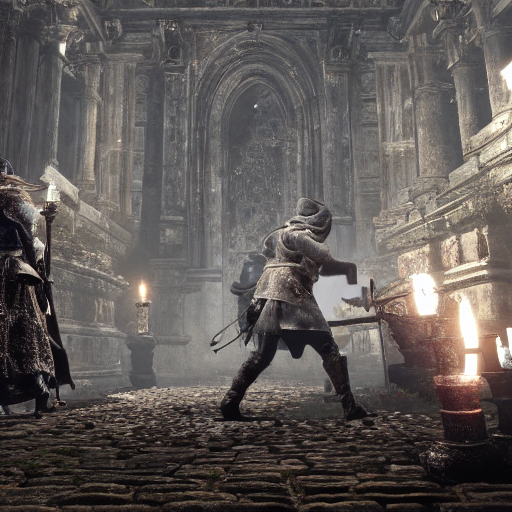}
        \caption{\textbf{$\rads$ (ours)}}
        \label{fig:quality_ours}
    \end{subfigure}
    \caption{
        \textbf{$\rads$ mitigates memorization while preserving image quality.}
        Generations for prompt \textit{\textless em\textgreater Bloodborne\textless /em\textgreater ~Video: Sony Explains the Game's Procedurally Generated Dungeons}.
    }
    \label{fig:quality}
\end{figure}

\begin{figure}[h]
    \small
    \centering
    \begin{subfigure}[t]{0.15\linewidth}
        \centering
        \includegraphics[width=\linewidth]{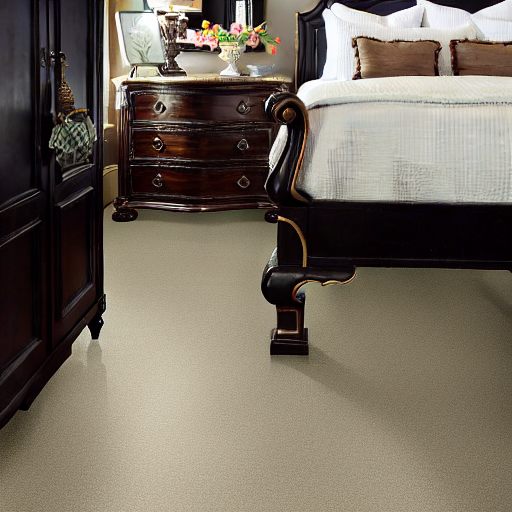}
        \caption{Memorized}
        \label{fig:alignment_memorized}
    \end{subfigure}
    \hfill
    \begin{subfigure}[t]{0.15\linewidth}
        \centering
        \includegraphics[width=\linewidth]{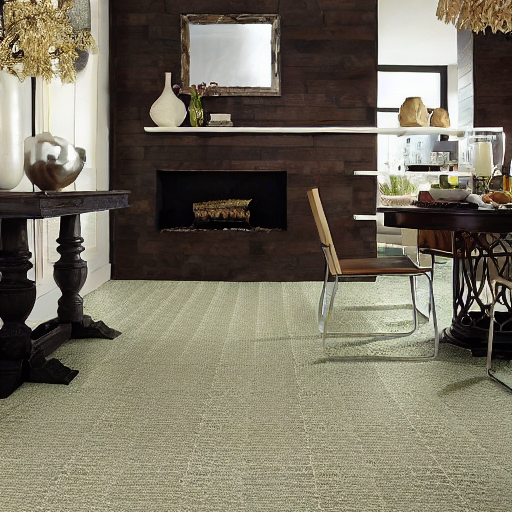}
        \caption{\citeauthor{wen2024detecting}}
        \label{fig:alignment_wen}
    \end{subfigure}
    \hfill
    \begin{subfigure}[t]{0.15\linewidth}
        \centering
        \includegraphics[width=\linewidth]{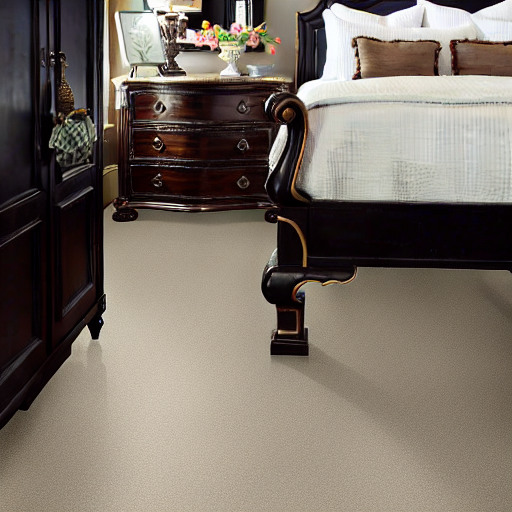}
        \caption{\citeauthor{ren2024crossattn}}
        \label{fig:alignment_ren}
    \end{subfigure}
    \hfill
    \begin{subfigure}[t]{0.15\linewidth}
        \centering
        \includegraphics[width=\linewidth]{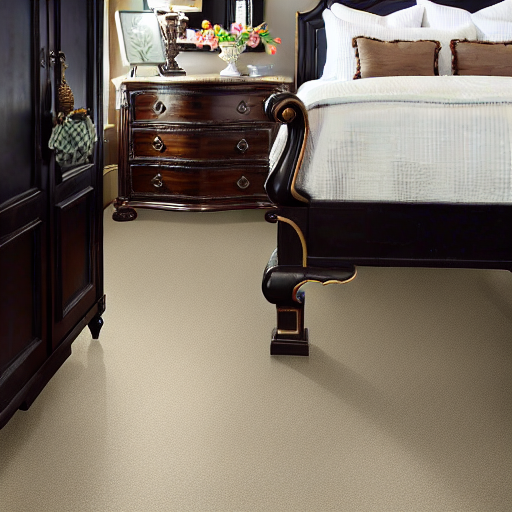}
        \caption{\tiny \citeauthor{hintersdorf2024nemo}}
        \label{fig:alignment_hintersdorf}
    \end{subfigure}
    \hfill
    \begin{subfigure}[t]{0.15\linewidth}
        \centering
        \includegraphics[width=\linewidth]{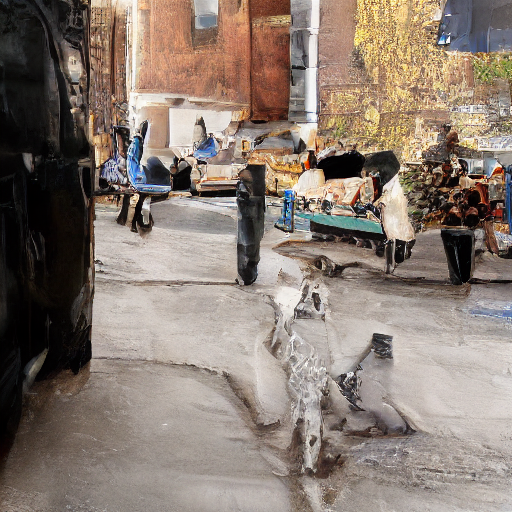}
        \caption{\citeauthor{jain2025attraction}}
        \label{fig:alignment_jain}
    \end{subfigure}
    \hfill
    \begin{subfigure}[t]{0.15\linewidth}
        \centering
        \includegraphics[width=\linewidth]{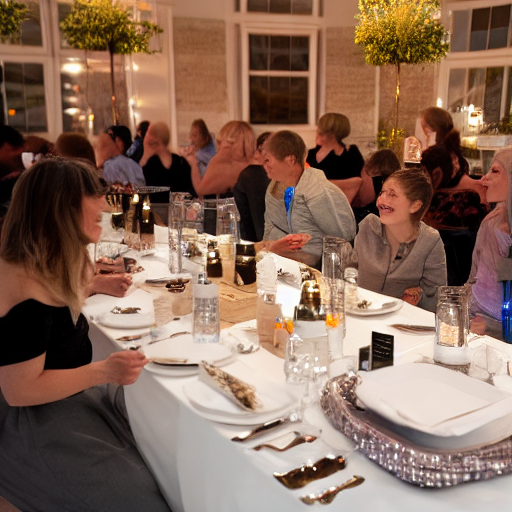}
        \caption{\textbf{$\rads$ (ours)}}
        \label{fig:alignment_ours}
    \end{subfigure}
    \caption{
        \textbf{$\rads$ can preserve prompt alignment while mitigating memorization.}
        Generations for prompt
        \textit{Shaw Floors Value Collections Fyc Tt II Net Dinner With Friends (t) 732T\_5E022}.
    }
    \label{fig:alignment}
\end{figure}

\begin{figure}[h]
    \small
    \centering
    \begin{subfigure}[t]{0.18\linewidth}
        \centering
        \includegraphics[width=\linewidth]{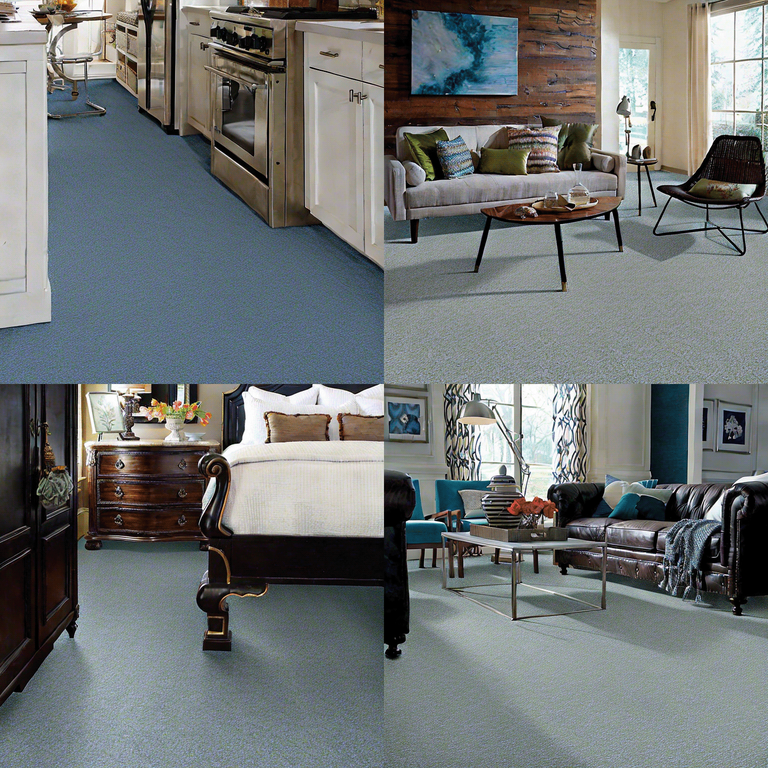}
        \caption{Memorized}
        \label{fig:sdv2_qualitative_none}
    \end{subfigure}
    \hfill
    \begin{subfigure}[t]{0.18\linewidth}
        \centering
        \includegraphics[width=\linewidth]{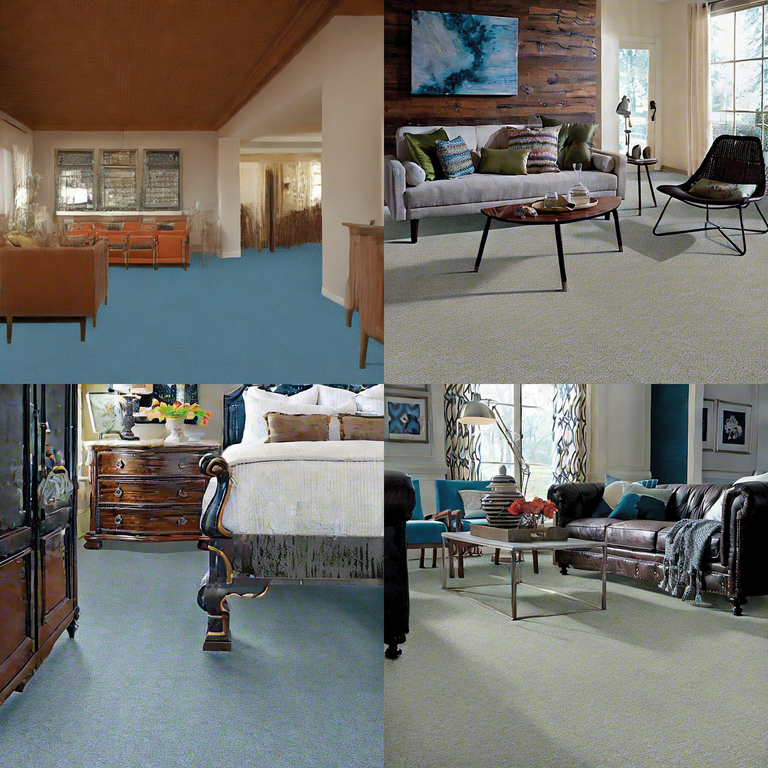}
        \caption{\citeauthor{wen2024detecting}}
        \label{fig:sdv2_qualitative_wen}
    \end{subfigure}
    \hfill
    \begin{subfigure}[t]{0.18\linewidth}
        \centering
        \includegraphics[width=\linewidth]{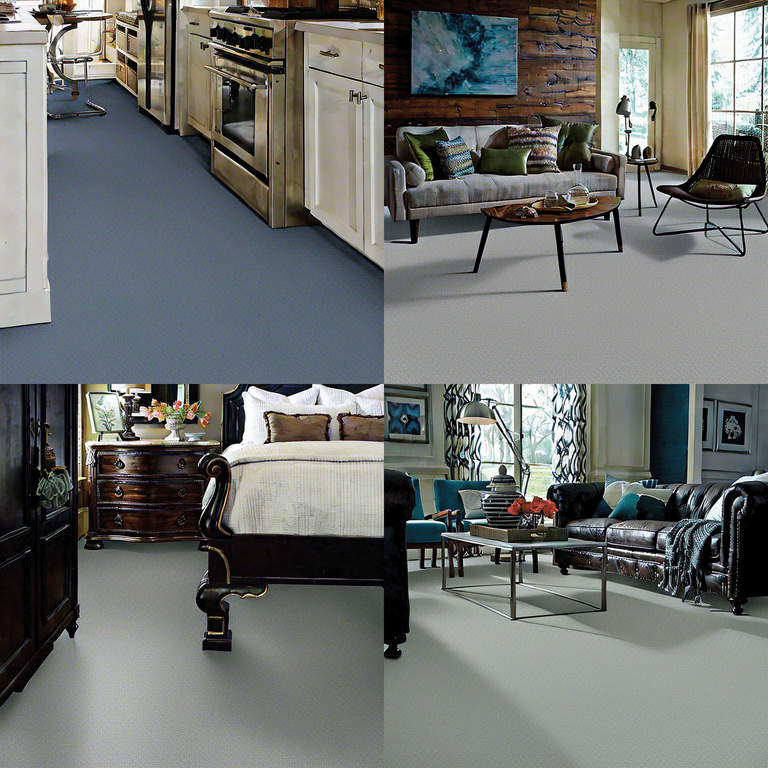}
        \caption{\citeauthor{ren2024crossattn}}
        \label{fig:sdv2_qualitative_ren}
    \end{subfigure}
    \hfill
    \begin{subfigure}[t]{0.18\linewidth}
        \centering
        \includegraphics[width=\linewidth]{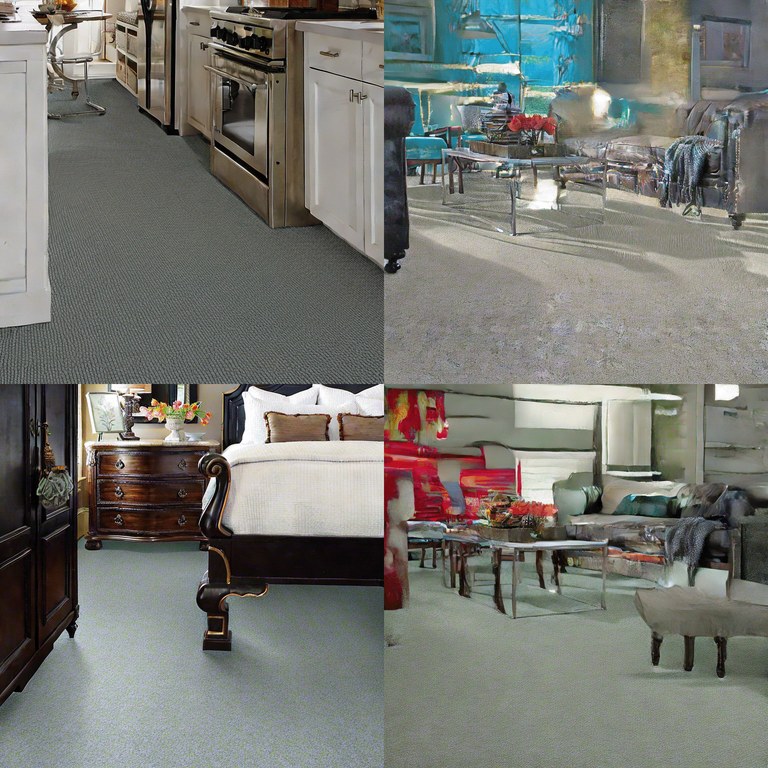}
        \caption{\tiny \citeauthor{jain2025attraction}}
        \label{fig:sdv2_qualitative_jain}
    \end{subfigure}
    \hfill
    \begin{subfigure}[t]{0.18\linewidth}
        \centering
        \includegraphics[width=\linewidth]{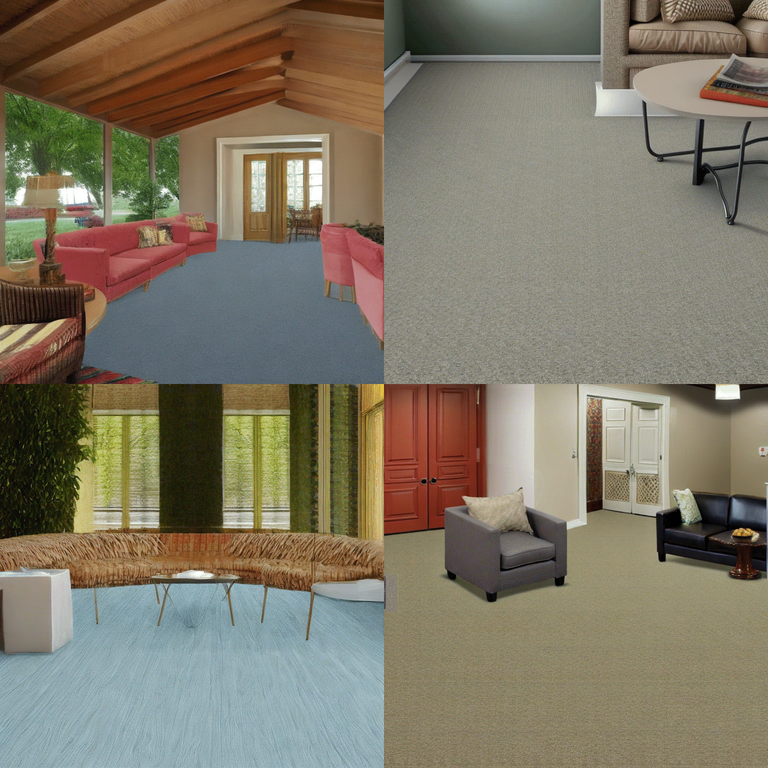}
        \caption{\textbf{\rads{} (ours)}}
        \label{fig:sdv2_qualitative_rads}
    \end{subfigure}
    \caption{
        \textbf{Similar patterns in prompt alignment and memorization mitigation extend to Stable Diffusion v2.1.}
        Generations for prompt \textit{Shaw Floors Roll Special Xv540 Tropical Wave 00420\_XV540}.
    }
    \label{fig:sdv2_qualitative}
\end{figure}

\begin{figure}[h]
\centering
\small
\setlength{\tabcolsep}{2pt}
\renewcommand{\arraystretch}{1.0}

\begin{tabular}{cccccc}
\footnotesize Memorized &
\footnotesize \citeauthor{wen2024detecting} &
\footnotesize \citeauthor{ren2024crossattn} &
\footnotesize \citeauthor{hintersdorf2024nemo} &
\footnotesize \citeauthor{jain2025attraction} &
\footnotesize \textbf{\rads{} (ours)} \\

\includegraphics[width=0.155\linewidth]{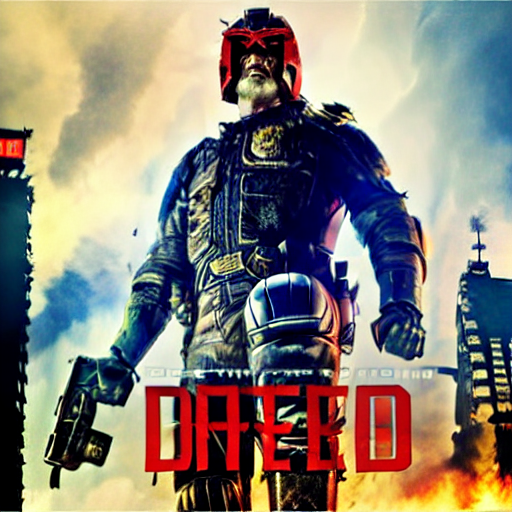} &
\includegraphics[width=0.155\linewidth]{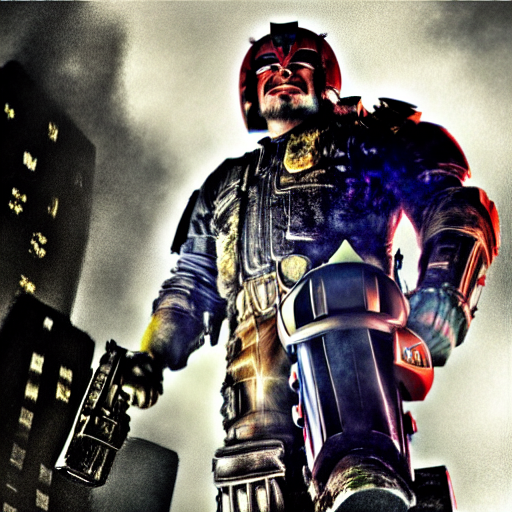} &
\includegraphics[width=0.155\linewidth]{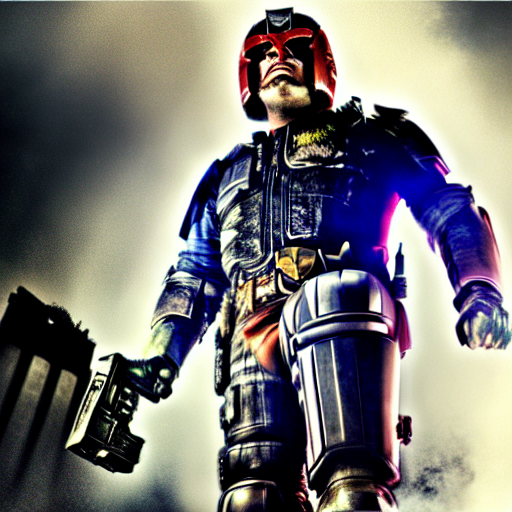} &
\includegraphics[width=0.155\linewidth]{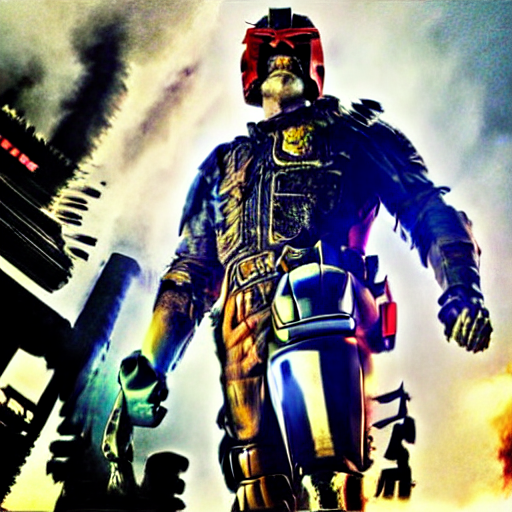} &
\includegraphics[width=0.155\linewidth]{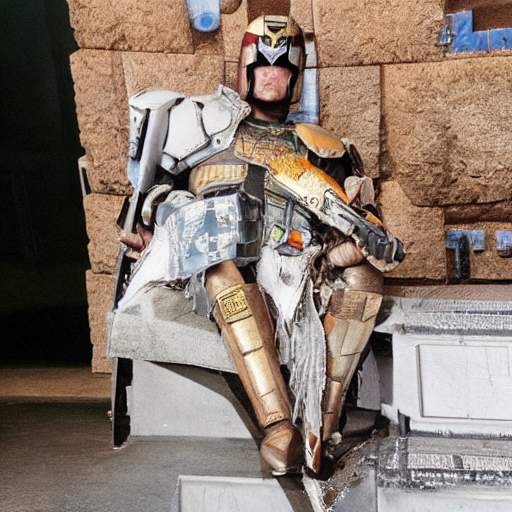} &
\includegraphics[width=0.155\linewidth]{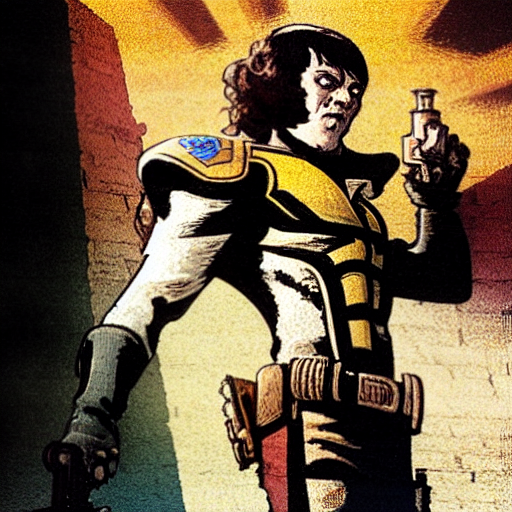} \\[-0.2em]
\multicolumn{6}{c}{\footnotesize \emph{Prompt:} ``<i>Dredd</i> Sequel Would Require More Star Power According to Producer.''} \\[0.5em]

\includegraphics[width=0.155\linewidth]{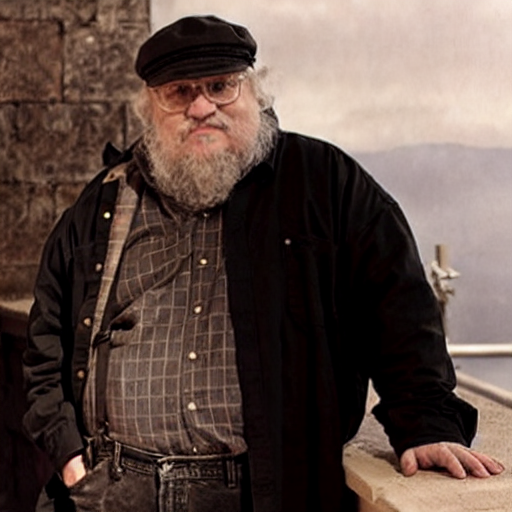} &
\includegraphics[width=0.155\linewidth]{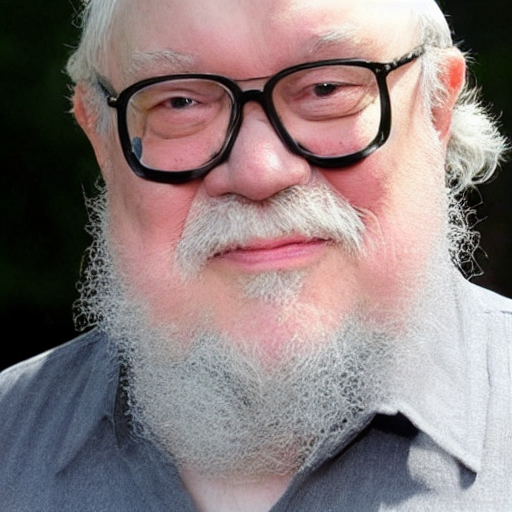} &
\includegraphics[width=0.155\linewidth]{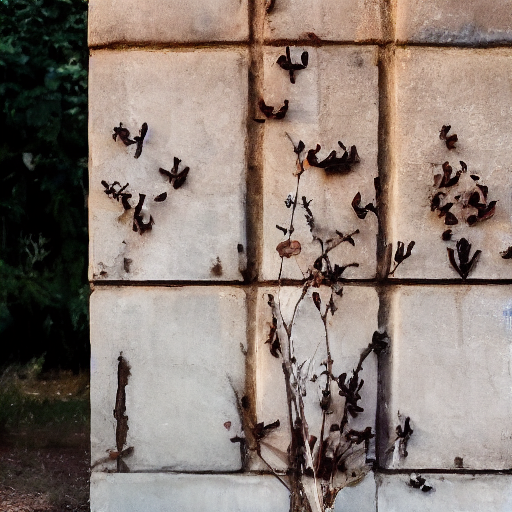} &
\includegraphics[width=0.155\linewidth]{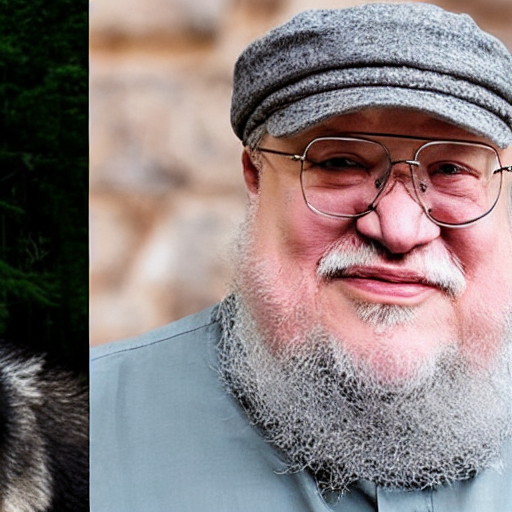} &
\includegraphics[width=0.155\linewidth]{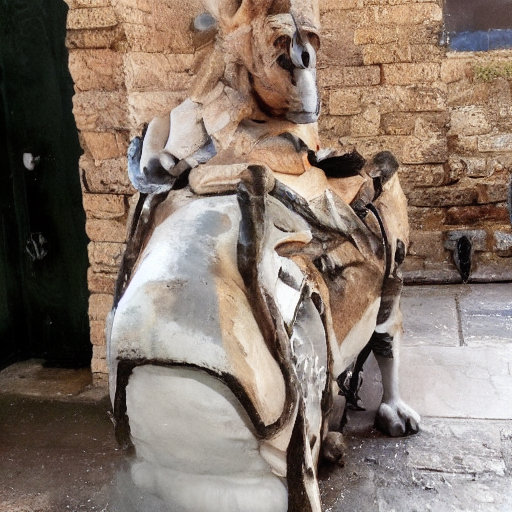} &
\includegraphics[width=0.155\linewidth]{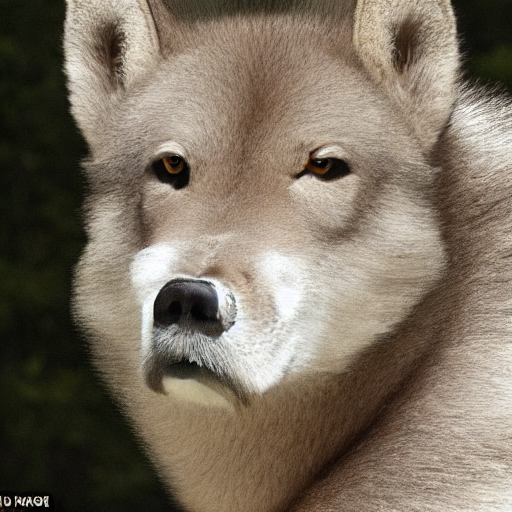} \\[-0.2em]
\multicolumn{6}{c}{\footnotesize \emph{Prompt:} ``George R.R. Martin Donates \$10{,}000 to Wolf Sanctuary for a 13-Year-Old Fan.''} \\[0.5em]

\includegraphics[width=0.155\linewidth]{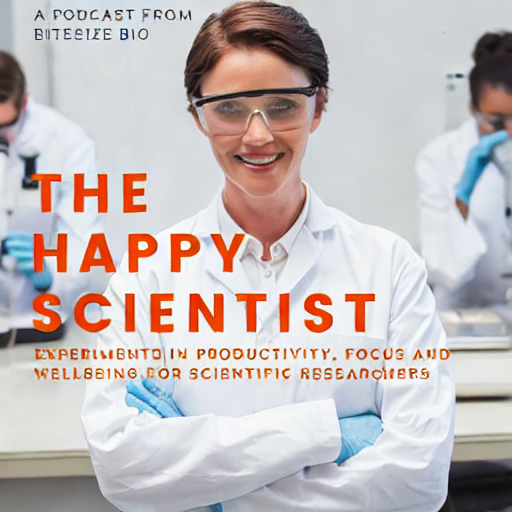} &
\includegraphics[width=0.155\linewidth]{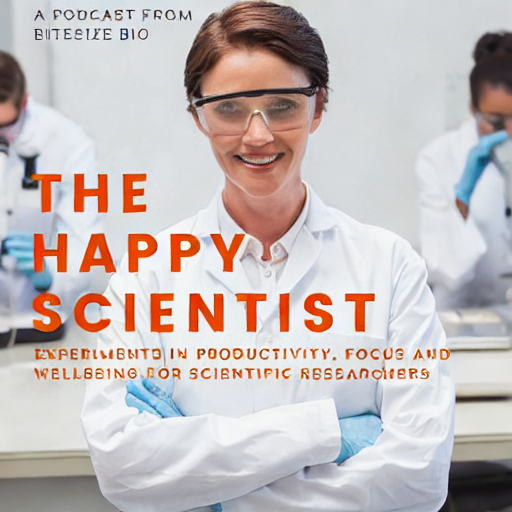} &
\includegraphics[width=0.155\linewidth]{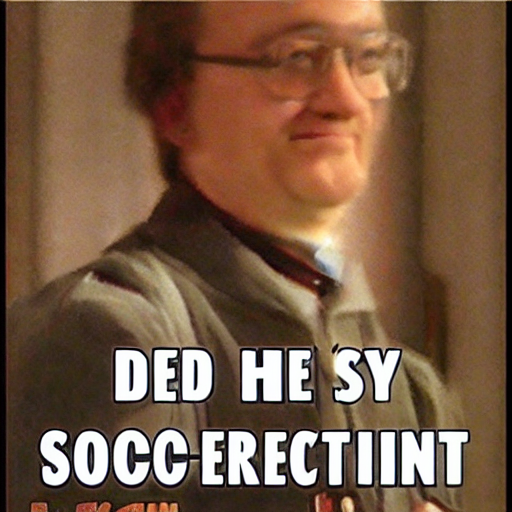} &
\includegraphics[width=0.155\linewidth]{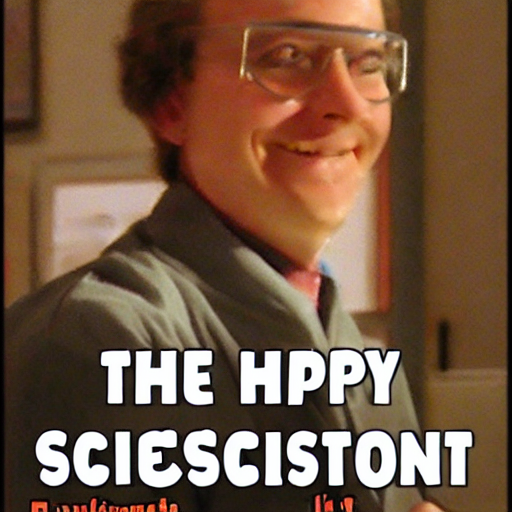} &
\includegraphics[width=0.155\linewidth]{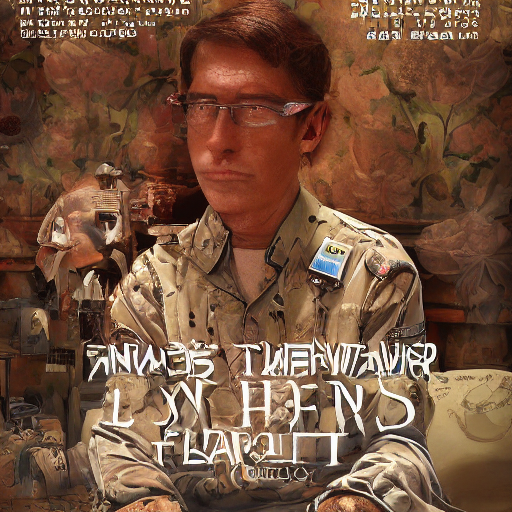} &
\includegraphics[width=0.155\linewidth]{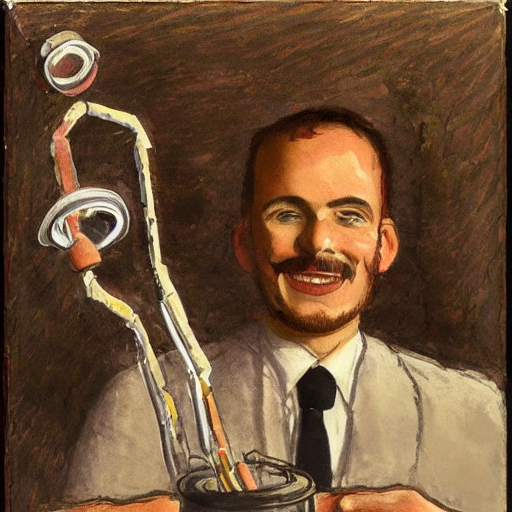} \\[-0.2em]
\multicolumn{6}{c}{\footnotesize \emph{Prompt:} ``The Happy Scientist''} \\[0.5em]

\includegraphics[width=0.155\linewidth]{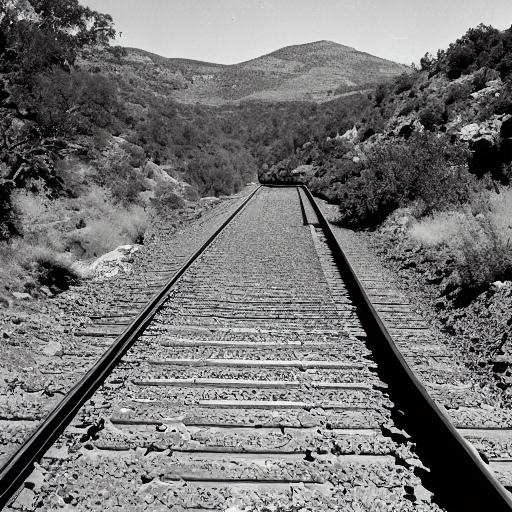} &
\includegraphics[width=0.155\linewidth]{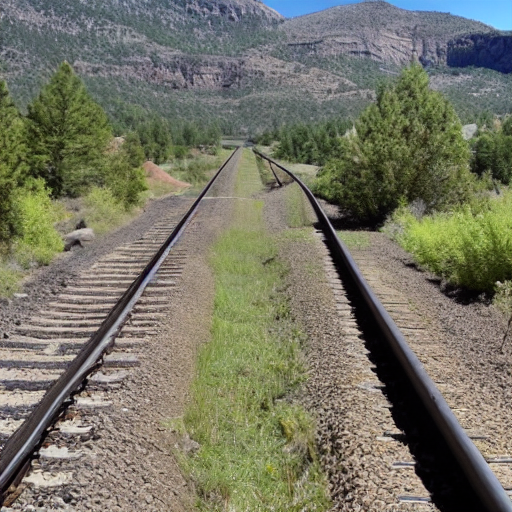} &
\includegraphics[width=0.155\linewidth]{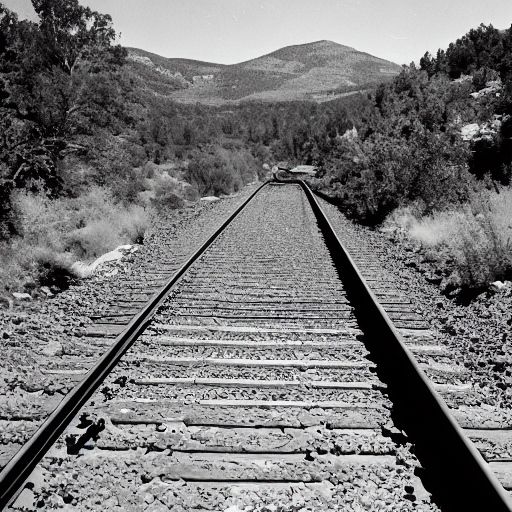} &
\includegraphics[width=0.155\linewidth]{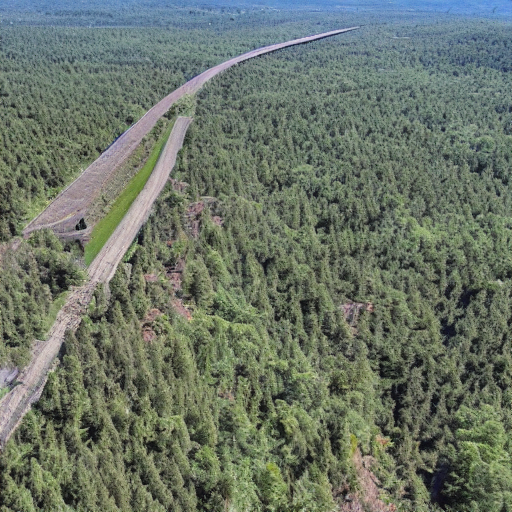} &
\includegraphics[width=0.155\linewidth]{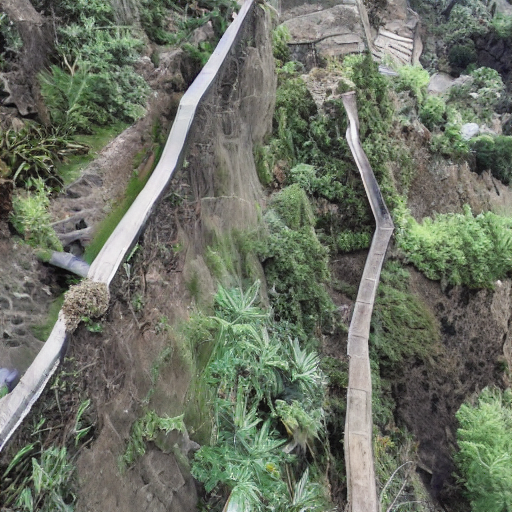} &
\includegraphics[width=0.155\linewidth]{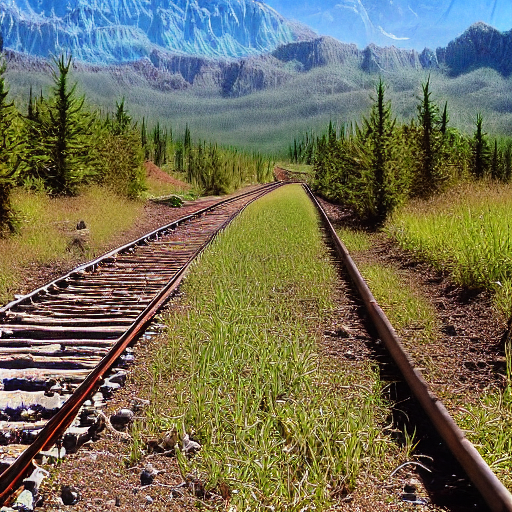} \\[-0.2em]
\multicolumn{6}{c}{\footnotesize \emph{Prompt:} ``View of straight railroad tracks near Deming through a rock cut with forested hills in distance.''} \\

\includegraphics[width=0.155\linewidth]{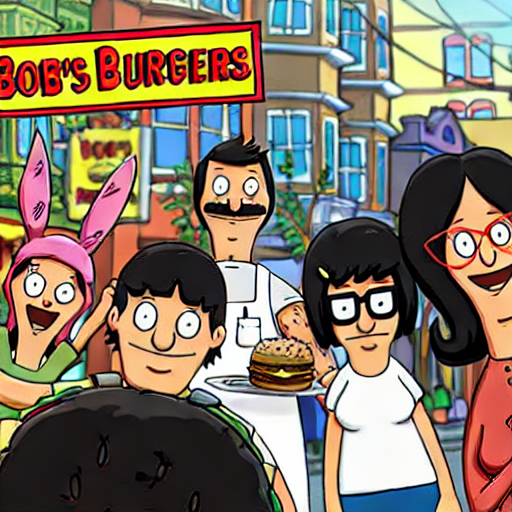} &
\includegraphics[width=0.155\linewidth]{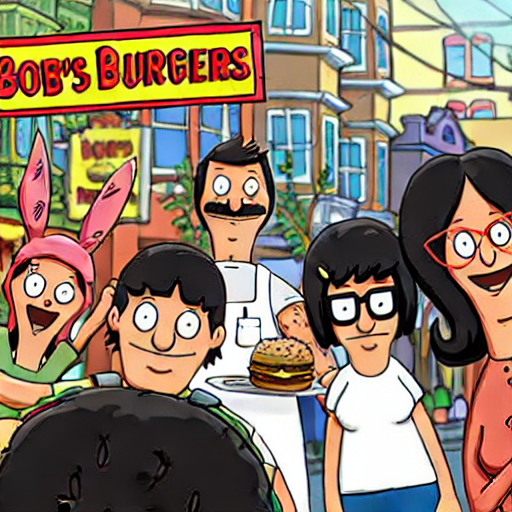} &
\includegraphics[width=0.155\linewidth]{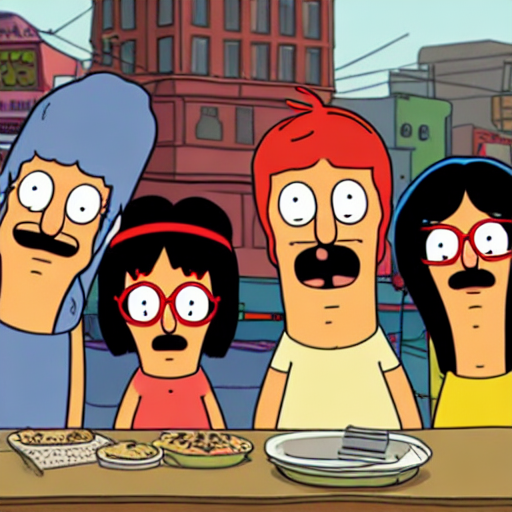} &
\includegraphics[width=0.155\linewidth]{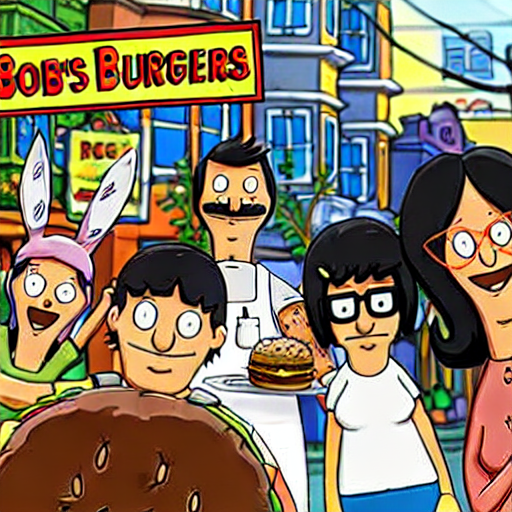} &
\includegraphics[width=0.155\linewidth]{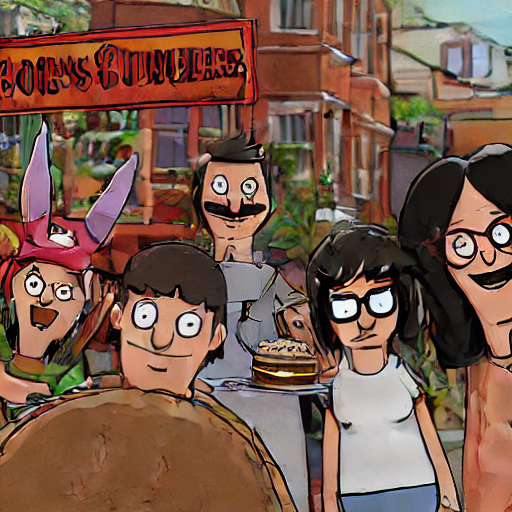} &
\includegraphics[width=0.155\linewidth]{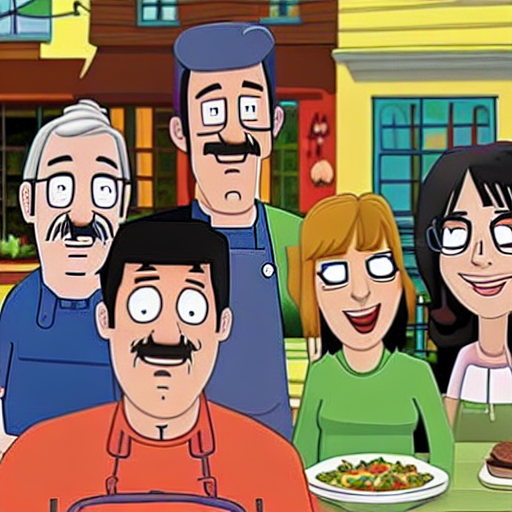} \\[-0.2em]
\multicolumn{6}{c}{\footnotesize \emph{Prompt:} ``<i>Bob's Burgers</i> To Publish Cookbook''} \\

\end{tabular}

\caption{
    \textbf{Additional qualitative comparisons across mitigation methods.}
    Each row corresponds to one prompt, with columns showing the unmitigated memorized output,
    prior inference-time baselines, and \rads. Overall, \rads{} maintains a stronger quality-diversity tradeoff.
}
\label{fig:additional_comparisons}
\end{figure}

Figures~\ref{fig:quality}--\ref{fig:additional_comparisons} provide additional qualitative comparisons across mitigation methods. In Figure~\ref{fig:sdv2_qualitative}, we illustrate a qualitative example of preserving prompt-alignment while mitigating memorization in Stable Diffusion v2.1, thus demonstrating that the trends extend to larger architectures.

\vspace{-0.5em}

\vspace{-0.5em}

\subsection{Failure Case of $\rads$: Semantic Drift}
\label{app:failure_cases}

\begin{wrapfigure}{r}{0.6\textwidth}
    \vspace{-1em}
    \centering
    \small

    \begin{minipage}[c]{0.46\linewidth}
        \centering
        \includegraphics[width=\linewidth]{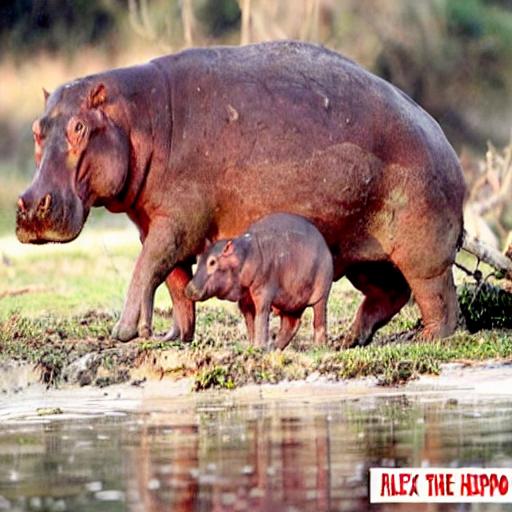}\\[-0.3em]
        {\scriptsize (a) Memorized}
    \end{minipage}
    \hspace{1em}
    \begin{minipage}[c]{0.44\linewidth}
        \centering
        \begin{tabular}{@{}cc@{}}
            \includegraphics[width=0.48\linewidth]{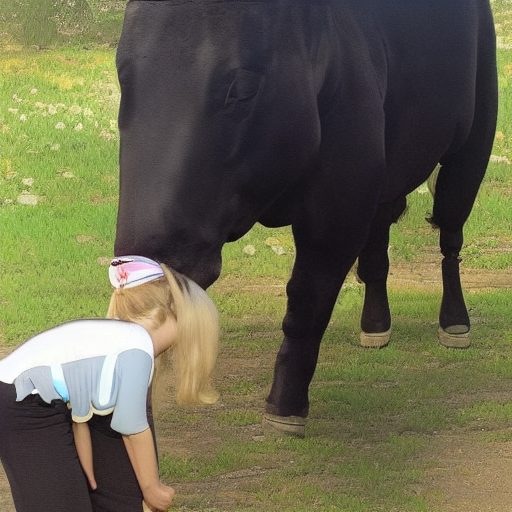} &
            \includegraphics[width=0.48\linewidth]{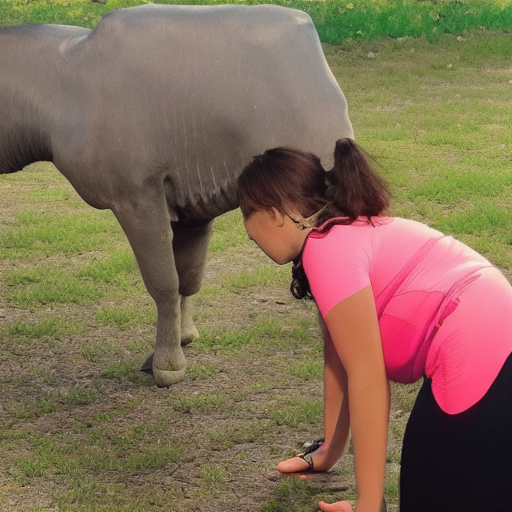} \\
            {\scriptsize (b) $\rads$ (seed 0)} & {\scriptsize (c) $\rads$ (seed 1)} \\[-0.1em]
            \includegraphics[width=0.48\linewidth]{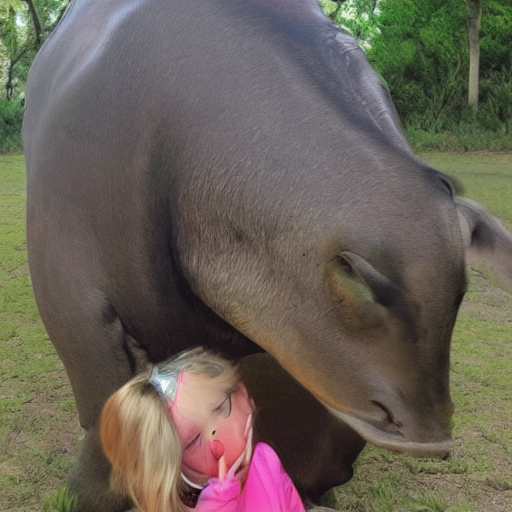} &
            \includegraphics[width=0.48\linewidth]{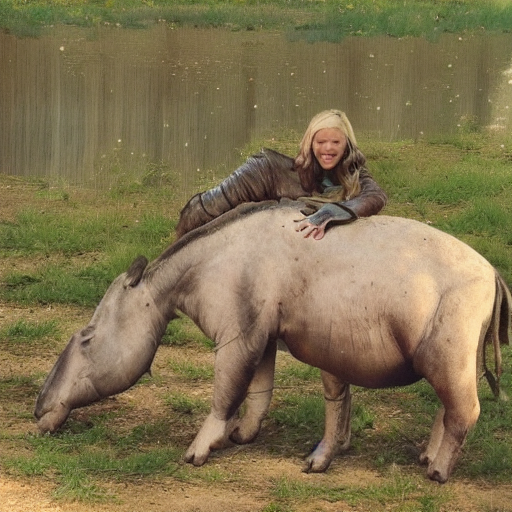} \\
            {\scriptsize (d) $\rads$ (seed 2)} & {\scriptsize (e) $\rads$ (seed 3)}
        \end{tabular}\\[-0.2em]
    \end{minipage}

    \caption{
        \textbf{Semantic drift.}
        For the held-out prompt \textit{Mothers influence on her young hippo}, $\rads$ mitigates
        the memorized output but can drift from the intended hippo semantics.
    }
    \label{fig:hippo_example}
    \vspace{-0.25em}
\end{wrapfigure}

One failure mode of $\rads$ is semantic drift on some out-of-distribution prompts.
For example, the held-out Webster prompt \textit{Mothers influence on her young hippo}
elicits a memorized image that matches the caption but reproduces the training instance.
After steering, $\rads$ avoids this memorized output but sometimes overemphasizes the
word ``mother,'' producing humans and inconsistently preserving the hippo concept
(Figure~\ref{fig:hippo_example}). This suggests that the current policy, trained on only
430 verified memorized prompts, may conflate specific memorization triggers with broader
semantic concepts. Incorporating more
memorized prompts with non-memorized prompts may improve prompt fidelity
while preserving mitigation.
\vspace{-0.5em}

\subsection{Artifacts from Guidance Truncation in \citet{jain2025attraction}}
\label{app:jain_artifacts}
\vspace{-0.5em}

\begin{figure}[H]
    \small
    \centering
    \begin{subfigure}[t]{0.14\linewidth}
        \centering
        \includegraphics[width=\linewidth]{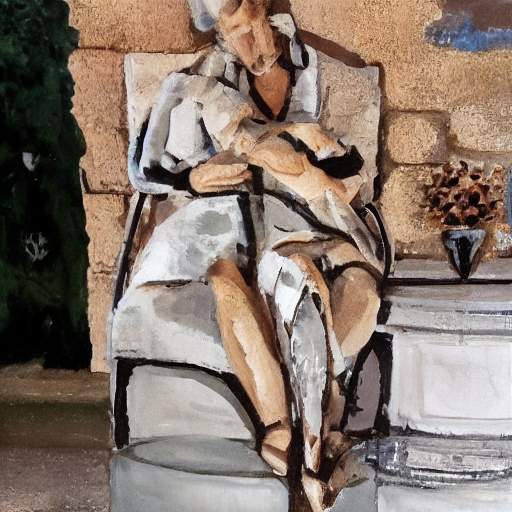}
        \caption{$g = 0$}
        \label{fig:jain_diversity_base}
    \end{subfigure}
    \hspace{1.5em}
    \begin{subfigure}[t]{0.14\linewidth}
        \centering
        \includegraphics[width=\linewidth]{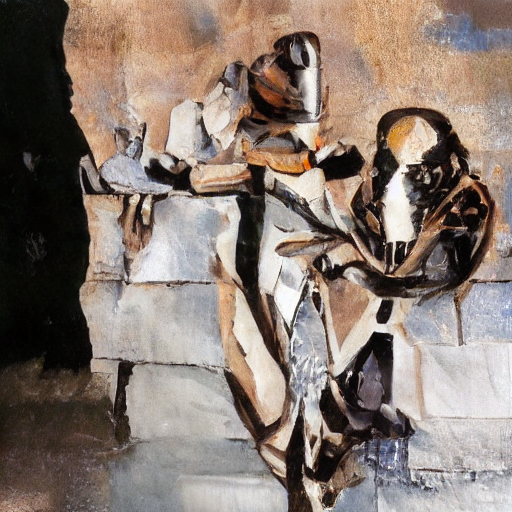}
        \caption{Prompt 1}
        \label{fig:jain_diversity_618}
    \end{subfigure}
    \hspace{1.5em}    
    \begin{subfigure}[t]{0.14\linewidth}
        \centering
        \includegraphics[width=\linewidth]{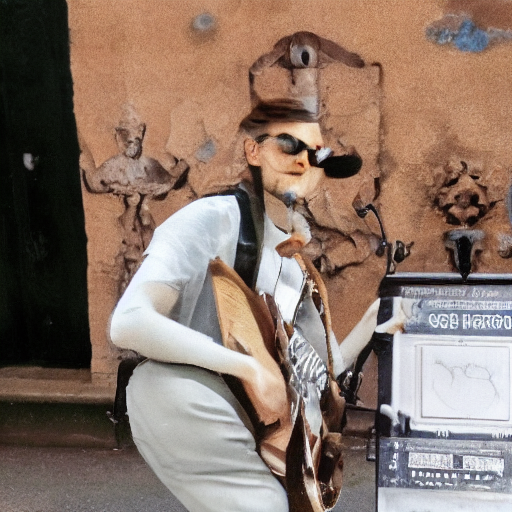}
        \caption{Prompt 2}
        \label{fig:jain_diversity_784}
    \end{subfigure}
    \hspace{1.5em} 
    \begin{subfigure}[t]{0.14\linewidth}
        \centering
        \includegraphics[width=\linewidth]{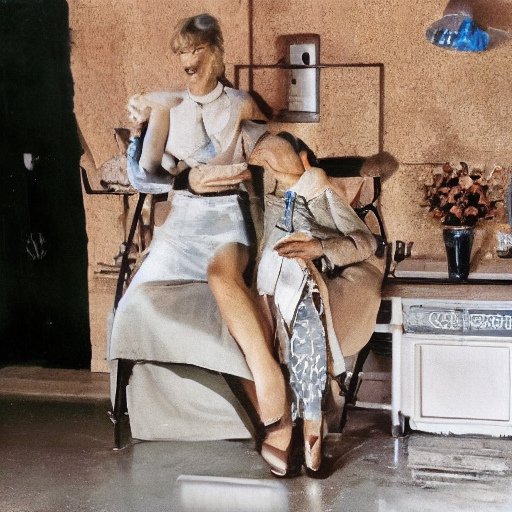}
        \caption{Prompt 3}
        \label{fig:jain_diversity_954}
    \end{subfigure}

    \caption{
        \textbf{Artifacts from guidance truncation in \citet{jain2025attraction}.}
        Images generated from the same initial latent $\latent_T$ resemble one another across prompts when text guidance is suppressed.
    }
    \label{fig:jain_diversity}
\end{figure}

\citet{jain2025attraction}
mitigates memorization by turning CFG on and off when an attraction
basin is detected. The resulting generation becomes
dominated by the initial latent $\latent_T$, which can produce artifacts and cause images
from different prompts to share visual similarities
(Figure~\ref{fig:jain_diversity}).

